\documentclass[preprint,12pt]{elsarticle}

\journal{Information Fusion}

\usepackage{caption}
\usepackage{subcaption}
\usepackage{multirow}
\usepackage{array}
\usepackage{amsmath,amsfonts,amssymb}
\usepackage{bm} 
\usepackage{mathtools} 

\usepackage{tikz}
\usetikzlibrary{mindmap}
\usepackage{tabularx}
\usepackage{booktabs}
\usepackage{pifont}
\usepackage{xcolor}
\usepackage[a4paper,margin=1in]{geometry}
\newcolumntype{C}{>{\centering\arraybackslash}m{1.5cm}} 
\newcolumntype{Y}{>{\centering\arraybackslash}X} 

\newcommand{\dtbox}[2]{%
  \begingroup
  \setlength{\fboxsep}{1pt}%
  \fcolorbox{black}{#1}{\scriptsize\textcolor{white}{\strut #2}}%
  \endgroup
}

\newcommand{\dtinactive}[1]{%
  \begingroup
  \setlength{\fboxsep}{1pt}%
  \fcolorbox{black}{gray!20}{\scriptsize\textcolor{black}{\strut #1}}%
  \endgroup
}

\definecolor{graphcol}{HTML}{4E79A7}
\definecolor{imagecol}{HTML}{F28E2B}
\definecolor{speechcol}{HTML}{E15759}
\definecolor{textcol}{HTML}{76B7B2}
\definecolor{tabcol}{HTML}{59A14F}
\definecolor{videocol}{HTML}{B07AA1}   
\definecolor{codecol}{HTML}{9C755F}    

\newcommand{\datatypebar}[7]{%
\ifnum#1=1\relax\dtbox{black}{G}\else\dtinactive{G}\fi
\ifnum#2=1\relax\dtbox{black}{I}\else\dtinactive{I}\fi
\ifnum#3=1\relax\dtbox{black}{A}\else\dtinactive{A}\fi
\ifnum#4=1\relax\dtbox{black}{L}\else\dtinactive{L}\fi
\ifnum#5=1\relax\dtbox{black}{T}\else\dtinactive{T}\fi
\ifnum#6=1\relax\dtbox{black}{V}\else\dtinactive{V}\fi
\ifnum#7=1\relax\dtbox{black}{C}\else\dtinactive{C}\fi
}

\newcommand{\evalbar}[8]{%
\ifnum#1=1\relax\dtbox{black}{FA}\else\dtinactive{FA}\fi
\ifnum#2=1\relax\dtbox{black}{LO}\else\dtinactive{LO}\fi
\ifnum#3=1\relax\dtbox{black}{RO}\else\dtinactive{RO}\fi
\ifnum#4=1\relax\dtbox{black}{CO}\else\dtinactive{CO}\fi
\ifnum#5=1\relax\dtbox{black}{RA}\else\dtinactive{RA}\fi
\ifnum#6=1\relax\dtbox{black}{AX}\else\dtinactive{AX}\fi
\ifnum#7=1\relax\dtbox{black}{QL}\else\dtinactive{QL}\fi
\ifnum#8=1\relax\dtbox{black}{QN}\else\dtinactive{QN}\fi
}

\usepackage{hyperref}
\begin{document}

\begin{frontmatter}



\title{Decoding the Multimodal Maze: A Systematic Review on the Adoption of Explainability in Multimodal Attention-based Models}

\author[aau]{Md Raisul Kibria\corref{cor1}}
\cortext[cor1]{Corresponding author.}

\author[aau]{Sébastien Lafond}
\author[psu]{Janan Arslan}
\affiliation[aau]{organization={Faculty of Science and Engineering, Information Technology, Åbo Akademi University},
            city={Turku},
            postcode={20500}, 
            country={Finland}}

\affiliation[psu]{organization={Sorbonne Université, Institut du Cerveau - Paris Brain Institute},
            city={Paris},
            postcode={F-75013}, 
            country={France}}

\tnotetext[doi]{Published version available at
\href{https://doi.org/10.1016/j.inffus.2026.104405}
{https://doi.org/10.1016/j.inffus.2026.104405}}
\begin{abstract}
Multimodal learning has witnessed remarkable advancements in recent years, particularly with the integration of attention-based models, leading to significant performance gains across a variety of tasks. Parallel to this progress, the demand for explainable artificial intelligence (XAI) has spurred a growing body of research aimed at interpreting the complex decision-making processes of these models. This systematic literature review analyzes research published between January 2020 and early 2024 that focuses on the explainability of multimodal models. Framed within the broader goals of XAI, we examine the literature across multiple dimensions, including model architecture, modalities involved, explanation algorithms and evaluation methodologies. Our analysis reveals that most studies are concentrated on vision-language and language-only models, with attention-based techniques being the most commonly employed for explanation. However, these methods often fall short in capturing the full spectrum of interactions between modalities, a challenge further compounded by the architectural heterogeneity across domains. Importantly, we find that evaluation methods for XAI in multimodal settings are largely non-systematic, lacking consistency, robustness, and consideration for modality-specific cognitive and contextual factors. To address these gaps, we not only synthesize findings from the surveyed works but also incorporate a complementary analysis that integrates recent and emerging advances driving multimodal explainability. Based on these insights, we provide a comprehensive set of recommendations aimed at promoting rigorous, transparent, and standardized evaluation and reporting practices in multimodal XAI research. Our goal is to support future research in more interpretable, accountable, and responsible multimodal AI systems, with explainability at their core.

\end{abstract}



\end{frontmatter}

\section{Introduction}
Explainability in deep neural networks refers to the study of uncovering the internal mechanisms of artificial intelligence (AI) models, making their decisions more interpretable and transparent, in contrast to their traditionally black-box and opaque nature. There currently exists no universal definition regarding explainability. However, in the context of our work, we adopt the definition highlighted by Guidotti \emph{et al.}, which describes explainability as an interface that functions both as a proxy for the decision-maker and as a construct comprehensible to end users \cite{guidotti2018survey}. In the literature, a distinction is often made between explainability and interpretability, with the latter generally referring to the cognitive process of generating meaning (e.g., a model-based feature could represent a clinical variability). However, in this paper, we use the terms interchangeably to encompass the broader concept \cite{arrieta2020explainable}. As machine learning (ML) models have grown increasingly powerful, diverse, and accurate, there has been a parallel surge in the development of methods aimed at explaining their decisions \cite{burkart2021survey}, driven by growing domain-specific demands \cite{arrieta2020explainable, yang2022unbox} and increasingly stringent regulatory requirements \cite{nannini2023explainability, chun2024comparative}. Nevertheless, explainable AI (XAI) remains an elusive goal for several reasons: the absence of a unified definition of a ``valid explanation", inconsistent reporting practices, differing and often subjective stakeholder needs, and the lack of universally accepted evaluation metrics. These challenges are further amplified in the case of multimodal ML models, which simultaneously process and learn from multiple data sources. Multimodal models vary significantly in terms of the number and types of modalities, fusion mechanisms, task objectives, and application domains. Consequently, they introduce additional complexity in both the generation and evaluation of explanations. Current research on explainability for such models mainly focuses on interpreting inter-modal interactions and identifying suitable evaluation criteria. However, comprehensive studies that systematically assess and compare existing explanation methods for multimodal models are still scarce, despite their importance for standardization and benchmarking efforts.

In the last decade, advances in storage capacity, internet infrastructure, and computational resources have facilitated the creation of numerous large-scale multimodal datasets across various domains, such as COCO \cite{lin2014microsoft}, GLUE \cite{wang-etal-2018-glue}, ROCO \cite{pelka2018radiology}, VQA \cite{antol2015vqa}, Visual Genome \cite{krishna2017visual}, and MIMIC \cite{johnson2016mimic}, among others. However, modeling these datasets presents substantial challenges, including increased task complexity and architectural limitations such as training collapse, non-convergence, and instability, particularly in generative adversarial networks (GANs) \cite{saxena2021generative}. These tasks often involve processing long sequences of intricately connected tokens, which traditional models struggled to handle effectively due to limitations in contextual understanding \cite{alzubaidi2021review, liu2022combining}. The introduction of the attention mechanism provided a breakthrough by enabling the generation of contextually weighted vectors that conditionally link input tokens to downstream modules. Originally developed for neural machine translation (NMT) \cite{bahdanau2014neural}, attention paved the way for the transformer architecture, which streamlined and generalized the attention mechanism \cite{vaswani2017attention}. Transformers eliminated the need for recurrent or convolutional layers and instead relied solely on attention to model relationships within and between input and output tokens. This made them highly adaptable to other modalities such as vision \cite{dosovitskiy2020image}, video \cite{liu2022video}, and various multimodal tasks, thereby significantly advancing the field of multimodal learning (MML).

Modern attention-based multimodal models have achieved substantial progress across diverse application areas, demonstrating improvements in benchmark performance \cite{xu2023multimodal} and generative modeling capabilities \cite{sengar2024generative}. In addition to their accuracy and scalability, attention-based models offer unique opportunities for explainability. Beyond traditional model-agnostic approaches, attention weights offer a model-specific lens through which token interactions can be interpreted, potentially yielding meaningful insights into model behavior \cite{abnar2020quantifying}. However, most explanation efforts have focused narrowly on self-attention mechanisms that capture intra-modal relationships. Other critical components of transformers, such as skip-connections, responsible for a significant share of information flow, have not been thoroughly examined \cite{qiang_attcat_nodate}. Cross-attention layers, which encode inter-modal dependencies by linking queries and values from different modalities, also remain difficult to interpret due to their inherent complexity. Attempts to adapt model-agnostic explanation methods to multimodal settings have been limited and often suffer from high computational costs \cite{parcalabescu2022mm}. Existing studies also tend to focus predominantly on specific modality combinations, such as vision-language (VL) tasks \cite{rodis2024multimodal}, leaving other modality pairings underexplored. Moreover, key aspects of the explanation framework, such as the interfaces used to present explanations and the methods used to evaluate them, remain understudied. While foundational work exists on evaluation frameworks for XAI \cite{doshi2017towards, gilpin2018explaining, guidotti2018survey}, the challenges in multimodal context have resulted in fragmented, non-systematic approaches to explanation evaluation.

Given the multitude of interconnected factors involved in explaining multimodal models, a systematic examination of the current research landscape is both timely and necessary. With the rapid advancements in state-of-the-art multimodal architectures, it is important to track the evolution, promises, and limitations of explainability within this context. Previous systematic reviews by Fantozzi \emph{et al.}~\cite{fantozzi2024explainability} and Rodis \emph{et al.}~\cite{rodis2024multimodal} have addressed related areas. Fantozzi \emph{et al.} focus on explanation algorithms for transformer models, while Rodis \emph{et al.} explore explainability in multimodal AI models. In contrast, our review concentrates specifically on attention-based architectures due to their widespread adoption and performance advantages in multimodal tasks. Unlike prior studies, our approach considers explainability holistically and includes additional studies that, although not strictly multimodal, utilize multiple input streams (hereby referred to as `multichannel' and further explained in Section~\ref{Modalities}) or involve generative modeling. Thus, while some overlap exists with previous work, our review protocol incorporates and extends existing taxonomies, and provides comprehensive discussions on all major components of the explanation framework \cite{mohseni2021multidisciplinary}. Table \ref{tab:survey} highlights key differences in focus and coverage between our review and prior work.

\begin{table*}[t]
\renewcommand{\arraystretch}{1.3}
\centering
\caption{Classification and Comparison of Related Literature Reviews \\
\textbf{Legend:} \ding{51} = Discussion included, \ding{55} = Discussion not included}
\label{tab:survey}
\scriptsize
\begin{tabularx}{\textwidth}{C|Y|Y|C|C|C|C}
\hline
\textbf{Reference} & \textbf{Review Type} & \textbf{Key Focus Area} & \textbf{Model Architecture} & \multicolumn{3}{c}{\textbf{Explainability}} \\
\cline{5-7}
 & & & & \textbf{Evaluation Dataset} & \textbf{Algorithm} & \textbf{Evaluation} \\
\hline
Xu \emph{et al.} \cite{xu2023multimodal} & Holistic Review & Transformer variants used in multimodal learning & \ding{51} & \ding{55} & \ding{55} & \ding{55} \\
\hline
Fantozzi \emph{et al.} \cite{fantozzi2024explainability} & Systematic Review & Explainability of Transformer models in unimodal and multimodal contexts & \ding{55} & \ding{55} & \ding{51} & \ding{55} \\
\hline
Rodis \emph{et al.} \cite{rodis2024multimodal} & Systematic Review (methodology not fully reported) & Explainability of AI models used in multimodal learning & \ding{55} & \ding{51} & \ding{51} & \ding{51} \\
\hline
Our work & Systematic Review (methodology fully reported) & Explainability of attention-based models used in multimodal learning & \ding{51} & \ding{51} & \ding{51} & \ding{51} \\
\hline
\end{tabularx}
\end{table*}

\begin{figure}
    \centering
    \includegraphics[width=1\linewidth]{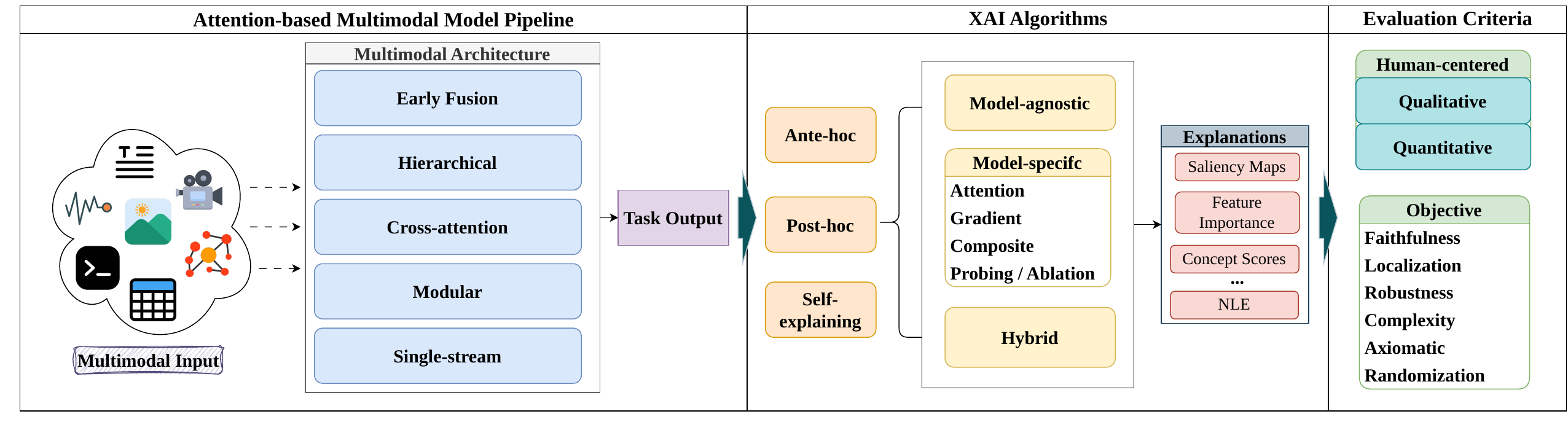}
    \caption{Pipeline for multimodal XAI evaluation, showing (1) attention-based multimodal architecture taxonomy, (2) XAI algorithms taxonomy, and (3) evaluation criteria.}
    \label{fig:block_whole}
\end{figure}

The main contributions of this paper are as follows:
\begin{enumerate}
    \item We present a comprehensive catalog of multimodal applications studied over the past four years, analyzed through the lens of explainability. Alongside systematically collecting key studies, we examine the associated domains, tasks, and the evaluation datasets — a critical factor in XAI research.
    \item We adopt and extend existing taxonomies to facilitate detailed discussion across three dimensions (as depicted in Figure~\ref{fig:block_whole}):
     \begin{itemize}
         \item \textit{Attention-based architectures:} Classification of studies based on how and where different input streams are fused within the model.
         \item \textit{XAI algorithms:} Categorization of explanation algorithms, along with technical analysis and application contexts.
         \item \textit{Explainability-oriented evaluation methods:} Grouping of evaluation strategies and discussion contextualized to the XAI method and use-case.
     \end{itemize}
    \item Beyond the surveyed literature, we incorporate a synthesis of recent advances in multimodal architectures, interpretability techniques, and emerging evaluation frameworks that complement and expand the scope of existing work.
    \item We identify major challenges in achieving explainability for multimodal models and provide extensive future research recommendations based on our findings.
\end{enumerate}

The remainder of this review is organized as follows: Section \ref{Method} provides a comprehensive discussion of the design and methodology for the systematic literature review (SLR), while Section \ref{Qualitative Synthesis Overview} presents an overall summary of the findings from the selected publications included in the review. All other sections focus on the qualitative synthesis of information from the publications, analyzed from various perspectives. Section \ref{Modalities} and Section \ref{Tasks and Datasets} describe the combinations of input-output modalities used and the application areas along with explainability evaluation datasets, respectively. Section \ref{Attention-based architectures} and Section \ref{Explanation Algorithm} systematically analyze the model architectures and explanation algorithms employed in the reviewed studies. Section \ref{Evaluation Criteria} elaborates on the criteria used to evaluate solutions based on explainability and contextualizes how decisions made in other sections influence these criteria. Publications introducing visualization tools for explainability as interfaces are discussed in Section \ref{Explanation Interfaces}. Section~\ref{sec:Beyond} synthesizes additional literature beyond the primary search to highlight emerging multimodal architectures, explainability methods, and evaluation frameworks. Finally, a critical assessment of the findings, along with future recommendations, is presented in Section \ref{Recommendation and Future Directions}, and the conclusion is drawn in Section \ref{Conclusion}.

\section{Method\label{Method}}
This SLR is conducted in accordance with the established guidelines proposed by Kitchenham \cite{kitchenham2004procedures}, focusing on the analysis of scholarly works related to the explainability of multimodal attention-based models. To comprehensively capture the multifaceted dimensions of this research area, the study formulates the following research questions:

\begin{enumerate}

\item What datasets are commonly used for evaluating explainability in various multimodal tasks and domains?
\textit{(Addressed in Section \ref{Tasks and Datasets})}
\item What types of attention-based architectures have been employed in the development of models for multimodal tasks?
\textit{(Addressed in Section \ref{Attention-based architectures})}
\item Which explanation methods are utilized in multimodal attention-based models, and how are these methods categorized? What is the distribution of these categories across the literature?
\textit{(Addressed in Section \ref{Explanation Algorithm})}
\item What are the key criteria that define effective explanations, and how do evaluation methodologies vary across different application domains?
\textit{(Addressed in Section \ref{Evaluation Criteria})}
\item In what ways do evaluation approaches account for the multimodal nature of the tasks being addressed?
\textit{(Discussed in Section \ref{sec:mm_interactions})}
\end{enumerate}

By analyzing the various elements surrounding our primary research objective, along with related concepts and integrated insights, we propose a comprehensive guideline aimed at advancing standardized practices for incorporating explainability into multimodal models and streamlining their evaluation across tasks and domains.

\subsection{Search}
The search protocol was developed by the first author (MRK) and reviewed by the other authors (SL and JA). The PICO (Population, Intervention, Comparison, and Outcomes) strategy is followed for the formulation of the search string. We define the population to be attention-based multimodal models for any task from an arbitrary domain, incorporating any combination of numbers and types of modalities. The different explainability or interpretability methods are defined as the intervention. The outcome includes qualitative insights for model architecture, explanations, metrics employed for evaluation of the results, and the criteria used for determining the validity of the explanations. Based on this, we formulate the search term as presented in Table \ref{tab:search}.

\begin{table}[t]
\renewcommand{\arraystretch}{1.3}
    \centering
    \caption{Searches in databases}
    \scriptsize
    \begin{tabular}{c|p{0.8\textwidth}}
    \hline
        Database & Search \\
        \hline
         Scopus & \texttt{( TITLE-ABS-KEY ( multimodal ) OR TITLE-ABS-KEY ( "multi-modal" ) OR TITLE-ABS-KEY ( modality ) OR TITLE-ABS-KEY ( generative ) OR TITLE-ABS-KEY ( vision ) OR TITLE-ABS-KEY ( auditory ) OR TITLE-ABS-KEY ( audio ) OR TITLE-ABS-KEY ( speech ) OR TITLE-ABS-KEY ( text ) OR TITLE-ABS-KEY ( language ) OR TITLE-ABS-KEY ( signal ) OR TITLE-ABS-KEY ( image ) OR TITLE-ABS-KEY ( coding ) ) AND ( TITLE-ABS-KEY ( encoder ) OR TITLE-ABS-KEY ( "encoder-decoder" ) OR TITLE-ABS-KEY ( decoder ) OR TITLE-ABS-KEY ( attention ) ) AND TITLE-ABS-KEY ( transformer ) AND ( ( TITLE ( explain* ) OR TITLE ( interpret* ) OR TITLE ( xai ) ) OR ( ( TITLE ( "tell us" ) OR TITLE ( mean* ) OR TITLE ( represent* ) ) AND ( TITLE-ABS-KEY ( explain* ) OR TITLE-ABS-KEY ( interpret* ) ) ) )} 
     \\
         \hline
         Web of Science & \texttt{(ALL=(multimodal) OR ALL=(multi-modal) OR ALL=(modality) OR ALL=(generative) OR ALL=(vision) OR ALL=(auditory) OR ALL=(audio) OR ALL=(speech)  OR ALL=(text) OR ALL=(language) OR ALL=(signal) OR ALL=(image) OR ALL=(coding) OR ALL=(code)) AND (ALL=(encoder) OR ALL=("encoder-decoder") OR ALL=(decoder) OR ALL=(attention)) AND ALL=(transformer) AND ((TI=(explain*) OR TI=(interpret*) OR TI=(XAI)) OR ((TI=("tell* us") OR TI=(mean*) OR TI=(represent*)) AND (ALL=(explain*) OR ALL=(interpret*))))} \\
         \hline
    \end{tabular}
    \label{tab:search}
\end{table}

\subsubsection{Search Method}
The search and study selection were conducted in accordance with the Preferred Reporting Items for Systematic Reviews and Meta-Analyses (PRISMA) guidelines \cite{moher2009preferred}. We utilized two major academic databases—\textit{Web of Science} and \textit{Scopus}—due to their extensive coverage, relevance to our study objectives, and complementary indexing, which helps mitigate the risk of missing relevant studies. The database-specific search terms, with slight modifications between platforms, are listed in Table \ref{tab:search}.

\subsection{Inclusion and Exclusion Criteria}
We decided to constrain the search to peer-reviewed studies published within the period January $1^{st}$, 2020 to January $31^{st}$, 2024. The initial date is capped as the use of attention-based models in MML has grown substantially in recent times. Besides, the explainability of models has shifted from classic model-agnostic methods to more model-specific ones. The rest of the criteria for eligibility to be considered as part of the review are as follows:
\begin{enumerate}
    \item The study either proposes a multimodal, generative, or multichannel attention-based model as the primary research objective or uses one as any part of the analysis.
    \item The study discusses the explainability of the model to any extent, including qualitative or quantitative analysis.
    \item Original peer-reviewed publications part of conference proceedings, journals, or workshops.
    \item Article published in the English language.
\end{enumerate}

The following criteria are used to exclude studies from the review:
\begin{enumerate}
    \item Secondary studies, including systematic reviews, meta-analyses, and narrative reviews.
    \item Editorials, opinion pieces, and commentary letters.
    \item Books and gray literature.
    \item Studies that are duplicates of other studies.
    \item Studies that use ablation studies or other ways to validate their model without explicitly applying established explainability or interpretability techniques.
\end{enumerate}

\subsection{Study Selection Procedure}
The study selection is done in the standard multi-stage approach. We first remove the duplicate studies from the final set of search results from the two databases. Next, the titles and abstracts of the studies are screened, and the eligibility criteria are applied to filter the results. Then, the full-text screening is conducted to further assess the eligibility of the studies. Finally, we perform forward and backward snowball sampling from the set of studies selected after the full-text screening.

During the title-abstract screening, we classify the studies into three categories: `include', `exclude', or `ambiguous'. Any study without complete certainty is labeled as ambiguous to be more inclusive and is resolved during subsequent steps. To further improve the reliability of the review and confidently resolve the ambiguous cases, we used an additional human-guided large language model (LLM)-based screening process as described in the following section.

\subsubsection{LLM-based Screening}
LLMs are increasingly being utilized at various stages of the scientific reading and writing process \cite{altmae2023artificial, dagdelen2024structured}, including during the screening phase of SLRs \cite{felizardo2024chatgpt}. Recent studies suggest that LLMs perform at a level comparable to human screeners, for example, in terms of classification accuracy when replicating the title-abstract screening of a prior SLR on \textit{time pressure in software engineering} \cite{huotala2024promise}. Based on this, we adopt a human-guided, LLM-assisted method to independently screen all studies categorized as either `include' or `ambiguous'. Full technical details, including model specifications, prompts, and example outputs, are provided in \textit{Supplementary Material}.

For this task, we employ a quantized version of the LLaMA 3.1 model, trained on data with a knowledge cutoff of December 2023 \cite{grattafiori2024llama}. The model is provided with instructions, eligibility criteria, and the title and abstract of each article. It is then prompted to follow the same three decision categories used by the authors during screening.



In addition to assessing each criterion, the LLM is instructed to extract relevant data to justify its screening decision. This simulates a chain-of-thought (CoT) reasoning process, which has been shown to improve LLM performance on complex tasks \cite{wei2022chain}. The generated explanations are then manually reviewed by the first author to verify the categorization.

It is important to note that the LLM outputs are not used for direct inclusion or exclusion decisions. Instead, they are incorporated during the full-text assessment stage to provide additional input for the authors. Table \ref{tab:llm} presents the screening results from the title–abstract phase (by both authors and LLM), as well as the final decisions based on full-text assessments.

\begin{table}[]
\renewcommand{\arraystretch}{1.3}
    \caption{LLM-assisted Screening Results \\
    \textbf{Abbreviations:} T/A = Title and abstract screening; Full-text = Full-text article assessment. \\
    \textbf{Legend:} $\times$ = Irrelevant / Any case possible.}
    \label{tab:llm}
    \scriptsize
    \centering
    \begin{tabular}{c|>{\centering\arraybackslash}p{0.14\linewidth}|>{\centering\arraybackslash}p{0.15\linewidth}|c|c}
    \hline
    Authors (T/A) & LLM (T/A) & Final (Full-text)  & Count & Percentage \\
    \hline
    Include/Ambiguous & Include & Include & 47 & 72.3\% \\
    Exclude/Ambiguous & Exclude & Exclude & 3 & 4.62\% \\
    Include/Ambiguous & Ambiguous & Include & 4 & 6.15\% \\
    Exclude/Ambiguous & Ambiguous & Exclude & 2 & 3.08\% \\
    $\times$ & Include & Exclude & 7 & 10.77\% \\
    $\times$ & Exclude & Include & 2 & 3.08\% \\
    \hline
    \multicolumn{2}{l}{Total} &&  65 & 100\% \\
    \hline
    \end{tabular}
\end{table}

As shown in Table \ref{tab:llm}, LLM-generated inputs are largely reliable. The LLM-assisted method provided usable screening suggestions in 90.77\% of cases, with the remaining labeled as ambiguous. Of all predictions, only 13.85\% were found to be incorrect after the full-text assessment. These results suggest that LLM-assisted screening is a promising approach, and we recommend further exploration and systematic adoption of such methods in the literature review process.

\subsubsection{Snowball sampling}
To enhance the theoretical validity and overall coverage of the study, we performed both forward and backward snowball sampling following the full-text screening phase, in line with established guidelines \cite{wohlin2014guidelines}. The sampling process was initiated from a seed study by Chefer \emph{et al.} \cite{chefer_generic_2021}, selected for its strong relevance to the review and its placement early in the study period. Given the high number of citations to and from this work, we applied filtering constraints based on citation counts to manage the volume of candidate studies. The sampling was conducted on December $17^{th}$, 2024, using only the Web of Science database, chosen for its more accessible filtering capabilities. Details of the sampling procedure, including inclusion counts at each step, are provided in Table \ref{tab:snow}.

\begin{table}[!t]
\renewcommand{\arraystretch}{1.3}
    \caption{Parameters for Snowball Sampling}
    \label{tab:snow}
    \centering
    \begin{tabular}{{p{0.12\linewidth}|p{0.12\linewidth}|p{0.1\linewidth}|p{0.12\linewidth}|p{0.12\linewidth}|p{0.1\linewidth}|p{0.1\linewidth}}}
    \hline
    Sampling direction & Population size & After exclusion criteria  & Percentage considered & Top cited candidates & No. of duplicates & After full-text evaluation \\
    \hline
         Forward & 94 & 62 & 15\% & 9 & 2 & 0 \\
    \hline
         Backward & 50 & 17 & 30\% & 5 & 0 & 1 \\
    \hline
    \end{tabular}

\end{table}
    
\section{Qualitative Synthesis Overview\label{Qualitative Synthesis Overview}}
\begin{figure}[]
    \centering
    \includegraphics[width=1\linewidth, trim={0 8cm 0 2cm},clip]{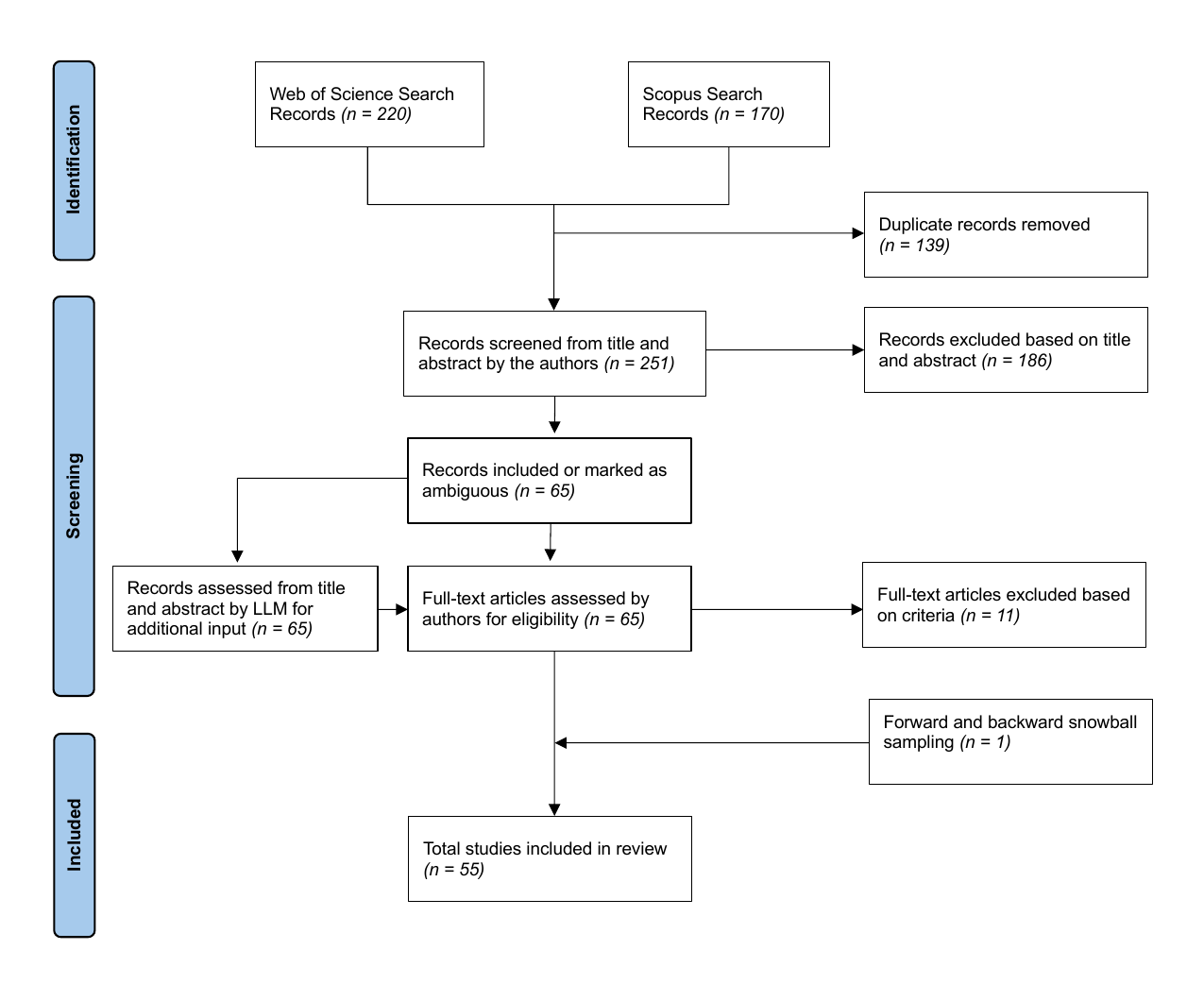}
    \caption{PRISMA flowchart for the number of selected studies during the different steps of the selection process.}
    \label{fig:prisma}
\end{figure}

After a rigorous selection process, the final approved set contained a total of 55 publications. A summary of the selection procedure is presented in Figure \ref{fig:prisma}. The key characteristics of the included explainability-focused studies are summarized in Table~\ref{tab:compact_papers}, while the studies focusing on explainability interfaces are reported in Table~\ref{tab:xai_interfaces}. The complete review dataset, including all computational procedures and results, is available in the supplementary material and online at: \url{https://decoding-the-multimodal-maze.notion.site/}. Data have been extracted from these studies for the analysis of how attention-driven multimodal models are adopted and how explainability is explored in different tasks in a diverse range of application domains. In addition, an overview of the important bibliometric analytics for the final set is presented in this section.

\begin{table}[]
\centering
\scriptsize
\caption{Overview of the included studies, year, primary task, and analysis dimensions. Modalities: G (graph), I (image/vision), A (audio), L (language), T (tabular), V (video), C (code). Fusion Architectures: EF (early), CF (single cross-attention), MC (multi-cross), HIER (hierarchical), SS (single-stream), MS (modular). XAI used: ANTE (ante-hoc), ATT (attention-based), GRAD (gradient-based), COMPOSITE (attention composite), PROBE (probing/ablation), HYB (hybrid), SELF (self-explaining), M-AGN (model-agnostic). Evaluation criteria: FA (faithfulness), LO (localization), RO (robustness), CO (complexity), RA (randomization), AX (axiomatic), QL (qualitative), QN (quantitative).}
\setlength{\tabcolsep}{6pt}
\begin{tabular}{l l c l c c c c c c}
\hline
\textbf{Ref} & \textbf{Year} & \textbf{Task} & \textbf{Mod} & \textbf{Fusion} & \textbf{XAI} & \textbf{Eval} & \textbf{Public Data} \\
\hline
\cite{janssens_360_2024} & 2024 & Classification & \datatypebar{0}{0}{0}{1}{1}{0}{0} & HIER & M-AGN & \evalbar{1}{0}{1}{0}{0}{0}{0}{0} & Y \\
\cite{kumar_bert_2022} & 2022 & Classification & \datatypebar{0}{0}{0}{1}{0}{0}{0} & MS & ATT & \evalbar{0}{0}{0}{0}{0}{0}{1}{0} & Y \\
\cite{du_case-based_2023} & 2023 & Classification & \datatypebar{0}{0}{0}{0}{1}{0}{0} & MS & COMP & \evalbar{0}{0}{0}{0}{0}{0}{1}{0} & Y \\
\cite{ukwuoma_hybrid_2023} & 2023 & Classification & \datatypebar{0}{1}{0}{0}{0}{0}{0} & EF & HYB & \evalbar{0}{0}{0}{0}{0}{0}{1}{0} & Y \\
\cite{wang_novel_2023} & 2023 & Classification & \datatypebar{0}{0}{0}{0}{1}{0}{0} & MC & ATT & \evalbar{0}{0}{0}{0}{0}{0}{1}{0} & Y \\
\cite{bhargava_adaptive_2020} & 2020 & MML & \datatypebar{0}{1}{0}{1}{0}{0}{0} & MC & PROBE & \evalbar{0}{0}{0}{0}{0}{0}{1}{0} & Y \\
\cite{guo_explainable_2023} & 2023 & NLE & \datatypebar{0}{0}{0}{1}{1}{0}{0} & EF & SELF & \evalbar{1}{0}{0}{0}{0}{0}{0}{0} & Y \\
\cite{yang_explainable_2023} & 2023 & Classification & \datatypebar{0}{1}{0}{0}{0}{0}{0} & HIER & GRAD & \evalbar{0}{0}{0}{0}{0}{0}{1}{0} & Y \\
\cite{qiang_attcat_nodate} & 2022 & XAI, NLP & \datatypebar{0}{0}{0}{1}{0}{0}{0} & SS & COMP & \evalbar{1}{0}{0}{0}{0}{0}{0}{0} & Y \\
\cite{koyama_attention_2023} & 2023 & Classification & \datatypebar{0}{0}{0}{1}{1}{0}{0} & MC & COMP & \evalbar{0}{0}{0}{0}{0}{0}{1}{0} & Y \\
\cite{ferrando_attention_2021} & 2021 & NLP, NMT & \datatypebar{0}{0}{0}{1}{0}{0}{0} & CF & COMP & \evalbar{0}{1}{0}{0}{0}{0}{1}{0} & Y \\
\cite{rigotti_attention-based_2022} & 2022 & XAI, Classification & \datatypebar{0}{1}{0}{0}{0}{0}{0} & CF & ANTE & \evalbar{1}{0}{0}{0}{0}{0}{0}{0} & Y \\
\cite{abdulkadir_augmenting_2022} & 2022 & Classification & \datatypebar{0}{1}{0}{0}{1}{0}{0} & EF & ANTE & \evalbar{0}{0}{0}{0}{0}{0}{1}{0} & Y \\
\cite{meng_bidirectional_2021} & 2021 & Classification & \datatypebar{0}{0}{0}{0}{1}{0}{0} & EF & ATT & \evalbar{0}{0}{0}{0}{0}{0}{1}{0} & Y \\
\cite{ding_deepstf_2023} & 2023 & Classification & \datatypebar{0}{0}{0}{0}{1}{0}{0} & HIER & ATT & \evalbar{0}{0}{0}{0}{0}{0}{1}{0} & Y \\
\cite{feucht_description-based_2021} & 2021 & Classification & \datatypebar{0}{0}{0}{1}{1}{0}{0} & HIER & ATT & \evalbar{0}{0}{0}{0}{0}{0}{0}{0} & Y \\
\cite{sun_dfyolov5m-m2transformer_2023} & 2023 & Multi-task & \datatypebar{0}{1}{0}{1}{0}{0}{0} & CF & SELF & \evalbar{1}{0}{0}{0}{0}{0}{0}{0} & N \\
\cite{yu_ex-vit_2023} & 2023 & Segmentation & \datatypebar{0}{1}{0}{0}{0}{0}{0} & MS & ANTE & \evalbar{0}{1}{0}{0}{0}{0}{0}{0} & Y \\
\cite{buoy_explainable_2023} & 2023 & Scene Text & \datatypebar{0}{1}{0}{0}{1}{0}{0} & SS & ANTE & \evalbar{0}{1}{0}{0}{0}{0}{0}{0} & Y \\
\cite{chiewhawan_explainable_2020} & 2020 & Regression & \datatypebar{0}{0}{0}{1}{1}{0}{0} & EF & GRAD & \evalbar{0}{0}{0}{0}{0}{0}{1}{0} & Y \\
\cite{ullah_explainable_2022} & 2022 & Classification & \datatypebar{0}{1}{0}{1}{0}{0}{0} & HIER & M-AGN & \evalbar{0}{0}{0}{0}{0}{0}{1}{0} & Y \\
\cite{zhang_explainable_2022} & 2022 & Regression & \datatypebar{0}{1}{0}{0}{1}{0}{0} & HIER & COMP & \evalbar{0}{0}{0}{0}{0}{0}{1}{0} & Y \\
\cite{mohammadkhani_explaining_2023} & 2023 & Code & \datatypebar{1}{0}{0}{0}{0}{0}{1} & CF & ATT & \evalbar{0}{0}{0}{0}{0}{0}{1}{0} & Y \\
\cite{wang_exploring_2021} & 2021 & NLP & \datatypebar{0}{0}{0}{1}{0}{0}{0} & SS & ATT & \evalbar{1}{0}{0}{0}{0}{0}{0}{1} & Y \\
\cite{chen_faster_2023} & 2023 & VL & \datatypebar{0}{1}{0}{1}{0}{0}{0} & EF & SELF & \evalbar{1}{0}{0}{0}{0}{0}{0}{0} & Y \\
\cite{huang_generic_2023} & 2023 & MML, XAI & \datatypebar{0}{1}{0}{1}{0}{0}{0} & MC & COMP & \evalbar{1}{0}{0}{0}{0}{0}{0}{0} & Y \\
\cite{chefer_generic_2021} & 2021 & MML, XAI & \datatypebar{0}{1}{0}{1}{0}{0}{0} & EF/CF & COMP & \evalbar{1}{1}{0}{0}{0}{0}{0}{0} & Y \\
\cite{wu_interpretable_2023} & 2023 & Regression & \datatypebar{0}{0}{0}{1}{1}{0}{0} & EF & ANTE & \evalbar{0}{0}{0}{0}{0}{0}{1}{0} & Y \\
\cite{parelli_interpretable_2023} & 2023 & VL & \datatypebar{0}{1}{0}{1}{0}{0}{0} & CF & SELF & \evalbar{0}{0}{0}{0}{0}{0}{1}{0} & Y \\
\cite{malkiel_interpreting_2022} & 2022 & NLP & \datatypebar{0}{0}{0}{1}{0}{0}{0} & MS & COMP & \evalbar{0}{0}{0}{0}{0}{0}{1}{1} & Y \\
\cite{boito_investigating_2020} & 2020 & NLP, NMT & \datatypebar{0}{0}{1}{1}{0}{0}{0} & CF & ATT & \evalbar{0}{0}{0}{1}{0}{0}{1}{0} & Y \\
\cite{treviso_ist-unbabel_2021} & 2021 & Regression & \datatypebar{0}{0}{0}{1}{0}{0}{0} & EF & HYB & \evalbar{1}{0}{0}{0}{0}{0}{0}{0} & Y \\
\cite{zanzotto_kermit_2020} & 2020 & NLP & \datatypebar{1}{0}{0}{1}{0}{0}{0} & HIER & HYB & \evalbar{0}{0}{0}{0}{0}{0}{1}{0} & Y \\
\cite{xu_logiformer_2022} & 2022 & NLP, reasoning & \datatypebar{1}{0}{0}{1}{0}{0}{0} & HIER & ATT & \evalbar{0}{0}{0}{0}{0}{0}{1}{0} & Y \\
\cite{xu_multi-granular_2020} & 2020 & NLP & \datatypebar{0}{0}{0}{1}{0}{0}{0} & EF & ATT & \evalbar{0}{0}{0}{0}{0}{0}{1}{0} & Y \\
\cite{che_multiscale_2023} & 2023 & Classification & \datatypebar{0}{0}{0}{0}{1}{0}{0} & CF & ATT & \evalbar{0}{0}{0}{0}{0}{0}{1}{0} & Y \\
\cite{heo_natural-language-driven_2023} & 2023 & MML & \datatypebar{0}{0}{1}{1}{0}{1}{0} & EF & SELF & \evalbar{1}{0}{0}{0}{0}{0}{0}{1} & Y \\
\cite{sun_neural_2021} & 2021 & Multi-task & \datatypebar{0}{1}{0}{1}{0}{0}{0} & SS & PROBE & \evalbar{0}{0}{0}{0}{0}{0}{1}{0} & Y \\
\cite{wang_odp-transformer_2023} & 2023 & Multi-task & \datatypebar{1}{1}{0}{1}{0}{0}{0} & CF & SELF & \evalbar{1}{0}{0}{0}{0}{0}{1}{0} & Y \\
\cite{li_oscar_2020} & 2020 & VL & \datatypebar{0}{1}{0}{1}{0}{0}{0} & EF & ATT & \evalbar{1}{0}{0}{0}{0}{0}{1}{0} & N \\
\cite{kandukuri_physical_2022} & 2022 & Multi-task & \datatypebar{0}{0}{0}{0}{1}{1}{0} & SS & ANTE & \evalbar{1}{0}{0}{0}{0}{0}{0}{0} & N \\
\cite{huang_representation_2023} & 2023 & Classification & \datatypebar{0}{0}{0}{0}{1}{0}{0} & MC & ATT & \evalbar{0}{0}{0}{0}{0}{0}{1}{0} & Y \\
\cite{jha_supervised_2023} & 2023 & NLP & \datatypebar{0}{0}{0}{1}{0}{0}{0} & HIER & ATT & \evalbar{0}{0}{0}{0}{0}{0}{1}{0} & Y \\
\cite{wang_tfregnci_2023} & 2023 & Regression & \datatypebar{0}{1}{0}{0}{1}{0}{0} & EF & HYB & \evalbar{0}{0}{0}{0}{0}{0}{1}{0} & Y \\
\cite{ferrando_towards_2022} & 2022 & NLP & \datatypebar{0}{0}{0}{1}{0}{0}{0} & CF & ATT & \evalbar{0}{0}{0}{0}{0}{0}{1}{0} & Y \\
\cite{kumar_towards_2021} & 2021 & Classification & \datatypebar{0}{0}{1}{1}{0}{0}{0} & HIER & ATT & \evalbar{1}{0}{0}{0}{0}{0}{1}{0} & Y \\
\cite{xiao_transformer_2024} & 2024 & Regression & \datatypebar{1}{1}{0}{0}{0}{0}{0} & MS & GRAD & \evalbar{0}{0}{0}{0}{0}{0}{1}{0} & Y \\
\cite{hiemstra_using_2021} & 2021 & NLP, NLE & \datatypebar{0}{0}{0}{1}{0}{0}{0} & MS & PROBE & \evalbar{1}{0}{0}{0}{0}{0}{1}{0} & Y \\
\cite{naseem_vision-language_2023} & 2023 & VL & \datatypebar{0}{1}{0}{1}{0}{0}{0} & EF & HYB & \evalbar{0}{0}{0}{0}{0}{0}{1}{0} & Y \\
\cite{ilinykh_what_2021} & 2021 & VL, NLE & \datatypebar{0}{1}{0}{1}{0}{0}{0} & CF & ATT & \evalbar{0}{0}{0}{1}{0}{0}{1}{0} & Y \\
\cite{dong_why_2023} & 2023 & VL & \datatypebar{0}{1}{0}{1}{0}{0}{0} & CF & ATT & \evalbar{1}{0}{0}{0}{0}{0}{1}{0} & Y \\
\cite{lin_zero-shot_2023} & 2023 & VL & \datatypebar{0}{1}{0}{0}{0}{0}{0} & CF & HYB & \evalbar{1}{0}{0}{0}{0}{0}{1}{0} & Y \\
\hline
\end{tabular}
\label{tab:compact_papers}
\end{table}

\begin{table}[]
\centering
\scriptsize
\caption{Attribute-level summary of the three included interface-focused explainability tools.}
\renewcommand{\arraystretch}{1.2}
\setlength{\tabcolsep}{6pt}
\begin{tabular}{lp{60pt}cp{50pt}p{60pt}p{55pt}}
\hline
\textbf{Tool / Paper} & \textbf{Model Type} & \textbf{Modality} & \textbf{Explanation Level} & \textbf{Visualization Type} & \textbf{Validation} \\
\hline
\textbf{Inseq} \cite{sarti_inseq_2023} & Sequence generation (encoder–decoder, GPT-2) & Language & Token / Layer & Aggregated attribution maps & MT gender bias, GPT-2 facts  \\
\textbf{VISIT} \cite{katz_visit_2023} & Transformer (GPT-style) & Language & Attention / Hidden state & Forward-flow graphs & IOI, layer-norm variants \\
\textbf{VL-InterpreT} \cite{aflalo_vl-interpret_2022} & Multimodal transformer (KD-VLP) & VL & Cross- / Intra-modal attention & Multi-head attention maps, embedding plots & VCR, WebQA on KD-VLP  \\
\hline
\end{tabular}
\label{tab:xai_interfaces}
\end{table}

\begin{figure}
    \centering
    \includegraphics[width=0.7\linewidth]{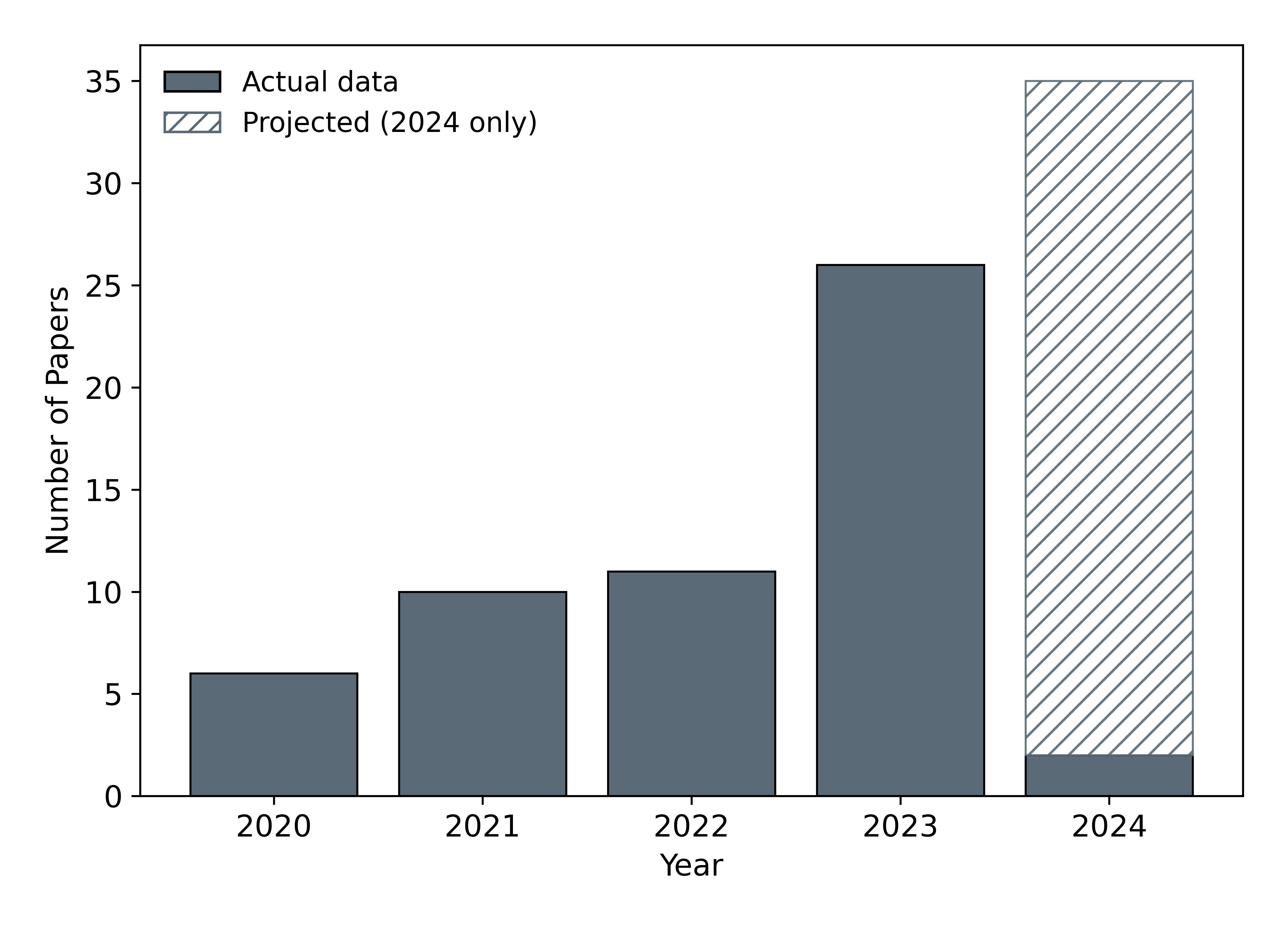}
    \caption{Publication per year}
    \label{fig:years}
\end{figure}

Primarily, the evolution of studies focused on the field over the selected period is presented in Figure \ref{fig:years}. There has been a rapid surge of studies in the last two years, which coincides with the significant development in attention-based multimodal models over the past years and general growth in works building on previously introduced explanation methods \cite{fantozzi2024explainability}. This increased interest can also be influenced by several other factors, such as stricter regulations on AI \cite{nannini2023explainability, chun2024comparative} and rapid mainstream adoption of AI-based agents (e.g., LLM-agents).

\begin{figure}
     \centering
     \begin{subfigure}[b]{\linewidth}
         \centering
         \includegraphics[width=\textwidth, trim={0 4cm 0 3cm},clip]{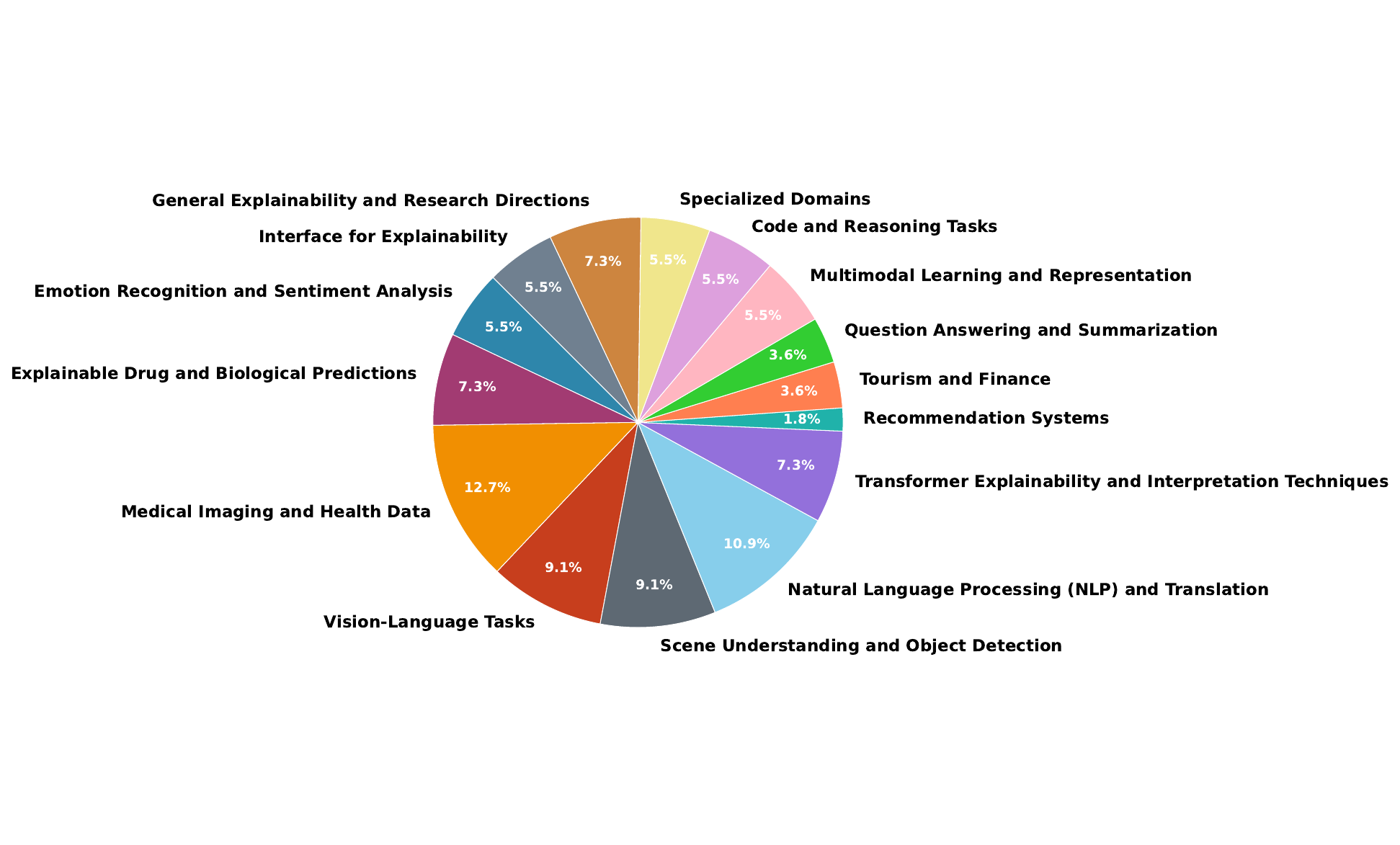}
         \caption{Thematic Distribution}
         \label{fig:topics}
     \end{subfigure}
     \hfill
     \begin{subfigure}[b]{\linewidth}
         \centering
         \includegraphics[width=\textwidth]{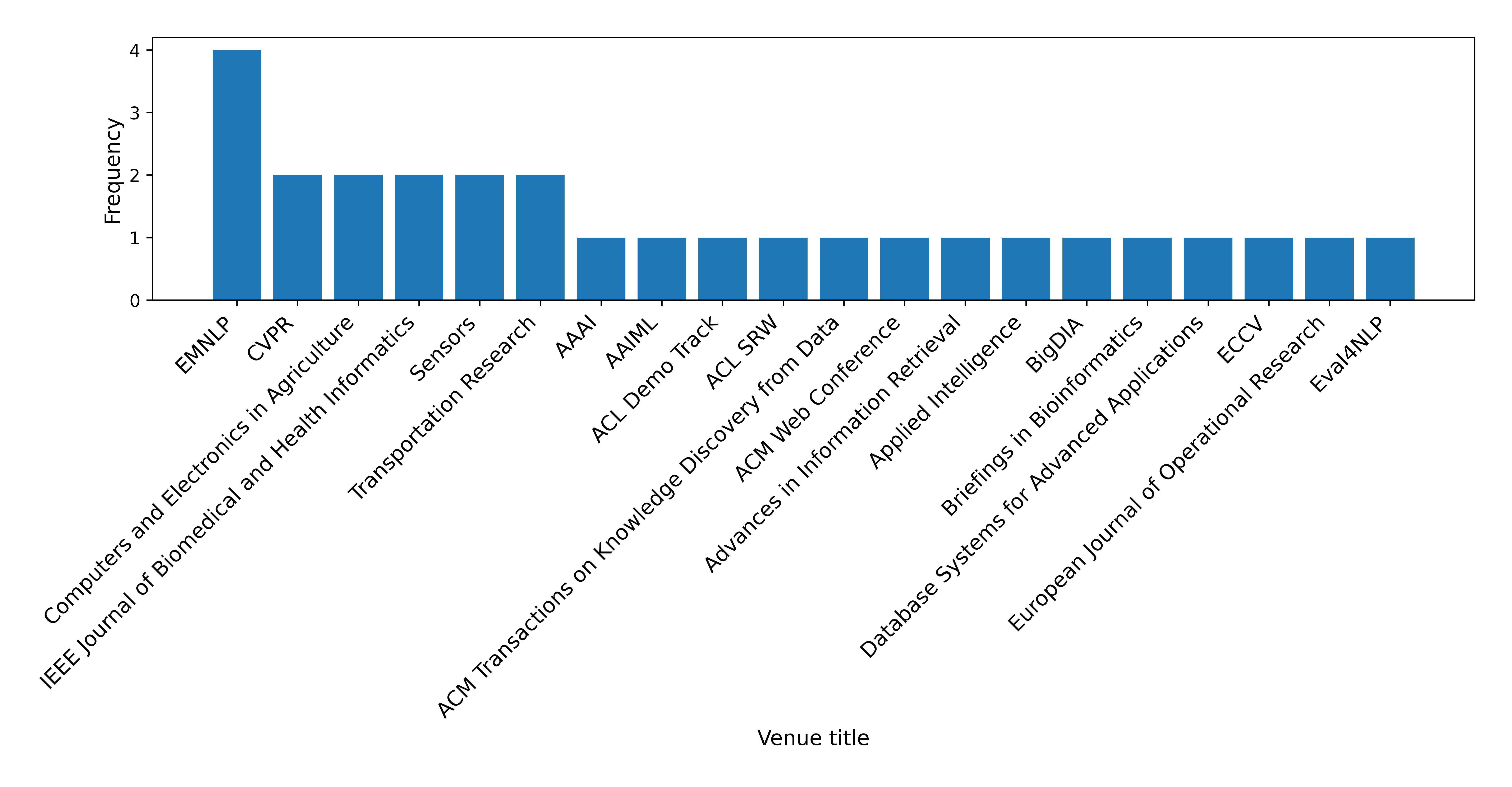}
         \caption{Top 20 Publication Venues}
         \label{fig:venues}
     \end{subfigure}
     \hfill
     \begin{subfigure}[b]{\linewidth}
         \centering
         \includegraphics[width=0.7\textwidth, trim={0 2cm 0 2cm}, clip]{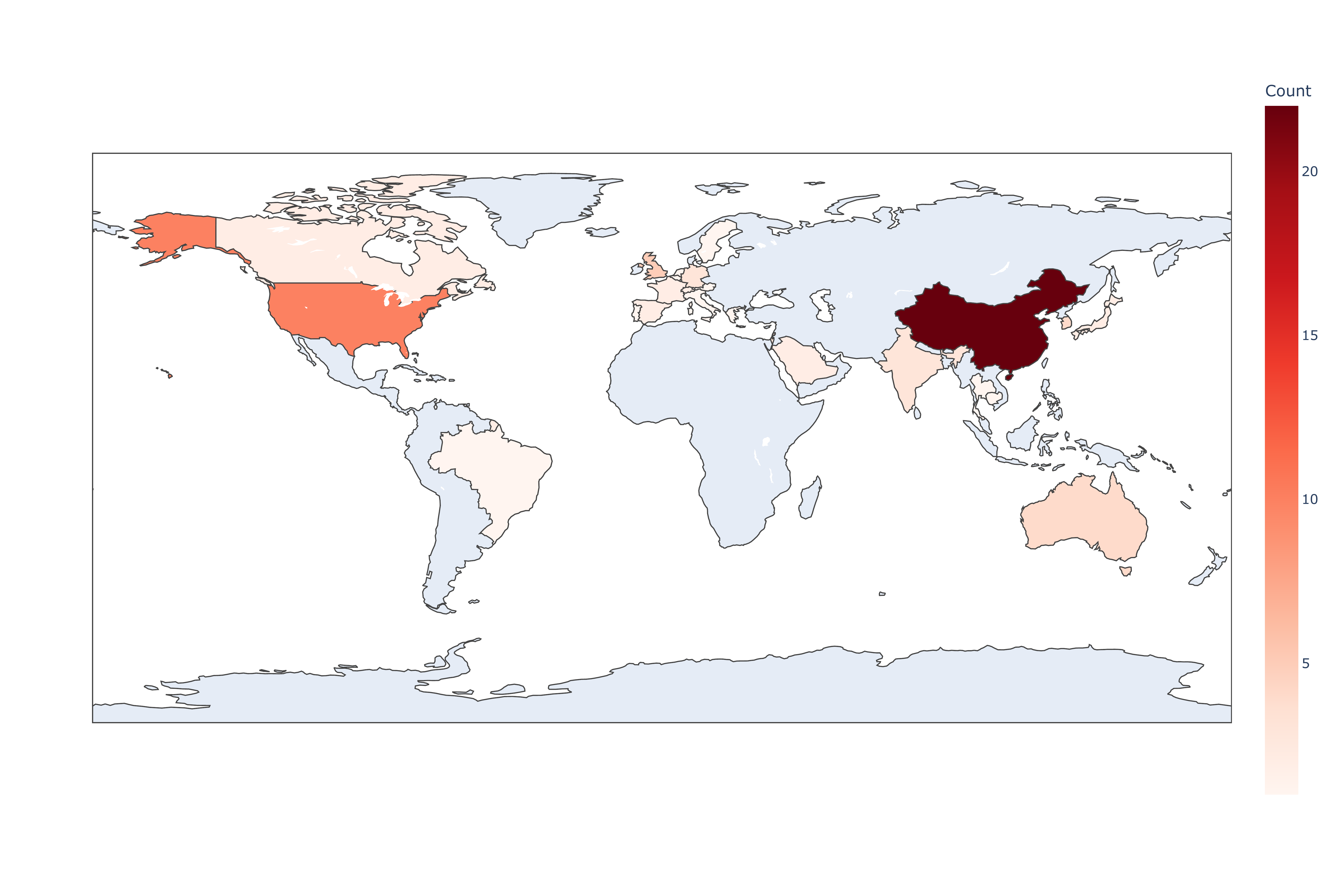}
         \caption{Choropleth Map of the Publications}
         \label{fig:geo}
     \end{subfigure}
        \caption{Key Bibliometric Analytics of the Publications}
        \label{fig:bib_stat}
\end{figure}

Additional analyses on thematic categorization, geographical distribution, and top publication venues are presented in Figure \ref{fig:bib_stat}. For a structured thematic analysis, studies were grouped into themes defined by application domains and tasks. Although each study was assigned to a single theme to streamline the analysis, it is important to note that both the themes and assigned studies are not strictly mutually exclusive and may span multiple disciplines (e.g., \textit{Natural Language Processing (NLP) and Translation} vs. \textit{Question Answering and Summarization}). The thematic distribution, shown in Figure \ref{fig:topics}, highlights the diversity of research interests across domains that demand explainability \cite{burkart2021survey}.

Likewise, the publication venues are varied, encompassing major conferences and journals in AI and computer vision. The most represented venue is EMNLP, the flagship conference in NLP (see Figure \ref{fig:venues}). The geographical origin of the research is illustrated in the choropleth map in Figure \ref{fig:geo}, revealing that the majority of publications originate from institutions based in the People's Republic of China, followed by the United States. Beyond these two countries, the distribution reflects a broad and globally diverse research landscape.

\section{Modalities\label{Modalities}}
Due to the wide application area considered for the generality of our study, the publications for this review reflect a diverse combination of modalities. We follow the data modality characterization by Nauta \emph{et al.} \cite{nauta2023anecdotal} used with slight exceptions for labeling the combination of modalities. The time-series data are labeled to the same category as tabular/structured due to similar representation used for models. In addition, we make a distinct category for representing programming code, due to the difference in syntax and semantics compared to natural language. Finally, there is a separate category for audio/speech data. We adopt a flexible approach to modality labeling, recording a modality for a given publication whenever the input, output, or any intermediate representation relevant to explainability aligns with our predefined categories.

\begin{figure}
     \centering
     \begin{subfigure}[b]{0.48\linewidth}
        \centering
        \includegraphics[width=\linewidth]{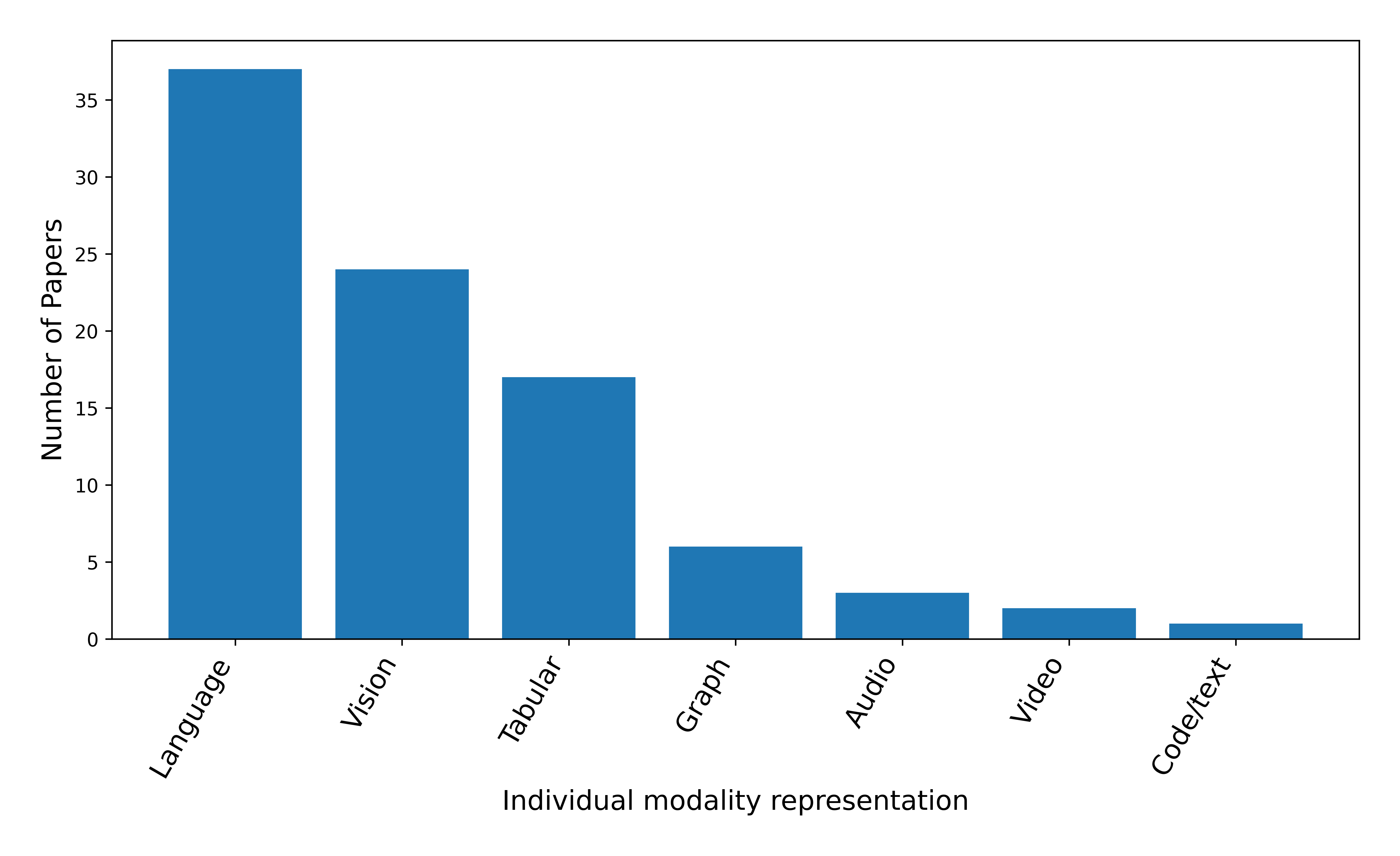}
        \caption{Distribution of papers represented as individual modalities}
        \label{fig:ind_mod}
     \end{subfigure}
     \hfill
     \begin{subfigure}[b]{0.48\linewidth}
        \centering
        \includegraphics[width=1\linewidth]{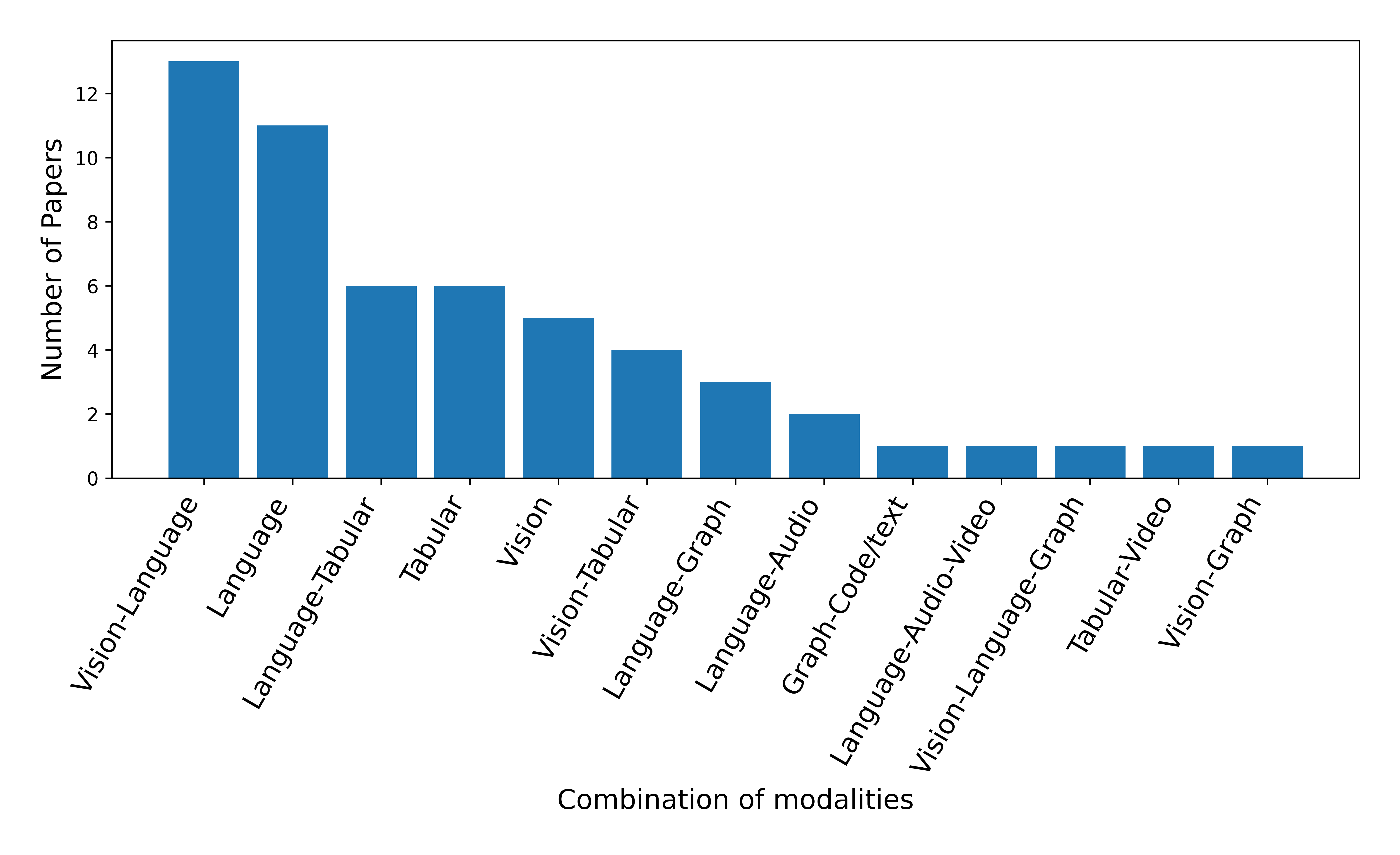}
        \caption{Distribution of papers by combination of modalities}
        \label{fig:cmb_mod}
     \end{subfigure}
    \caption{Representation of modalities}
    \label{fig:modalities}
\end{figure}

The different combinations of modalities observed in the review are shown in Figure \ref{fig:cmb_mod}. These combinations are an aggregation of all types of data from the input and output space. The most common tasks in the area of multimodal explainability are targeted at VL and NLP tasks. As an individual modality of data represented either as input or output, language is quite significantly represented, with vision being the second. The least amount of work is targeted towards code modeling. These findings are aligned with other surveys on XAI (e.g., \cite{nauta2023anecdotal}).

\begin{figure}
    \centering
    \includegraphics[width=0.75\linewidth]{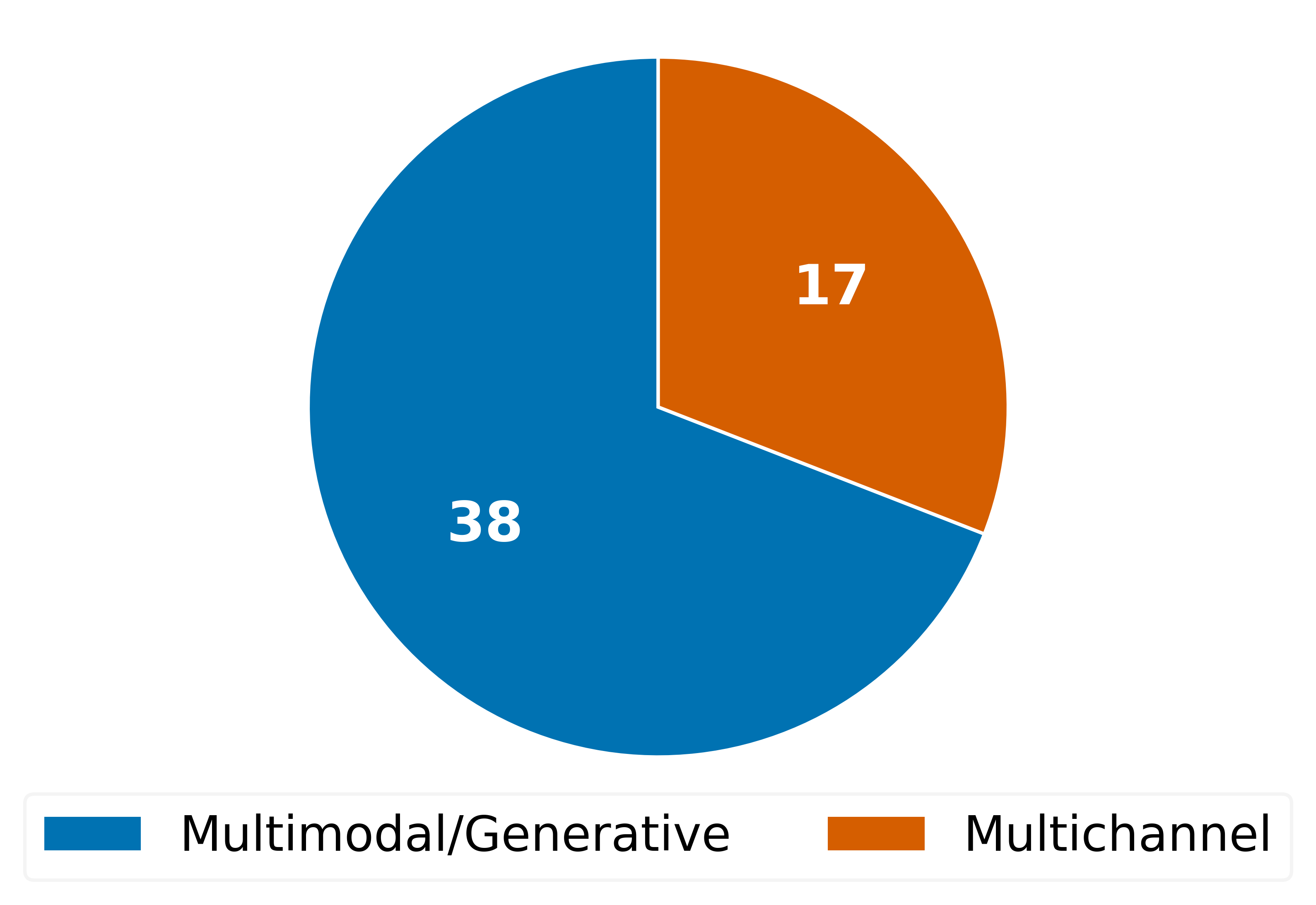}
    \caption{Distribution of multimodal/generative and multichannel modeling approaches}
    \label{fig:cvm}
\end{figure}

An important decision regarding the eligibility criteria in our study is the inclusion of ``multichannel" modeling approaches. These criteria cover models that make decisions based on multiple inputs generated by processing the same source input. For instance, Yang \emph{et al.} introduce a remote scene classification model that globally models multiscale representations from both spatial and frequency domains of an input remote scene image \cite{yang_explainable_2023}, hence making the approach multichannel and eligible for our study. The motivation behind the inclusion is that explaining multichannel approaches requires generating explanations from different representations and from different portions of the model. The other categories include multimodal models, which process and generate outputs using multiple modalities, and generative models, which produce sequences, typically within the same modality as the input. We represent these two categories by grouping them as one. We do not explicitly include foundation models—large, general purpose models—as our review targets explainability techniques rather than model scale or generality, and their inclusion could add complexity that obscures insights specific to multimodal and generative explainability. The distribution of multimodal or generative and multichannel approaches is presented in Figure \ref{fig:cvm}. Almost $69\%$ of the publications in our dataset belong to the category of multimodal/generative models and $31\%$ of them use a multichannel approach before attempting to explain model decisions.

\section{Tasks and Datasets\label{Tasks and Datasets}}
\begin{figure}
    \centering
    \includegraphics[width=0.9\linewidth]{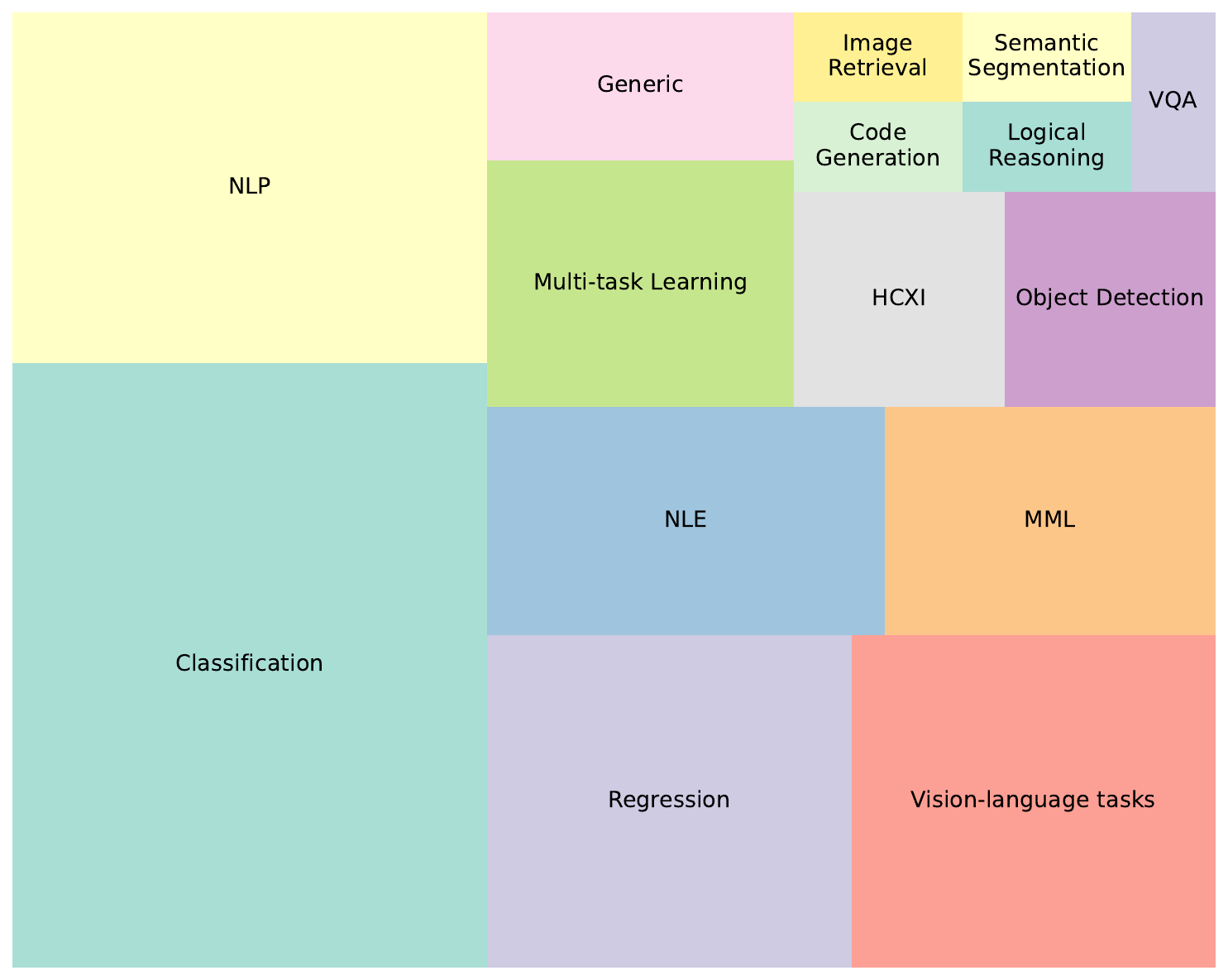}
    \caption{Primary training task objectives}
    \label{fig:task}
\end{figure}

Explainability in the set of publications is often explored for one or more primary models. These models are trained with primary task objectives that determine the datasets applicable (also determined by the domain) and how these are trained, including the architecture. The general categorization of the models used in the publications is presented in Figure \ref{fig:task}. Studies in the area generally address more than one research objective. Among the observed task groups, the most common one is classification problems utilizing multimodality. Next to that, studies for which the primary objectives can be labeled as NLP, regression, or VL tasks are also quite common. All these groups incorporate a very diverse range of tasks.

The reviewed publications utilize a wide range of datasets, depending on the specific task and research domain. In general, datasets may be used for training models, validating and testing their performance, or both. Since our primary focus is on explainability, when authors specify different datasets for model performance evaluation and explainability evaluation, we report only the latter. Given the diversity of evaluation approaches (see Section \ref{Evaluation Criteria}), datasets are often not systematically used for assessing explainability. Therefore, in this work, we organize our discussion of datasets based on their source tasks and included modalities, with a summary provided in Table \ref{tab:ds}.

Due to the focus on explainability, a lot of studies involve tasks oriented towards explainability. NLP datasets for sentiment analysis, like Amazon polarity, SST-2, and YELP, or language inference datasets for establishing hypothesis-premise relations, like MNLI, can be leveraged for explainability-oriented studies \cite{qiang_attcat_nodate}. Benchmark datasets for the visual question answering (VQA) task, such as the VQA dataset, consist of images paired with questions about their content and the corresponding answers. A later version of the dataset, titled VQAv2, included improvements to reduce language bias by including a pair of positive and negative images for each question. These datasets have also been adopted in explainability-oriented tasks in addition to being used for MML \cite{huang_generic_2023, chefer_generic_2021}.

Natural language explanations (NLE) refer to works on self-explaining models that can be evaluated against ground-truth labels from the datasets. For example, Guo \emph{et al.} used sentiment analysis datasets such as YELP and Amazon-movie for personalized recommendation generation \cite{guo_explainable_2023}. More complicated question-answering datasets, such as multiple option choice-based OpenBookQA or context-driven QA for retrieval problems-ReQA (SQuAD and Natural Questions), can be leveraged for NLE targeted tasks \cite{hiemstra_using_2021}. 

NLP-oriented tasks are quite diverse both in general and in the datasets used for explainability evaluation. In addition to some of the previously mention studies, annotated news corpus for abstractive summarization such as the CNN/DailyMail dataset or NYT50 by H. Wang \emph{et al.} \cite{wang_exploring_2021}, sentiment analysis datasets like Fashion and Wine reviews by Malkiel \emph{et al.} \cite{malkiel_interpreting_2022}, or word alignment datasets with curated gold-standard alignment named Gold Alignment Dataset by Ferrando and Costa-jussà \cite{ferrando_attention_2021} are examples of the NLP-task group. Datasets such as the autonomous vehicle data, Lyft, or MLQE-PE for machine translation, are purely oriented for regression-driven studies \cite{zhang_explainable_2022, treviso_ist-unbabel_2021}. Datasets used for classification problems in this context often also involve solving other complex tasks such as object detection \cite{wang_odp-transformer_2023} or regression \cite{sun_neural_2021}.

Large-scale datasets like MS COCO and PASCAL VOC are extended for evaluating explainability for semantic segmentation works \cite{yu_ex-vit_2023, chefer_generic_2021, huang_generic_2023}. These datasets are particularly valuable due to their vast sizes and the semantic annotations that can be leveraged for evaluations. VCR dataset for visual commonsense reasoning specifies a very sophisticated objective of cognition-level visual understanding. This dataset is used in two different studies \cite{parelli_interpretable_2023, aflalo_vl-interpret_2022}. 

A more unique category of publications on human computer interfaces for explainability (HCXI) study interfaces for explainability and validate them via case studies (discussed in Section \ref{Explanation Interfaces}). For that purpose, Counterfact, a dataset generally designed for assessing whether, in LLMs, factual associations can be modified without affecting other facts, is used by Katz and Belinkov \cite{katz_visit_2023}, and WebQA, a multi-hop, vision-language QA benchmark, is used by Aflalo \emph{et al.} \cite{aflalo_vl-interpret_2022}.

All of the listed datasets are publicly accessible either directly or through some approval process. The only private datasets are the study-specific ones manually collected by the authors and are used for various tasks such as VL \cite{li_oscar_2020, kandukuri_physical_2022} and object detection with explanation \cite{sun_dfyolov5m-m2transformer_2023}.

\begin{table}[htbp]
    \centering
    \caption{Evaluation datasets used for explainability}
    \label{tab:ds}
    \scriptsize
    \renewcommand{\arraystretch}{0.8}
    \begin{tabular}{>{\raggedright\arraybackslash}p{0.22\linewidth}
                    >{\raggedright\arraybackslash}p{0.15\linewidth}
                    >{\raggedright\arraybackslash}p{0.28\linewidth}
                    >{\raggedright\arraybackslash}p{0.2\linewidth}
                    >{\raggedright\arraybackslash}p{0.1\linewidth}}
        \toprule
        \textbf{Dataset} & \textbf{Modality} & \textbf{Brief description} & \textbf{Tasks} & \textbf{Paper} \\
        \midrule

        Amazon Polarity & Language & Product review sentiment & \multirow{6}{*}{XAI-focused} & \multirow{6}{*}{\cite{qiang_attcat_nodate}} \\
        IMDB & Language & Movie review sentiment &  &  \\
        MNLI & Language & Sentence-pair NLI &  &  \\
        QQP & Language & Paraphrase question pairs &  &  \\
        SQuAD & Language & QA over Wikipedia &  &  \\
        SST2 & Language & Sentence-level sentiment &  &  \\      
        \midrule

        Amazon-Movie & Language & Amazon movie reviews & NLE & \cite{guo_explainable_2023} \\
        \midrule
        \multirow{2}{*}{YELP} & \multirow{2}{*}{Language} & \multirow{2}{*}{Business review sentiment} & NLE & \cite{guo_explainable_2023} \\
         &   &  & XAI-focused & \cite{qiang_attcat_nodate} \\
        \midrule

        APTV-99 & Video+vision & Driving scenes & Object detection, Classification, NLE & \cite{wang_odp-transformer_2023} \\
        \midrule
        
        aPY & Vision & Object categories with attributes & \multirow{3}{*}{XAI-focused, Classification} & \multirow{3}{*}{\cite{rigotti_attention-based_2022}} \\
        CUB-200-2011 & Vision & Fine-grained bird categories &  &  \\
        MNIST Even/Odd & Vision & Digits labeled by parity &  &  \\
        \midrule

        AVSD & Video + audio + language & Dialog about scenes & MML & \cite{heo_natural-language-driven_2023} \\
        \midrule

        BDD-OIA & Vision & Driving scenes with interactions & VL, NLE & \cite{dong_why_2023} \\
        \midrule

        CNN/DailyMail & Language & News + summaries & \multirow{2}{*}{NLP} & \multirow{2}{*}{\cite{wang_exploring_2021}} \\
        NYT50 & Language & News with topic labels &  &  \\
        \midrule

        Code datasets & Code & Snippets and repositories & Code generation & \cite{mohammadkhani_explaining_2023} \\
        \midrule

        CounterFact & Language & Factual triples for probing & HCXI & \cite{katz_visit_2023} \\
        \midrule

        \multirow{3}{*}{\textit{Custom}\textsuperscript{*}} 
        & Vision+language & Proprietary VL data & VL & \cite{li_oscar_2020} \\
        & Vision+sensor & Private driving data & Object detection, NLE & \cite{sun_dfyolov5m-m2transformer_2023} \\
        & Numeric+vision & Physical measurements & Regression, VL & \cite{kandukuri_physical_2022} \\
        \midrule

        CUTE80 & Vision & Curved scene text 
        & \multirow{7}{*}{Scene text recognition} 
        & \multirow{7}{*}{\cite{buoy_explainable_2023}} \\
        
        ICDAR-2013 & Vision & Focused scene text &  &  \\
        ICDAR-2015 & Vision & Incidental scene text &  &  \\
        IIT5k-Words & Vision & Web word images &  &  \\
        SVT & Vision & Street View text &  &  \\
        SVTP & Vision & Perspective Street View text &  &  \\
        SynthTiger & Vision & Synthetic word images &  &  \\
        \midrule

        Fashion reviews & Language & Fashion product reviews & \multirow{2}{*}{NLP} & \multirow{2}{*}{\cite{malkiel_interpreting_2022}} \\
        Wine reviews & Language & Wine tasting reviews &  &  \\
        \midrule

        fMRI neural activation & Brain signals + text & Language stimuli 
        & \multirow{5}{*}{Classification, Regression} 
        & \multirow{5}{*}{\cite{sun_neural_2021}} \\
        
        SICK & Language & Sentence relatedness/entailment &  &  \\
        STS Benchmark & Language & Semantic similarity pairs &  &  \\
        Toronto Book Corpus & Language & Long-form book narratives &  &  \\
        UN-EWT & Language & Web sentences with syntax &  &  \\
        \midrule

        Gold Alignment Dataset & Language & Word alignments for MT & NLP, NMT & \cite{ferrando_attention_2021} \\
        \midrule

        Lyft & Trajectories+sensor & Driving logs & Regression & \cite{zhang_explainable_2022} \\
        \midrule

        Mboshi–French corpus & Speech+language & Low-resource parallel data & NLP, NMT & \cite{boito_investigating_2020} \\
        \midrule

        MLQE-PE & Language & MT outputs with QE labels & Regression & \cite{treviso_ist-unbabel_2021} \\
        \midrule

        \multirow{2}{*}{MS COCO 2014} 
        & Vision+caption & Objects and scenes & Semantic segmentation & \cite{yu_ex-vit_2023} \\
        & Vision+caption & Multimodal grounding & MML & \cite{chen_faster_2023, chefer_generic_2021, huang_generic_2023} \\
        \midrule

        OpenBookQA & Language & Science QA with facts 
        & \multirow{3}{*}{NLE, NLP} 
        & \multirow{3}{*}{\cite{hiemstra_using_2021}} \\
        
        ReQA Natural Questions & Language & Retrieval-based QA &  &  \\
        ReQA SQuAD & Language & Retrieval over SQuAD-style QA &  &  \\
        \midrule

        PASCAL VOC 2012 & Vision & Object masks and labels & Semantic segmentation & \cite{yu_ex-vit_2023} \\
        \midrule

        \multirow{2}{*}{VCR} & \multirow{2}{*}{Vision+language} & \multirow{2}{*}{Visual commonsense QA} & VL & \cite{parelli_interpretable_2023} \\
         & & &  VL, HCXI & \cite{aflalo_vl-interpret_2022} \\
        \midrule

        \multirow{2}{*}{VQA/VQAv2} & \multirow{2}{*}{Vision+language} & \multirow{2}{*}{Visual QA pairs} & XAI-focused, MML & \cite{chefer_generic_2021} \\
          &   &   & XAI-focused & \cite{huang_generic_2023} \\
        \midrule

        WebQA & Vision+language & Web-scale multi-hop QA & VL, HCXI & \cite{aflalo_vl-interpret_2022} \\
        \bottomrule
        \addlinespace[2pt]
        \multicolumn{5}{l}{\textsuperscript{*} Datasets with no public access available.}
    \end{tabular}
\end{table}

\section{Attention-based Architectures\label{Attention-based architectures}}
Attention mechanism originally gained popularity in sequence modeling and transduction models paired with recurrent or convolutional units. It allowed selective usage of frames in encoder representations for sequence-to-sequence generation tasks \cite{jaitly2016online, prabhavalkar2017analysis}. To improve the computational scalability, Vaswani \emph{et al.} introduced purely attention-based transducers, stripping away recurrence, referred to as the Transformer \cite{vaswani2017attention}. The vanilla transformer model used sub-word level tokenization in an NMT use-case with several special tokens (e.g., for masking). The embeddings of the tokens were combined with positional embeddings for encoding the positional information to account for the removal of recurrence. This token-based approach, along with partial transduction, allowed the architecture to model interactions between not only the inputs but also between the input and output tokens. As a result, transformers, with slight variations, have been adapted to modeling a diverse range of modalities (e.g., vision, time-series, graph, etc.) in addition to language models. The scalability also allowed effective adaptation in multimodal contexts. 

\subsection{Background}
There are different variants of attention used in the literature. We briefly introduce the major ones.

\subsubsection{Self-attention}
Self-attention is one of the most commonly used forms of attention and serves as a core component in the vanilla transformer encoder \cite{vaswani2017attention}. The input to self-attention is a fixed-shape vector, $Z_E$, which is generated through an embedding layer in transformer models. This embedding vector is often combined with positional encoding (either predefined or learned) as part of the processing pipeline. 

The embedding vector from the same source input is projected linearly onto three distinct matrices: query ($Q_S$), key ($K_S$), and value ($V_S$), calculated as in equation \ref{eq:qkv}.

\begin{equation}
    Q_S = Z_EW^{Q_S}, \quad K_S = Z_EW^{K_S}, \quad V_S = Z_EW^{V_S},
\label{eq:qkv}
\end{equation}

Where self-attention, $SA(Q_S, K_S, V_S)$, in the vanilla transformer model can be defined as in equation \ref{eq:sa}.

\begin{equation}
    U_S = SA(Q_S, K_S, V_S) = \mathbf{softmax}\left(\frac{Q_S {K_S}^{\intercal}}{\sqrt{d_k}}\right) V_S.
\label{eq:sa}
\end{equation}

Here, $\frac{1}{\sqrt{d_k}}$ is a scaling factor applied to stabilize the attention scores. 

A key advantage of self-attention is that it allows every token in the input sequence to attend to every other token, thereby capturing long-range dependencies and global interactions within the data. Moreover, self-attention can be augmented with a masking operation before the $softmax$ step. This masked self-attention is particularly useful in scenarios such as auto-regressive decoders, where the model must prevent access to future tokens during training, ensuring that predictions are based solely on past and present context.

\subsubsection{Cross-attention}
Cross-attention facilitates interactions between tokens from distinct source sequences. Its formulation mirrors self-attention with key differences outlined in equation \ref{eq:ca}.

\begin{equation}
    U_T = CA(Q_T, K_S, V_S) = \mathbf{softmax}\left(\frac{Q_T {K_S}^{\intercal}}{\sqrt{d_k}}\right) V_S.
\label{eq:ca}
\end{equation}

This mechanism allows one set of representations to focus on another, making it ideal for multimodal applications. Typically, for cross-attention in a single direction, the query ($Q$) originates from one sequence while keys ($K$) and values ($V$) come from another. However, without additional mechanisms, cross-attention does not provide global context from each set of representations in isolation, but only models their interactions.

\subsubsection{Sparse attention}
In self-attention, attention scores for each token are calculated from all other tokens in the sequence, which affects the encoding of the locality of references. Sparse self-attention, specifically, in case of structural sparsity, addresses this issue by restricting each query ($Q_i$) to selectively attend to only a subset ($S(i)$) of keys, as formulated in equation \ref{eq:sparse}.

\begin{equation}
SPARSE(Q, K, V)_i = \sum_{j \in S(i)} \alpha_{ij} V_j,
\label{eq:sparse}
\end{equation}

with the attention weights \( \alpha_{ij} \) defined as in equation \ref{eq:sparse_alpha}.

\begin{equation}
\alpha_{ij} = \frac{\exp\left(\frac{Q_i \cdot K_j}{\sqrt{d_k}}\right)}{\sum_{j' \in S(i)} \exp\left(\frac{Q_i \cdot K_{j'}}{\sqrt{d_k}}\right)}.
\label{eq:sparse_alpha}
\end{equation}

Here, \( S(i) \) is the set of key indices that the $i^{th}$ query is allowed to attend to. The $softmax$ normalization is computed only over the indices in \( S(i) \), ensuring that only the selected keys contribute to the output.

This sparsity reduces computational complexity and can be particularly useful in handling long sequences, as seen in architectures like the Sparse Transformer and BigBird. However, other variants of sparse attention for functional sparsity are also commonly used in cases of adaptive sparsity.

\subsection{Multimodal Attention}
\begin{figure*}
    \centering
    \includegraphics[width=\linewidth]{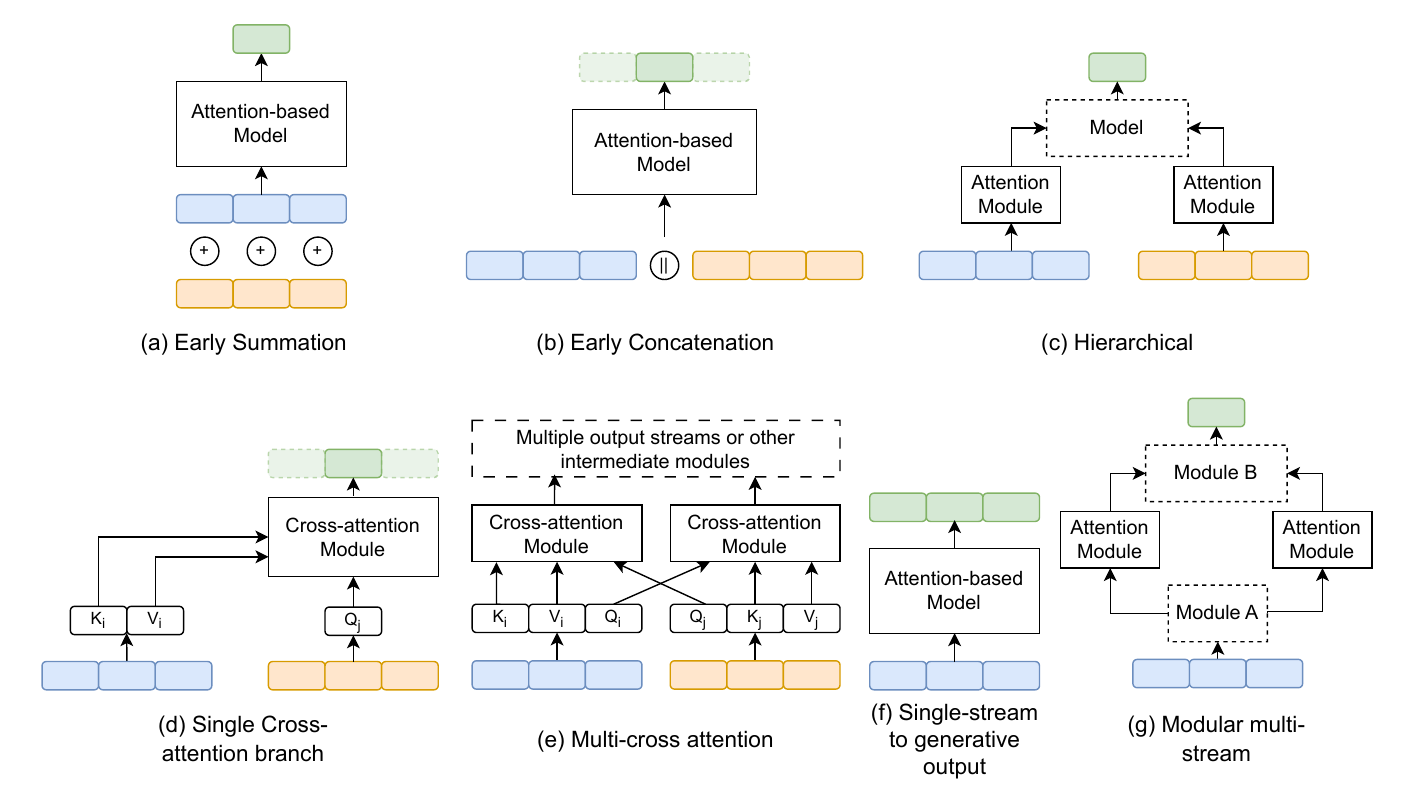}
    \caption{Block diagram illustrating various fusion architecture types: Early fusion (a, b); Hierarchical fusion (c); Cross-attention variants (d, e); and other fusion categories (f, g). Blue and orange rectangles denote input representations from different sources. Green rectangles indicate simpler outputs (e.g., classification probabilities), while a series of them suggests outputs can also be in different modalities. Shaded models or  modules can have arbitrary architectures. $Q$, $K$, and $V$ are the query, key, and value matrices in transformers.}
    \label{fig:arch}
\end{figure*}

Attention-based architectures, such as transformers, have been effectively extended to handle multimodal inputs and generative outputs. Several architectural variants have emerged to support multimodality, differing primarily in how and when they fuse information from multiple sources. To classify the models employed in the retrieved papers, we adapt the taxonomy proposed by Xu \emph{et al.} \cite{xu2023multimodal}, generalizing it from transformer-based to all attention-based architectures. This classification is particularly useful because it is agnostic to tokenization and embedding strategies, and is based instead on the fusion mechanism of input modalities. We group the original six categories into three broader architectural classes: 1) Early Fusion, 2) Hierarchical Architectures, and 3) Cross-Attention Variants. In addition to these, we observe two additional categories based on input-output structure:
Single-stream to Generative Output and Modular Multi-stream Processing. The block diagram in Figure \ref{fig:arch} illustrates the general architecture categories and Table \ref{tab:arch} summarizes the distribution of reviewed papers across these categories.

\begin{table}[h]
\caption{Architecture Variants of Multimodal Attention-based Models\\
\textbf{Legends: *} indicates studies that employ different architectures.}
\label{tab:arch}
\scriptsize
\begin{tabular}{p{0.25\linewidth} | p{0.5\linewidth} | p{0.08\linewidth}}
\hline
\textbf{Multimodal Architecture} & \textbf{Papers} & \textbf{Count} \\
\hline
\multicolumn{3}{l}{\textbf{Early Fusion}} \\
\hline
Early Summation & Classification: \cite{meng_bidirectional_2021}. & 1 \\
\hline
Early Concatenation & Classification: \cite{abdulkadir_augmenting_2022, ukwuoma_hybrid_2023}; Regression: \cite{chiewhawan_explainable_2020, wu_interpretable_2023, treviso_ist-unbabel_2021, wang_tfregnci_2023}; VL: \cite{chen_faster_2023, naseem_vision-language_2023, li_oscar_2020}; MML: \cite{chefer_generic_2021}*, \cite{heo_natural-language-driven_2023}; NLP: \cite{xu_multi-granular_2020}; NLE: \cite{guo_explainable_2023}. & 13 \\
\hline
\multicolumn{3}{l}{\textbf{Hierarchical Architectures}} \\
\hline
Hierarchical Multi-to-One & Classification: \cite{janssens_360_2024, yang_explainable_2023, ding_deepstf_2023, feucht_description-based_2021, kumar_towards_2021, ullah_explainable_2022}; Regression: \cite{zhang_explainable_2022}; NLP: \cite{zanzotto_kermit_2020, xu_logiformer_2022, jha_supervised_2023}. & 10 \\
\hline
\multicolumn{3}{l}{\textbf{Cross-Attention Variants}} \\
\hline
Single Cross-Attention Branch & Classification: \cite{rigotti_attention-based_2022, che_multiscale_2023}; VL: \cite{parelli_interpretable_2023, ilinykh_what_2021, dong_why_2023, lin_zero-shot_2023}; Multi-task: \cite{sun_dfyolov5m-m2transformer_2023, wang_odp-transformer_2023}; NLP: \cite{ferrando_attention_2021, boito_investigating_2020, ferrando_towards_2022}; Code: \cite{mohammadkhani_explaining_2023}; MML: \cite{chefer_generic_2021}*. & 13 \\
\hline
Multi-Cross Attention (Bidirectional) & Classification: \cite{huang_representation_2023, wang_novel_2023, koyama_attention_2023}; MML: \cite{chefer_generic_2021}, \cite{bhargava_adaptive_2020, huang_generic_2023}. & 6 \\
\hline
\multicolumn{3}{l}{\textbf{Other Architectures}} \\
\hline
Single-Stream to Generative Output & Multi-task: \cite{buoy_explainable_2023, sun_neural_2021, kandukuri_physical_2022}; NLP: \cite{qiang_attcat_nodate, wang_exploring_2021}. & 5 \\
\hline
Modular Multi-Stream Processing & Classification: \cite{du_case-based_2023, kumar_bert_2022}; NLP: \cite{malkiel_interpreting_2022, hiemstra_using_2021}; Segmentation: \cite{yu_ex-vit_2023}; Regression: \cite{xiao_transformer_2024}. & 6 \\
\hline
\end{tabular}
\end{table}

\subsubsection{Early Fusion}
A very classical method in ML for integrating information from multiple sources is through creating a joint representation before passing to the downstream model. Modern ML often uses embedding vectors as representations and are often extracted from other pretrained models. The two methods to use early fusion are as follows:

\paragraph{Early Summation} A simple way to fuse embedding vectors into a meaningful combined representation is to add them together. The only work using this method in the review, proposed by Meng \emph{et al.}, is a fusion technique for their depression prediction system using the BRLTM (Bidirectional Representation Learning model with a Transformer architecture on Multimodal EHR) architecture \cite{meng_bidirectional_2021}. The coded embeddings for diagnoses, procedures, medications, and topics from electronic health records (EHR) are summed together with embeddings for age and gender, and a special segment embedding for distinguishing between multiple visits. This early fusion enables access to demographic information as features along with a temporal representation.

\paragraph{Early Concatenation} This is a general alternative to summation when the embedding vectors come from very different sources and when semantic information between modalities is crucial. Concatenation enables the modeling of dense interactions between tokens of different modalities. This straightforward integration can be particularly useful in simpler tasks such as classification or regression. For instance, in classification tasks, an ensemble of features from pre-trained vision models can be used \cite{ukwuoma_hybrid_2023}, or for medical images, relevant tabular features (e.g., encoding region volumes, cortical thickness, and radiomics properties) can be combined together with visual features \cite{abdulkadir_augmenting_2022}. For forecasting problems, contextual data can be readily provided along with the input \cite{chiewhawan_explainable_2020, wu_interpretable_2023}. In VL tasks like VQA, representations of the question and visual cues can be concatenated \cite{chefer_generic_2021, naseem_vision-language_2023}. In contrast, when learning generic image-text representations, additional tokens that capture object-level information can be incorporated \cite{li_oscar_2020, chen_faster_2023}. Other works that use early concatenation are in translation quality estimation (constrained) \cite{treviso_ist-unbabel_2021}, multi-granular text pair classification \cite{xu2023multimodal}, non-covalent interaction correction \cite{wang_tfregnci_2023}, audio-visual scene aware dialog system \cite{heo_natural-language-driven_2023}, and recommendation generation system \cite{guo_explainable_2023}.

\subsubsection{Hierarchical Architectures: Hierarchical Multi-
to-One}
Hierarchical architectures are designed to capture complex inter-dependencies across modalities while offering flexibility. These models employ a modular hierarchy where multiple streams representing different modalities are processed independently before being fused later in the network, often through a dedicated module. Importantly, these modules can be non-attention-based as well. With the availability of powerful pre-trained language models, hierarchical architectures are widely used for NLP and text-based classification tasks, encompassing diverse application areas. For instance, in social media it can be used for rumor detection by combining text and structured features for the context \cite{janssens_360_2024}, in malware detection this can include combining encoding of HTTP flow and TCP streams as text and malware as image \cite{ullah_explainable_2022}, and in speech emotion recognition this involves combining text transcripts with speech audio \cite{kumar_towards_2021}. Multichannel solutions on document understanding \cite{xu_logiformer_2022, jha_supervised_2023} and sequence generation \cite{zanzotto_kermit_2020} can also employ this architecture. Additionally, it supports classification tasks such as transcription factor binding site (TFBS) prediction \cite{ding_deepstf_2023} and remote sensing scene classification \cite{yang_explainable_2023}. The only work on a regression problem by Zhang and Li is multimodal trajectory prediction by encoding environmental context images using a Swin transformer and historical context using a gated recurrent unit (GRU) \cite{zhang_explainable_2022}.

\subsubsection{Cross-Attention Variants}
Cross-attention, by design, models cross-modal interactions. This ability allows varied implementation of the cross-attention mechanism in different architectures, such as in encoder-decoder models for the target sequence tokens to attend to the source tokens, or in multimodal models where cross-interactions within all individual modalities are required to be modeled. We further classify the cross-attention-based methods into two different classes based on how many branches of cross-attention are used in the network:

\paragraph{Single Cross-Attention Branch}
Architectures with a single cross-attention branch only require one modality to attend to the other and not the other way around. The vanilla transformer used such an approach, which has been adopted extensively in various other tasks. The most common use of this design is in explainability-driven VL tasks. This includes VQA tasks where text-based justification for the correct answer is used as an additional supervision in a dual decoder setup \cite{parelli_interpretable_2023}, image captioning model for reason-induced autonomous driving systems~\cite{dong_why_2023} or captioning from visual concepts of objects~\cite{ilinykh_what_2021}, zero-shot image retrieval from sketches modeled as cross-modal matching problem~\cite{lin_zero-shot_2023}. Conceptual inputs for cross-reasoning can also be used in classification problems \cite{rigotti_attention-based_2022}. Another notable work by Che \emph{et al.} implemented classification for bearing fault diagnosis, which involved training sparse attention-based multiple encoders on different scales of signal representations and combining them through a single decoder with a cross-attention mechanism \cite{che_multiscale_2023}. In NLP, the architecture is mostly used in observing alignment in different NMT setups \cite{ferrando_attention_2021, boito_investigating_2020, ferrando_towards_2022}. In multi-task learning, single-branch cross-attention can be used in two-stage image captioning models as a basis for recognition. The first stage involves extracting regions of interest, and the second stage generates dense captions \cite{sun_dfyolov5m-m2transformer_2023} or captions with classification outputs \cite{wang_odp-transformer_2023}. Other uses include implementing LLMs in coding tasks (code document generation, code refinement, and code translation) \cite{mohammadkhani_explaining_2023} and MML experiments \cite{chefer_generic_2021}.

\paragraph{Multi-cross Attention (Bidirectional)}
 In contrast to single cross-attention branches, multi-cross attention architectures enable interactions in more than one direction within tokens from different source/target modalities. This design is particularly valuable for complex tasks requiring reciprocal attention and integrated modality interactions, such as MML and multimodal classification. In MML, Chefer \emph{et al.} \cite{chen_faster_2023} and Bhargava \cite{bhargava_adaptive_2020} study the LXMERT model \cite{tan2019lxmert}, which is a VL encoder for representing cross-modal learning. In LXMERT, a cross-modality encoder is stacked on top of object-relationship and language encoders for bidirectionally aligning the modalities. Bhargava replaces the $softmax$ function in self-attention with an $\alpha-entmax$ \cite{correia2019adaptively} for introducing functional sparsity to make the attention weights sparse. Other work using MML includes the study of the CLIPmapper model on different VL tasks such as image captioning and VQA \cite{huang_generic_2023}. For classification, application in drug discovery involves the use of structural features extracted from multiple source sequences (e.g., protein or amino acids) and processing through separate encoders \cite{wang_novel_2023, koyama_attention_2023}. The architecture can also be used in multichannel classification cases such as heart failure prediction from electronic medical record data \cite{huang_representation_2023}. 
 
\subsubsection{Other Architectures}
Besides the multichannel and multimodal architectures, other key architecture groups can be defined based on how many input streams the networks handle. 

\paragraph{Single-Stream to Generative Output}
Some tasks involve generating outputs from complex spaces (e.g., image, text). The input is generally unimodal, forming a single stream of information that can be encoded into a latent representation before generating the output. Studies using such architectures in most cases are multi-task learning models, for instance, in neural encoding and decoding of distributed semantic models \cite{sun_neural_2021}, learning physically interpretable latent representation of videos \cite{kandukuri_physical_2022}, or scene text recognition \cite{buoy_explainable_2023}. Other than that, this architecture can be implemented in NLP-based tasks such as controlled document summarization \cite{wang_exploring_2021} or question-answering \cite{qiang_attcat_nodate}.

\paragraph{Modular Multi-Stream Processing}
Often, an unimodal task is processed as a multimodal problem by splitting the unimodal signal into multiple streams, which are often then merged using hierarchical fusion techniques. This architecture is flexible and can be employed in various tasks. In classification, emotion recognition from selected channels of EEG \cite{du_case-based_2023} or from text using a dual-channel network \cite{kumar_bert_2022}. In NLP, this type can be used for unsupervised text similarity problems \cite{malkiel_interpreting_2022} or evidence retrieval for QA tasks \cite{hiemstra_using_2021}. Other example uses include gene expression prediction from histopathological images \cite{xiao_transformer_2024} or semantic segmentation learned from different transformations of the same input \cite{yu_ex-vit_2023}.

The rest of the studies excluded from this section discuss interfaces to present explainability \cite{sarti_inseq_2023, katz_visit_2023, aflalo_vl-interpret_2022} and are not primarily focused on specific model-based analysis.

\subsection{Discussion}
Architectures play a crucial role in determining how the explainable algorithm is going to be applied, particularly when using model-specific methods. The wide range of possible combinations of modalities inherently increases the interpretive complexity, while the variety in architectures further complicates it. 

In this review, the multimodal architectural groups of early summation and single-stream to generative outputs are used only by a few studies. By design, these groups only fit very specific use cases and are limited in encoding explainable multimodal interactions. On the other hand, variants of early concatenation and single cross-attention branch are more frequently used in a wide range of tasks, apart from a few exceptions, such as the single cross-attention branch not being used for regression tasks as frequently. After classification, VL and NLP are two broad areas that were the most frequent in this review. Publications from the review on NLP inhibit diverse architectures having representations from almost all the architecture groups, whereas publications on VL tasks used either early concatenation or single cross-attention branches. Hierarchical and modular architectures are preferred for simpler tasks such as classification or regression, and are underexplored in other domains. Notably, cross-attention-based models are gaining traction, particularly in multimodal and multi-task learning. This trend may indicate a shift toward architectures that allow greater modality interaction flexibility while allowing robust interpretability techniques. Overall, the diversity in types of architecture in solving similar problems suggests that there is no specific architecture that fits well with different multimodal problems. Hence, future studies must focus on benchmarking different architecture types across tasks on cross-modal interactions and multimodal task performance with a specific focus on explainability.

\section{Explanation Algorithm\label{Explanation Algorithm}}
\begin{figure}[]
    \centering
    \includegraphics[width=1\linewidth]{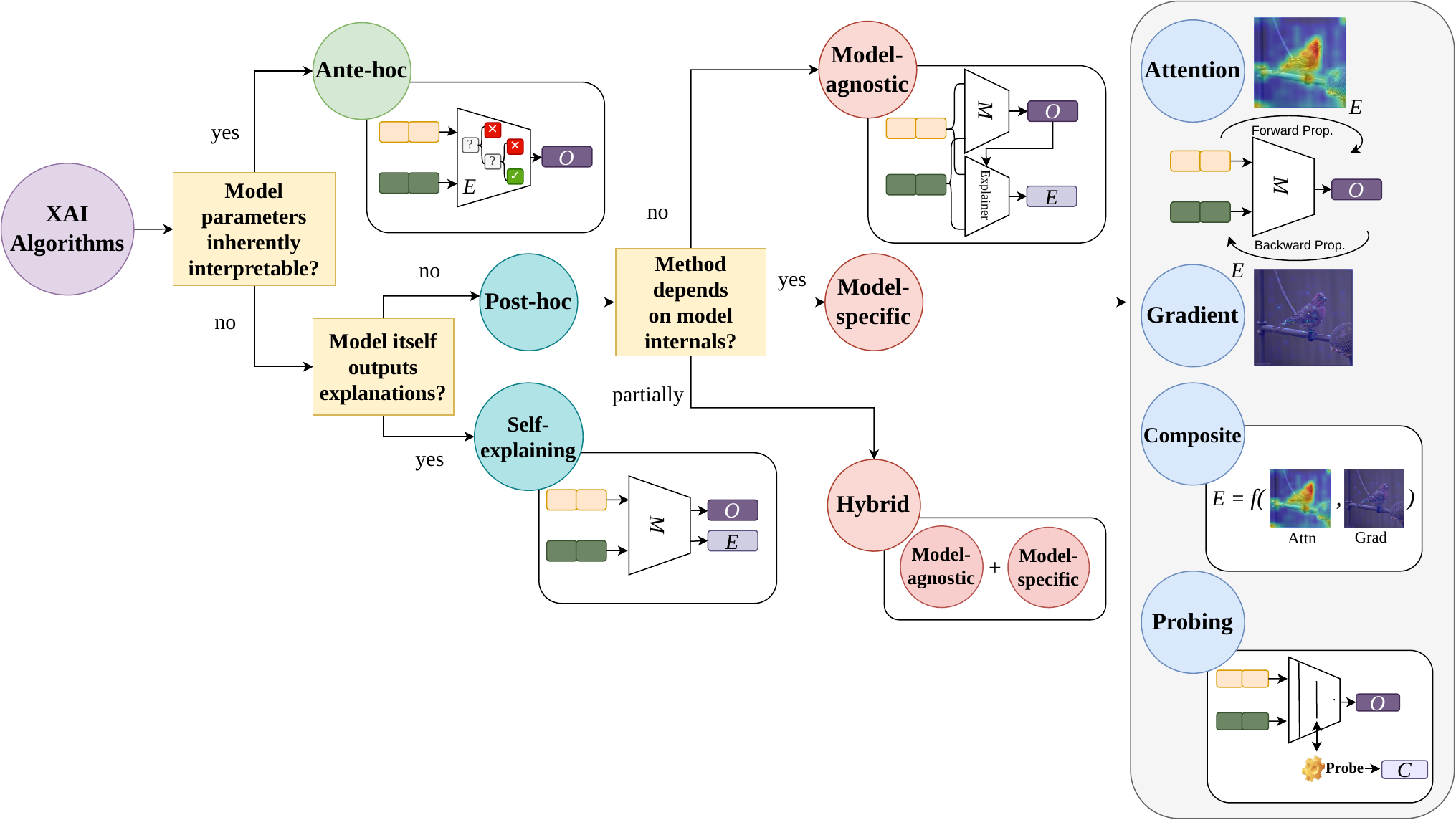}
    \caption{Taxonomy of XAI algorithms for multimodal models. Colored rectangles denote multimodal inputs to model $\mathit{M}$, producing output $\mathit{O}$; $\mathit{E}$ represents generated explanations that may contain interpretable concepts $\mathit{C}$.}
    \label{fig:xai_class}
\end{figure}

\begin{figure*}[]
\begin{subfigure}[b]{\linewidth}
\begin{tikzpicture}[
    level 1/.style={level distance=3.5cm, sibling distance=-5cm},
    level 2/.style={level distance=3cm, sibling distance=-3cm,},
    level 3/.style={level distance=6cm, sibling distance=-2.3cm},
    grow=right,
    font=\footnotesize,
    scale=0.6,
    every node/.style={
        rectangle, 
        draw, 
        align=center,
        anchor=west 
    },
    edge from parent path={
        (\tikzparentnode.east) -- (\tikzchildnode.west)
    },
    child anchor=west 
]
\node {Explanation\\Methods}
    child { node {Ante-hoc \\ \cite{rigotti_attention-based_2022, yu_ex-vit_2023, buoy_explainable_2023, kandukuri_physical_2022, abdulkadir_augmenting_2022, wu_interpretable_2023}}}
    child { node {Post-hoc}
        child { node {Model-agnostic \\ \cite{janssens_360_2024, ullah_explainable_2022}}}
        child { node {Model-specific}
                child { node {Attention-based \\\cite{meng_bidirectional_2021, li_oscar_2020, mohammadkhani_explaining_2023, xu_multi-granular_2020, boito_investigating_2020, wang_exploring_2021, dong_why_2023, ding_deepstf_2023, kumar_towards_2021}, \\\cite{feucht_description-based_2021, xu_logiformer_2022, huang_representation_2023, ferrando_towards_2022, wang_novel_2023, ilinykh_what_2021, kumar_bert_2022, jha_supervised_2023, che_multiscale_2023}} }
                child { node {Gradient-based \\\cite{xiao_transformer_2024, yang_explainable_2023, chiewhawan_explainable_2020}} }
                child { node {Attention composite \\\cite{du_case-based_2023, zhang_explainable_2022, malkiel_interpreting_2022, qiang_attcat_nodate, chefer_generic_2021, huang_generic_2023, koyama_attention_2023, ferrando_attention_2021}} }
                child { node {Probing and ablation \\\cite{sun_neural_2021, hiemstra_using_2021, bhargava_adaptive_2020}} }
        }
        child { node {Hybrid \\ \cite{zanzotto_kermit_2020, wang_tfregnci_2023, ukwuoma_hybrid_2023, treviso_ist-unbabel_2021, lin_zero-shot_2023, naseem_vision-language_2023}}}
    }
    child {node {Self-explaining \\ \cite{sun_dfyolov5m-m2transformer_2023, wang_odp-transformer_2023, chen_faster_2023, parelli_interpretable_2023, guo_explainable_2023, heo_natural-language-driven_2023}}};
\end{tikzpicture}
\caption{Classification tree of the explanation algorithms}
\label{fig:algo_tree}
\end{subfigure}
\hfill
\centering
\begin{subfigure}[b]{\linewidth}
    \centering
    \includegraphics[width=1\linewidth]{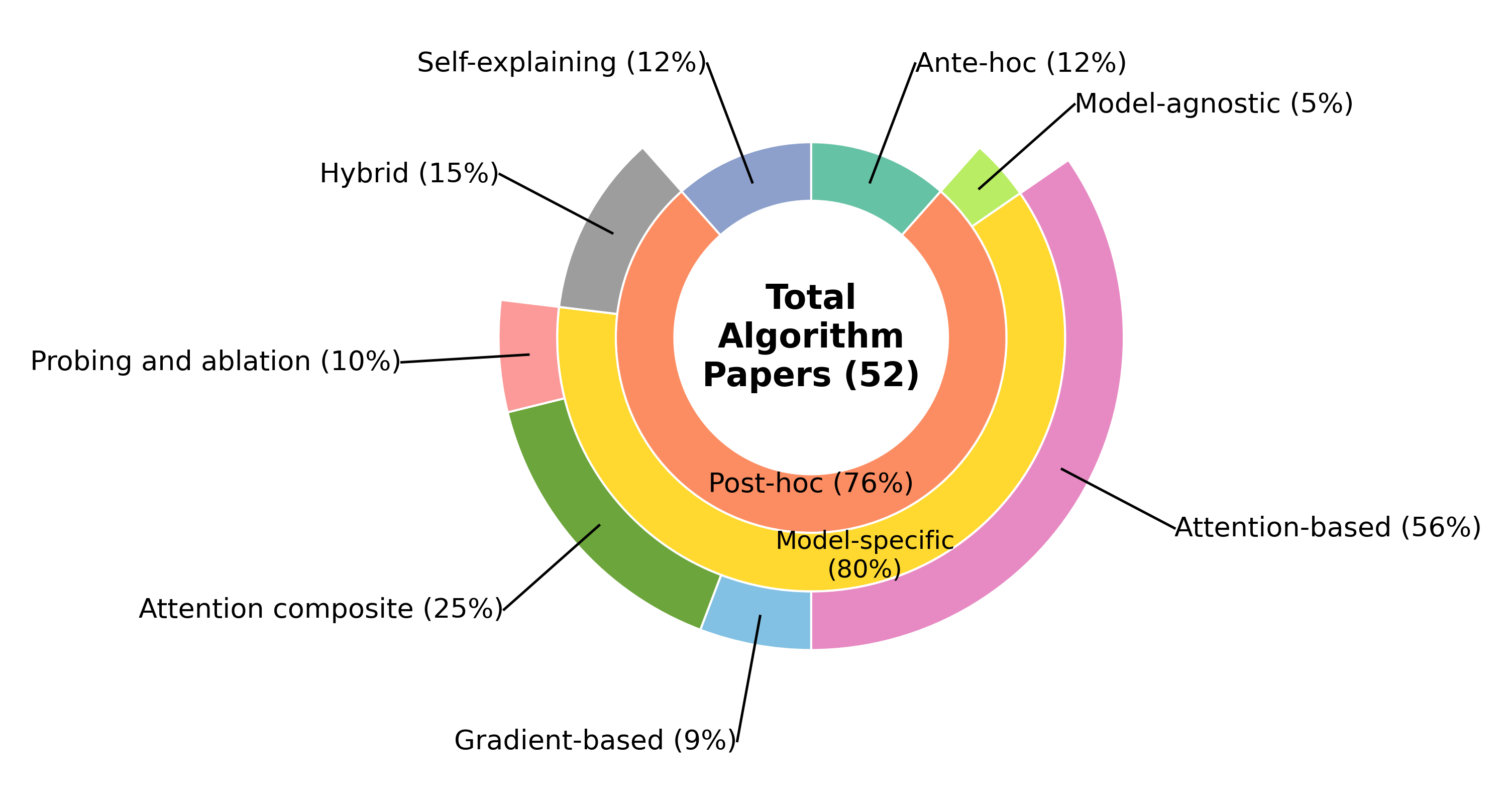}
    \caption{Distribution of implemented XAI algorithms for different classes}
    \label{fig:algo_donut}
\end{subfigure}
\caption{Classification and distribution of explanation algorithms}
\end{figure*}

A key focus of this review is to explore how explanations are generated in multimodal applications. Due to the diverse constraints imposed by different domains and use cases, a wide range of explainable algorithms have been developed. These algorithms can be categorized in various ways, depending on the dimension under consideration \cite{vilone2021classification}. In this review, we adopt a hybrid classification framework, synthesizing hierarchical structures proposed in prior studies. The primary basis for categorization is the stage at which the explanation is provided \cite{xu2019explainable, burkart2021survey}. The resulting taxonomy is illustrated as a flow-based classification tree in Figure~\ref{fig:xai_class}. Figure~\ref{fig:algo_tree} presents the same hierarchy, listing representative studies at each leaf. Finally, Figure~\ref{fig:algo_donut} summarizes the distribution of papers across the proposed categories.

\subsection{Ante-hoc Explanations}
Ante-hoc explainability refers to transparency-oriented models that are intrinsically interpretable. The ante-hoc interpretability can be used for local decisions using local explainability methods. Although ante-hoc models sometimes may refer to white-box, fully transparent models, here we consider models that are designed with a focus on built-in interpretability. In this review, three publications provide ante-hoc interpretations. 

Interpretability can be achieved by training models to learn high-level concepts or attributes. Concept Transformers, introduced by Rigotti \emph{et al.}, generalizes attention weights from low-level features to externally provided high-level concept representations \cite{rigotti_attention-based_2022}. These concepts originate in the target domain and can be specific spatial features or global features. Concepts are learned for classification tasks without additional overhead during training. This is achieved by integrating the vector representations of the concepts directly into the cross-attention process. The query vector, $Q_I$, remains as the patch representation from the original input. But the key, $K_C$, and value, $V_C$, are linearly projected and concatenated representations of the concept vectors. The logits are calculated for providing the probability distribution over the output classes by multiplying an output matrix, $O$, with the attention outputs and then averaging over all the input patches. The formulation of the logit calculation is as in equation \ref{eq:ct}.

\begin{equation}
    {logit}_i = \frac{1}{P} \sum_{p=1}^{P}{{CA(Q_I,K_C,V_C).O]}_{pi}}{}
    \label{eq:ct}
\end{equation}

Yu \emph{et al.} extend intrinsic explainability through a self-supervised learning method in eXplainable Vision Transformer (eX-ViT) \cite{yu_ex-vit_2023}. The model includes two specific components—Explainable Multi-head attention (E-MHA) and attribute-guided explainer—to learn high-level interpretable attributes that are distinctly identifiable for different objects. The E-MHA calculates the attention scores in an attempt to maximize alignment with input tokens. This is achieved by introducing a scaling function (L2 norm) into the self-attention mechanism. The formulation for the proposed self-attention, $SA'$, is as presented in equation \ref{eq:yu_sa}.

\begin{equation}
\begin{split}
    A =& \mathbf{f_{\theta}}{\left(\frac{Q {K}^{\intercal}}{\sqrt{d}} + b\right)}^{\intercal} \\
    where, \mathbf{f_{\theta}}(x) =& \frac{x}{||x||^2} \\
    SA' =& A^\intercal V
\label{eq:yu_sa}
\end{split}
\end{equation}

The upper bound to the E-MHA ensures alignment of the attention weights, $A$, to the discriminative features in $V$, removing the distortion caused by the probability of $softmax$ as in regular self-attention. As a result, it allows a better understanding of how each token contributes to the final output. On the other hand, the attribute-guided explainer helps decompose the attention map into a set of distinct high-level attributes. This module is applied to the feature maps from the final layer of the eX-ViT's encoder for extracting trainable attributes that encode the concept of objects better. In addition, eX-ViT is trained with a composite attribute-guided loss function that combines loss values for discriminability and diversity. Discriminability loss penalizes the different types of input views, whereas diversity loss promotes effective decomposition of attribute diversity.

For scene text recognition, Buoy \emph{et al.} introduced a single-stream to generative output architecture consisting of a ViT encoder and a 1D CTC decoder \cite{buoy_explainable_2023}. The 2D spatial feature maps ($F$) from the ViT are converted with a marginalization method to be used with the 1D CTC decoder. This is achieved by first passing the feature maps from the ViT through a linear layer to create unnormalized probability distribution scores with the same height and width as the feature maps, however, with the embedding dimensions remapped to the number of classes. Softmax operation is then applied to the scores across the height and the class dimensions before marginalizing over the height dimension. As a result of this marginalization method, the efficiency of CTC is preserved without losing the information of the height dimension, as is the case with 1D CTC decoders generally. The probability scores before marginalization serve as an association map (AM), which highlights the most important image regions for the prediction. Thus, along with prediction outputs, the AM provides inherent interpretability to the prediction model.

The only physics-aware neural network in this review is introduced by Kandakuri \emph{et al.} \cite{kandukuri_physical_2022} for physical representation learning from videos. The proposed model uses physical parameter extraction from frames of videos and reconstructs the frame using a spatial transformer network (STN). The differential physics layer for pose estimation ensures that the model is inherently interpretable.

Lee \emph{et al.} used a two-stage network for medial temporal lobe atrophy (MTA) score prediction from multimodal imaging data: A ViT for accumulating image features and a TabNet classifier that processes the merged features for classification \cite{abdulkadir_augmenting_2022}. Here, the TabNet is inherently interpretable, and the feature importance from TabNet is used to identify the most influential tabular features. Another inherently interpretable model for tourism demand forecasting was proposed by Wu \emph{et al.} \cite{wu_interpretable_2023}. They introduce a temporal fusion transformer model, which is optimized by the adaptive differential evolution algorithm. As a result, the explainable model can be used to observe the feature importance of past and future inputs.

\subsection{Post-hoc Explanations}
Post-hoc explainability is designed to explain the inference process of a model after it has been trained. These types of explanations may include, for instance, providing analytics and insights, visualizing different aspects of the decision-making process, and offering explanations by example \cite{xu2019explainable}. This wide set of possibilities and not requiring a change in the way modeling is done makes post-hoc explanation methods a popular choice. We discuss further classification of post-hoc explanation methods based on whether the explanation process is influenced by the choice of the model architecture.

\subsubsection{Model-agnostic methods}
Model-agnostic methods for interpretability are not strictly dependent on the selection of the model type and are generally applicable to any network architecture for the same task. This can be achieved by observing the inputs and outputs of the model without requiring access to the model parameters. These methods have been well-established and quite deeply analyzed in several studies \cite{burkart2021survey, vilone2021classification, fantozzi2024explainability}; hence, for brevity, we just mention how the techniques are used. 

The most common way to achieve model-agnostic interpretability is through perturbation and by example. Local Interpretable Model-agnostic Explanation (LIME) and Shapley Additive exPlanations (SHAP) are two of the most commonly used algorithms. Janssens \emph{et al.}, for rumor detection from social media data, explored interpretability by applying LIME due to its computational efficiency and adaptability for high-dimensional data \cite{janssens_360_2024}. They applied LIME on both structured and unstructured data through both interpretable and non-interpretable representations. The result of the produced explanations was local and global feature contributions. SHAP was adopted for explaining the malware-detection model by Ullah \emph{et al.} \cite{ullah_explainable_2022}. SHAP values for the total of 32 features were used to explain how each feature influenced model decisions compared to the expected predictions. Also, SHAP was used to identify feature contributions when restricted to a specific class (benign vs. malware).

Despite their prominence in the field of XAI, implementations of these techniques in multimodal contexts remain limited. Moreover, other perturbation-based approaches are often used in conjunction with model-specific methods.

\subsubsection{Model-specific methods}
Model-specific or the decompositional approach generates explanations from the internal structure, parameters, and feature representation of the prediction model. As this review focuses on attention-based methods, most studies applying model-specific XAI methods leverage the attention weights. The use of attention weights as explanations, although debated \cite{jain2019attention}, allows the generation of intuitive methods of observing the internal mechanics of a model \cite{abnar2020quantifying}. Apart from attention weights, there are several other methods found in the literature for generating model-specific explanations. These methods are described following the classes used by Fantozzi \emph{et al.} \cite{fantozzi2024explainability}.

\paragraph{Attention-based methods}
The attention scores are a matrix of cross-token probabilities. Hence, these can depict the amount of influence the tokens have on each other with respect to the final prediction. Observing the self-attention interactions was demonstrated by Vaswani \emph{et al.} when introducing the vanilla transformer model \cite{vaswani2017attention}. However, a major issue with the interpretability through attention is the aggregation of the matrices from different attention heads and layers.

Final layers of neural networks are known to encode high-level, abstract features that best represent the input-output relationship. Consequently, one of the most common attention-based XAI methods is to analyze the attention scores only from the final layer. In their study, Meng \emph{et al.} visualized the attention weights of the BERT-based encoder, which express the latent associations within EHR data of patients for depression detection \cite{meng_bidirectional_2021}. The work uses the BertViz tool \cite{vig-2019-multiscale} to depict the attention component from the last layer of the network; however, the head aggregation mechanism is not specified. Ding \emph{et al.} visualized the attention weights from the final output layers of both the sequence processing and shape processing modules. They then visualized the attention weights from the final output layers after fusing the two TFBS streams \cite{ding_deepstf_2023}. Additionally, they presented the attention weights of all eight heads in the transformer component of their DeepSTF model. Interactions between TFBSs were illustrated by varying color intensity in proportion to the attention strength. For their DLAC model, Feucht \emph{et al.} demonstrated explainability through ``top attention scores" \cite{feucht_description-based_2021}. For different target ICD-9 classes, the attention scores reflect how different parts of the discharge summary (one out of two input streams) are significant. However, the method for calculating the top attention scores is not specified. For the explainability analysis of their code models, Mohammadkhani \emph{et al.} used normalized attention scores (in [0,100] range) by averaging the values in all the decoder layers and reporting them for each category of code tokens grouped by their types (naming, structural, others) \cite{mohammadkhani_explaining_2023}. In their work of controlled abstractive summarization, H. Wang \emph{et al.} proposes the use of an interaction matrix, $Q^s$, where $s$ denotes the sentences in the input \cite{wang_exploring_2021}. These matrix incorporates the directional influence, $q_{AB}$, of sentence A on sentence B where $q_{AB}$ is measured as follows:

\begin{equation}
\begin{aligned}
q_{AB}(h_A, s_A, h_B, d) &= \sigma\Big(
    \underbrace{\bm{W}_c h_A}_{\text{informativeness}} 
    + \underbrace{h_A^\top \bm{W}_r d}_{\text{relevance}} \\
    &\quad + \underbrace{h_A^\top \bm{W}_s h_B - h_A^\top \bm{W}_n \tanh(a_A)}_{\text{novelty}} 
    + \bm{b}_{\text{matrix}}
\Big)
\end{aligned}
\end{equation}

Here, $W$ and $b$ are trainable parameters, $h$ denotes the representation of the respective sentence, and $a$ is the summary representation accumulated with respect to the current sentence. The use of attributes specific for informativeness, relevance, and novelty gives the matrix a meaningful representation, and hence, the matrix can be visualized for analyzing each of the attributes.

Averaging the weights over all the attention heads is a commonly used method implemented in techniques such as attention rollout \cite{abnar2020quantifying}. Dong \emph{et al.} used averaged attention maps across all heads of the final cross-attention layer in the decoder to illustrate cross-modal interactions as a form of explanation \cite{dong_why_2023}. Che \emph{et al.} gathered the attention weights of each head from different layers and averaged them into a single matrix reflecting patch importance \cite{che_multiscale_2023}. This helps visualize all three encoders and the decoder weights. Similarly, Y. Huang \emph{et al.} \cite{huang_representation_2023}, F. Xu \emph{et al.} \cite{xu_logiformer_2022}, and S. Xu \emph{et al.} \cite{xu_multi-granular_2020} used average attention scores from the last attention layer to visualize model rationale. In each of these, interactions between all modeled input streams are covered. Other than these, Zheng \emph{et al.} also presented the interpretability of their work on drug-target interaction by visualizing the attention weights \cite{wang_novel_2023}. Specifically, they visualize the attention scores of the input protein sequences as a 2D heatmap, but the aggregation technique is not explicitly mentioned.

\begin{figure}[]
    \centering
    \includegraphics[width=0.7\linewidth]{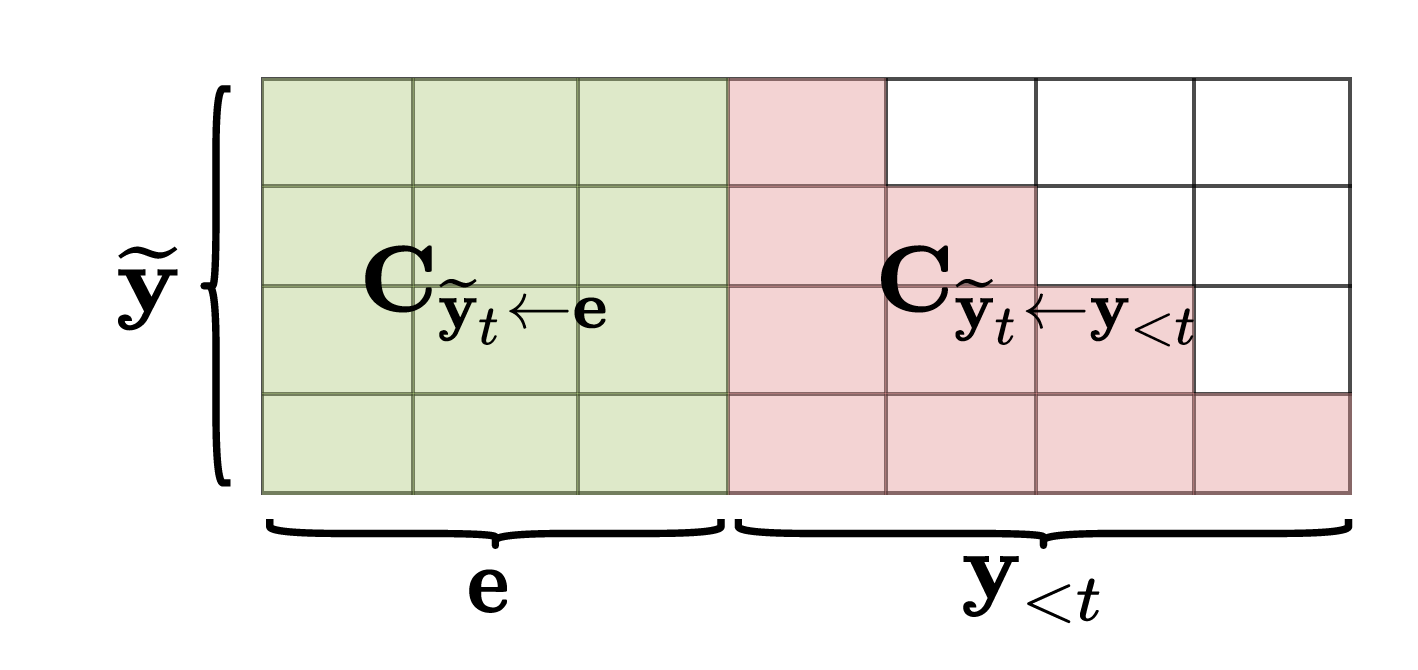}
    \caption{Decoder Layer Contribution Matrix in ALTI+ Method \cite{ferrando_towards_2022}}
    \label{fig:alti}
\end{figure}

The Aggregation of Layer-wise Token-to-token Interactions (ALTI) method, proposed by Ferrando \emph{et al.} \cite{ferrando2022measuring}, calculates the token contribution matrices in transformer encoders in a way similar to attention rollout by forming a directed graph of information flow. However, instead of using raw attention weights, it quantifies contributions using the Manhattan distance between linearly transformed input and output vectors, which in turn is a measure of the information flow. This method was extended to decoder cross-attention in NMT as the ALTI+ method \cite{ferrando_towards_2022}. ALTI+ method decomposes the cross-attention output into two parts: the encoder outputs via the cross-attention mechanism and the contributions of prior decoder outputs via the residual self-attention connection. The final decoder contribution matrix, $[C_{\tilde{y}\leftarrow \mathbf{e}}; C_{\tilde{y}\leftarrow y_{<t}}]$ is then formed by combining the transformed encoder output ($\mathbf{e}$) and residual ($y_{<t}$) contributions, as presented in Figure~\ref{fig:alti}. Here, each row in $\tilde{y}$ represents cross-attention outputs at different decoder time steps.

Other studies process and present attention scores differently from these methods for interpretability. Boito \emph{et al.} used soft-alignment as the interpretation technique for low-resource NMT \cite{boito_investigating_2020}. The attention mechanism used in this case produces a soft-alignment matrix that provides alignment interpretability between the target and source sequences. The proposed attention mechanism works as in equation \ref{eq:boito}.

\begin{equation}
\begin{aligned}
\alpha_{t,j} &= \mathbf{softmax}(\text{align}(h_j, s_{t-1})) \\
c_t &= \text{Att}(H, s_{t-1}) = \sum_{j=1}^{|s|} \alpha_{t,j} h_j
\end{aligned}
\label{eq:boito}
\end{equation}

Here, the soft-alignment probabilities, $\alpha_{t,j}$, are calculated by applying a $softmax$ operation over the alignment scores between the source annotation $h_j$ and the previous decoder state $s_{t-1}$. The context vector, $c_t$, is computed from the attention, which is the weighted sum of source annotations, $h_j$, and the probabilities $\alpha_{t,j}$.

Often, clustering can help with the analysis of attention scores by associating additional meaning with them. For the image captioning task, Ilinykh and Dobnik defined thematic clusters of objects from their textual labels to observe how attention links (attention scores in different heads) in the image encoder form between objects of the same cluster in varying layer depth \cite{ilinykh_what_2021}. A cluster-based interpretability analysis for the target classes is also presented by Kumar and Raman, and Kumar \emph{et al.} \cite{kumar_bert_2022, kumar_towards_2021}. However, it is important to note that in both cases, attention weights were not used for the analysis; rather, they used the last three dense layers from the modular architectures. For text emotion recognition, across the experimented datasets, the dense vectors from the three layers are used to create clusters for each of the emotion classes, which are then used for calculating inter- and intra-cluster distances \cite{kumar_bert_2022}. Similarly, for multimodal speech emotion recognition, principal component analysis (PCA) is used for each layer before creating emotion clusters for analysis \cite{kumar_towards_2021}. Finally, for long-form document matching, although not cluster-based, Jha \emph{et al.} proposed the use of multi-level similarity scores for better interpretability \cite{jha_supervised_2023}. The similarity scores are weighted across different sections and chunks of the text input and are calculated using a combination of contrastive loss and a multi-headed attention layer. 

Li \emph{et al.}, for their Oscar model, discuss the concept of full attention, where tokens from object, vision, and language representations attend to one another, and partial interaction, which restricts attention to only image and language representations \cite{li_oscar_2020}. These are used for the same model through the use of attention masks.

\paragraph{Gradient-based methods}
In contrast to attention maps, which reflect where the model allocates focus during inference, gradient-based attribution methods compute the sensitivity of the output to input features, thus providing class-specific explanations by highlighting input regions that most influence a particular prediction. As most modern deep neural networks are trained via backpropagation, several gradient-based methods have been established for explainability, and are commonly applied in attention-based models \cite{fantozzi2024explainability}. However, among the studies in this SLR, only two gradient-based methods were investigated.

Grad-CAM is one of the more popular gradient-based methods, introduced by Selvaraju \emph{et al.} \cite{selvaraju2017grad}. It was originally introduced for CNNs and involves calculating the average of gradients for each filter until the last convolutional layer, making it architecture agnostic. Among the studies in this review, Xiao \emph{et al.} used Grad-CAM visualization for analyzing important pixels in the histopathological image inputs to their Transformer with Convolution and Graph-Node co-embedding (TCGN) vision backbone \cite{xiao_transformer_2024}. Grad-CAM visualization is generated for different classes for the same input. Due to the graph-based architecture, they also visualized the relationship between patches (nodes). For their dual-channel model for remote scene classification, Yang \emph{et al.} proposed the use of category activation map for both the spatial and frequency channels \cite{yang_explainable_2023}. Although the specific method is not mentioned, the category activation maps are calculated similarly to Grad-CAMs.

The only other gradient-based method used is integrated gradients (IG). IG integrates the gradients of the outputs with respect to the model inputs along a path from some baseline to the inputs. Chiewhawan and Vateekul used IG to determine the positive and negative word predictions in Thai stock market index prediction from the text part of the multimodal input \cite{chiewhawan_explainable_2020}. 

\paragraph{Attention-centric composite methods\label{sec:att-cen}}
While attention scores offer a valuable means of modeling inter-token interactions, they often produce diffused attention maps due to the absence of localized perception. To address this limitation, complementary techniques—such as gradient-based or perturbation-based methods—can be integrated with attention mechanisms for more precise interpretability. A notable approach, called transformer attribution, in this direction was proposed by Chefer \emph{et al.} \cite{chefer2021transformer}, who combined LRP with Grad-CAM-style attributions derived from attention scores to enhance explanation quality. Relevance score, $R^{(k)}$, is assigned through generic Deep Taylor Decomposition to each transformer block, $k$. These scores are then combined with the gradients, $\nabla{M^{(k)}}$, computed with respect to the attention scores $M^{(k)}$ through the Hadamard product. The final relevance map, aR, is produced following equation \ref{eq:chef}.

\begin{equation}
\begin{aligned}
\tilde{M}^{(k)} &= \mathbb{E}_h (\nabla{M^{(k)}} \odot R^{(k)})^+  \\
aR &= \tilde{M}^{(1)} \cdot \tilde{M}^{(2)} \cdots \tilde{M}^{(K)}
\end{aligned}
\label{eq:chef}
\end{equation}

Here, $\mathbb{E}_h$ is the average attention relevance of multiple self-attention heads.

Transformer attribution was used by Du \emph{et al.} to generate interpretations of their transformer-based emotion recognition model \cite{du_case-based_2023}. The relevance scores are calculated by taking the row-wise sum of the relevance maps, which are then also used for the channel selection task. EEG topography of the relevance scores serves as the interpretation technique. For trajectory prediction from multimodal traffic data, Zhang and Li used a modified version of transformer attribution for explaining the Swin-based image transformer \cite{zhang_explainable_2022}. The attribution method is adapted for the regression problem by replacing the predicted class labels with the predicted trajectory with different confidences. In addition, they introduce an up-sampling layer in the transformer model that can capture finer details in the attention maps, and a down-sampling layer to recover the smaller scale weight to be used during training.

Malkiel \emph{et al.} presented BERT Interpretations (BTI), an interpretable text similarity calculation method for BERT models \cite{malkiel_interpreting_2022}. The method combines gradient scores with respect to the similarity score between a pair of paragraphs with corresponding activation maps for calculating per-word saliency scores. The final word-pair similarity between the paragraphs is calculated by multiplying the importance of the words in their respective paragraphs and the similarity between the pair's embedding vectors, as in equation \ref{eq:BTI}.

\begin{equation}
    \begin{aligned}
    Score(w_{p_1}, w_{p_2}) &= Saliency(w_{p_1}) .\\& Saliency(w_{p_2}) . similarity(w_{p_1}, w_{p_2})
    \label{eq:BTI}
    \end{aligned}
\end{equation}

Another technique inspired by Grad-CAM, called Attentive Class Activation Tokens (AttCAT), was presented by Qiang \emph{et al.} \cite{qiang_attcat_nodate}. Yet another unimodal explanation technique, AttCAT first measures the Class Activation Tokens ($CAT_i^l$) for the $l^{th}$ self-attention layer by taking the Hadamard product between the token representations ($h^l_i$) and their corresponding gradients, calculated with respect to the outputs for each input token, $i$. The final AttCAT scores are then calculated by averaging across all heads and summing over all the layers. The formulation is as presented in equation \ref{eq:attcat}.

\begin{equation}
    \text{AttCAT}_i = \sum_{j=1}^L \mathbb{E}_h \left( \alpha_i^j \cdot \text{CAT}_i^j \right)
    \label{eq:attcat}
\end{equation}
Where $\text{CAT}_i^j = \nabla \mathbf{h}_i^j \odot \mathbf{h}_i^j$, and $\mathbb{E}_h(\cdot)$ denotes an averaging operation.

\begin{figure}
    \centering
    \includegraphics[width=1\linewidth, trim={4.5cm 14cm 2cm 6.5cm},clip]{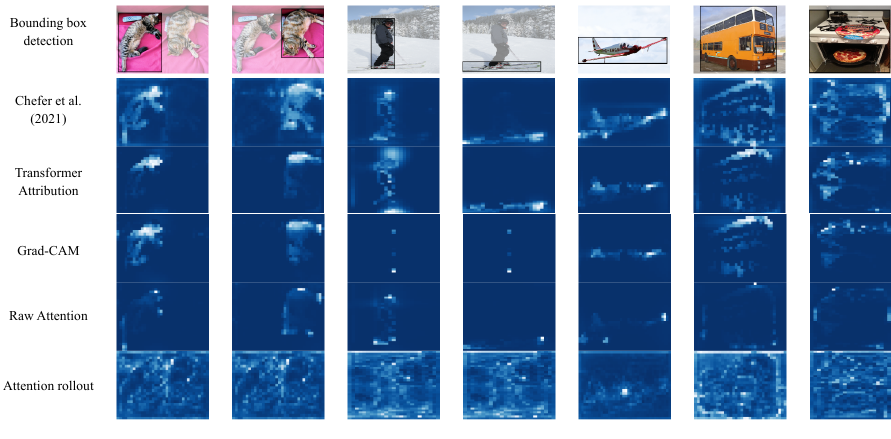}
    \caption{Visual explanation using the multimodal attention-composite method by Chefer \emph{et al.} for object detection, compared with transformer attribution, Grad-CAM, raw attention, and attention rollout \cite{chefer_generic_2021}.}
    \label{fig:chefer_genatt}
\end{figure}

A more multimodal approach, introduced as an extension to the transformer attribution method by Chefer \emph{et al.}, expands relevance-based explanations to both self-attention and cross-attention mechanisms \cite{chefer_generic_2021}.
This method calculates relevance scores separately for self-attention and cross-attention layers. An example of the explanations generated by this method, along with a comparison in the context of object detection, is shown in Figure~\ref{fig:chefer_genatt}.

Initially, the unimodal relevance scores are initialized as identity matrices, while bi-modal (cross-modal) relevance maps are initialized with zeros. The attention map update rule differs depending on whether the interaction occurs within self-attention or cross-attention layers. For self-attention, the relevance aggregation rule accounts for both query-query ($qq$) and key-query ($kq$) token interactions, and follows the update rule presented in equation \ref{eq:chef_sa}.

\begin{equation}
\textbf{R} \leftarrow \textbf{R} + \bar{\textbf{A}} \cdot \textbf{R}
\label{eq:chef_sa}
\end{equation}

Here, $\bar{\textbf{A}}$ denotes the attention map modified by element-wise multiplication with its gradient and clamped to positive values, averaged across attention heads to reflect the contribution of each token.

For cross-attention layers, the update rules handle relevance propagation across modalities. The key-query relevance ($qk$) is initialized with normalization and updated using the equations \ref{eq:chef_ca}.

\begin{align}
\textbf{R}^{qk} &\leftarrow \textbf{R}^{qk} + \left(\bar{\textbf{R}}^{qq}\right)^\top \cdot \bar{\textbf{A}} \cdot \bar{\textbf{R}}^{kk} \\
\textbf{R}^{qq} &\leftarrow \textbf{R}^{qq} + \bar{\textbf{A}} \cdot \textbf{R}^{kq}
\label{eq:chef_ca}
\end{align}

In this context, $\textbf{R}^{qq}$, $\textbf{R}^{qk}$ and $\textbf{R}^{kq}$ represent relevance propagation within modality $q$, from modality $q$ to modality $k$, and from modality $k$ to modality $q$, respectively. $\bar{\textbf{R}}^{qq}$ and $\bar{\textbf{R}}^{kk}$ are normalized relevance matrices, and $\bar{\textbf{A}}$ is defined as:
\begin{equation*}
\bar{\textbf{A}} = \mathbb{E}_h \left( \left( \nabla \textbf{A} \odot \textbf{A} \right)^{+} \right)
\end{equation*}

The update method of relevancy scores assumes all tokens are equally important from the outset, applying uniform accumulation without adjusting for their varying contributions. A variation of this method, proposed by Huang \emph{et al.}, introduces an adaptive weighting strategy based on token attributions for both attention and residual connections \cite{huang_generic_2023}.

Other than these, input perturbations can be effective tools to generate explanations from the resulting changes in the attention weights. Such a method was adopted by Koyama \emph{et al.} to interpret the interactions between T-cell receptors (TCR) and peptide–major histocompatibility complex molecules in their Cross-TCR-interpreter model \cite{koyama_attention_2023}. They perturbed the amino acid residues in positions of interest to observe how that influenced the prediction accuracy and the attention values. In addition, they also analyzed the attention values from all four attention heads in the cross-attention layer. Similarly, Ferrando and Costa-jussà used input perturbation on the source sequence in NMT along with attention norms to generate saliency scores for each layer \cite{ferrando_attention_2021}. These saliency scores represent the contributions of source and target tokens in the prediction of future tokens.

\paragraph{Probing and ablation}
Probing can be designed based on the task or the structure of the model to test the internal representations learned. Ablation, on the other hand, is specific to the model architecture and can incorporate the alteration or removal of specific components to understand their contribution. Although probing can also be used in a model-agnostic context, more commonly in the context of explanations, probing and ablation-based experiments are carefully designed in model-specific ways. Three studies in this review use different probing and ablation studies for interpretability. For effective visio-linguistic representations from the LXMERT, in addition to the previously mentioned adaptive sparse attention, Bhargava used Layerdrop \cite{fan2019reducing} as the regularization mechanism \cite{bhargava_adaptive_2020}. These different elements of the model work as probes to understand how it behaves under different conditions. Firstly, a masking mechanism informs how the span of attention varies for each modality across layers. Then, changing the $\alpha$ parameter's value in $\alpha-entmax$ helps understand preference for sparsity. Finally, Layerdrop regularization demonstrates the trade-off between the compute runtime and the accuracy of the model. Additionally, ablation studies are conducted to conclude that adaptive span works better with denser representations of attention weights from $softmax$ in comparison to $entmax$ and that Layerdrop is effective only up to a certain depth. Overall, these provide a general understanding of how different elements of the model function across modalities and constraints, which then can be used for generating interpretations.

Similarly, Sun \emph{et al.} explored the explainability of the encoding and decoding of their distributed semantic models through probing and ablation-based studies \cite{sun_neural_2021}. These examine how surface-level, syntactic, and semantic features captured by distributed semantic models contribute to modeling cortical responses during sentence processing. In contrast to the other studies, Liang \emph{et al.} proposed a lexical probing method to test the hypothesis that traditional approaches for information retrieval can outperform modern transformer-based solutions in certain cases \cite{hiemstra_using_2021}. The results from the probing experiment are used to determine methods suitable for different kinds of queries, which essentially serves as a rationale for how the resulting routing system works.

\subsubsection{Hybrid}
Compared to attention-centric composite methods like transformer attribution, a more decoupled combination of LRP and attention weights can be found in the work of Zanzotto \emph{et al.} \cite{zanzotto_kermit_2020}. Their proposed hybrid architecture includes Kermit, a non-attention neural layer, that complements the transformer model in syntactic interpretations. To explain the encoding method of Kermit, they used an LRP-driven method called KERMITviz. KERMITviz produces heatmaps from the input syntactic tree, which provides a much better representation of causal relationships than the attention-based BertViz.

A different combination of multiple explainability methods was presented by D. Wang \emph{et al.} \cite{wang_tfregnci_2023} for their multimodal model that fuses chemical properties with electron density data. The dimensionality of the extracted feature vectors is reduced using T-Distributed Stochastic Neighbor Embedding (t-SNE) for visualizing the clusters formed for different binding types. Additionally, they use plots for mean attention distances—distance between features from training samples weighted by attention—across self-attention heads, and compare heatmaps from 2D features and location maps from 3D features. The location maps are retrieved from the last block layers of their TFRegNCI and TFViTNCI models through Grad-RAM, a visualization technique adapted from Grad-CAM for regression models. In another study, Ukwuoma \emph{et al.} used both Grad-CAM for non-attention vision backbones and attention score visualization for their transformer encoder as an explainability method \cite{ukwuoma_hybrid_2023}.

For their explainable QE task in NMT, Treviso \emph{et al.} applied multiple explanation techniques spanning across model-specific and agnostic categories \cite{treviso_ist-unbabel_2021}. They applied attention-based (attention weights, cross-attention weights, attention $\times$ norm), gradient-based (gradient $\times$ hidden states, IG), attention-composite (gradients $\times$  attention), perturbation-based (leave-one-out) techniques as well as Relaxed-Bernoulli rationalizer. Due to the availability of a validation set for the constrained track in the shared task, these methods could be validated against the ground-truth.

\begin{figure}
    \centering
    \includegraphics[width=0.8\linewidth]{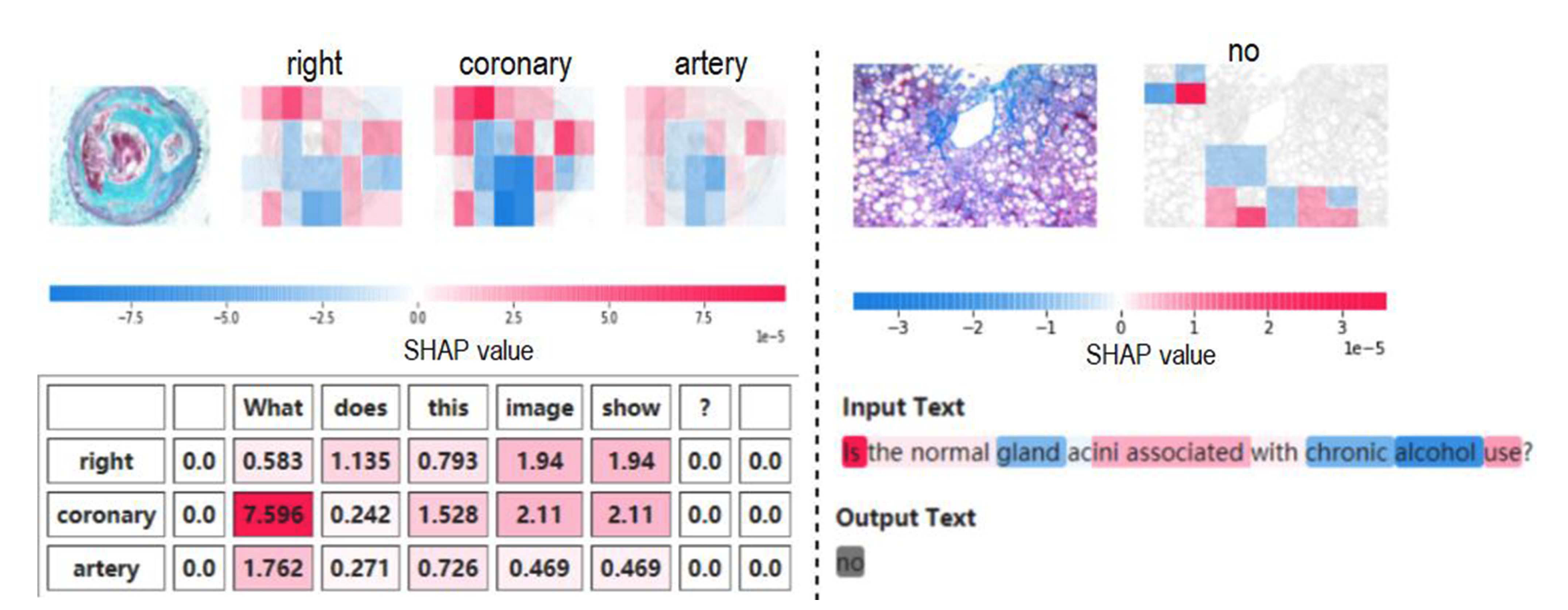}
    \caption{SHAP values to explain pathology images and corresponding questions in VQA tasks \cite{naseem_vision-language_2023}.}
    \label{fig:naseem_SHAP}
\end{figure}

Unlike Treviso \emph{et al.}, Naseem \emph{et al.} implemented different explanation algorithms across categories for explaining different aspects of the decision-making \cite{naseem_vision-language_2023}. SHAP was used for local interpretations for both the pathology input image and the question in the pathology VQA task. SHAP provides SHAP scores of different parts of the inputs, indicating how significant these are for the model's decisions, as depicted in Figure \ref{fig:naseem_SHAP}. In addition, they also compared different CNN-based vision backbones using Grad-CAM visualizations for the same set of input samples. Similarly, for zero-shot matching, in addition to using self-attention maps to understand how the network learns different visual objects, Lin \emph{et al.} used occlusion to observe reductions in similarity scores to find the most influential token pairs \cite{lin_zero-shot_2023}.

\subsection{Self-explaining Models\label{sec:nle}}
An emerging category of explainable models involves systems that are trained to explain their own decision-making processes. This is often achieved through supervised training of an additional decoder that generates high-level textual explanations alongside the original task output. However, since these explanations are also produced by black-box models, their reliability remains questionable. Nevertheless, they offer a key advantage: being high-level, they are more accessible, easier to generate post-training, and can even enhance performance on the original task \cite{hartmann2022survey}. One such work was introduced by Sun \emph{et al.} with their disease detection model from images of leaves \cite{sun_dfyolov5m-m2transformer_2023}. They introduce the $M^2$-transformer with an encoder for encoding regional information of diseases from the images and a text decoder. The $M^2$-transformer uses additional parameters in self-attention for improved memorization of object features and masked meshed cross-attention in the decoder for aggregating insights from different layers and different regions of the image. Two other similar works by S. Wang \emph{et al.} and Chen \emph{et al.} uses natural language explanations along with attention weight visualization as interpretability method \cite{wang_odp-transformer_2023, chen_faster_2023}. S. Wang \emph{et al.} proposed a two-stage model called ODP-Transformer for pest image classification and generating corresponding textual description \cite{wang_odp-transformer_2023}. The first stage encodes different objects into features, which are then used to generate the captions from all the different objects and a classification output. In addition to text-based explanations, the attention weights are visualized for sample inputs to demonstrate the correlation between visual and text modalities. For image captioning, Chen \emph{et al.} \cite{chen_faster_2023} propose the use of the multimodal Oscar model \cite{li_oscar_2020} along with attention score visualization for each token in the generated text across all attention heads.

Self-explaining models can also be useful in sophisticated application use-cases. Parelli \emph{et al.} introduced a reasoning-guided VQA architecture \cite{parelli_interpretable_2023}. The architecture consists of a BERT-based encoder for encoding the question words and a ResNet-based encoder for vision. Final attended-by-the-question representation of the image, $\mathrm{V_q}$, is achieved by taking the weighted sum of the attention weights ($\alpha^Q$) and the image embedding ($\mathcal{F}$) and passing them through a linear layer as in equation \ref{eq:parelli}. Attended-by-the-reasoning representation is also calculated similarly.

 \begin{equation}
     \centering
     \mathrm{V_q} = \mathrm{Linear(}\alpha^Q \odot \mathcal{F}\textit{)}
     \label{eq:parelli}
 \end{equation}
 
Reasoning supervision is achieved by aligning the attention weights conditioned on the questions with the attention weights conditioned on the reasoning. They use a two-stage training process that separates the training from the reasoning distillation, and they use attention-map visualization for explainability. 
 
Guo \emph{et al.} introduce a diffusion-based explainable recommendation model to generate product recommendations along with a corresponding personalized explanation \cite{guo_explainable_2023}. Diffusion and reverse are two processes of the model used for training and inference, respectively. During the diffusion process, noise is incrementally added to the clean data of \textit{user ID}, \textit{item ID}, and \textit{user comment embedding}. A transformer component learns to be robust against the noise for more expressive latent representations. Through the reverse process, the model can be randomized through a Gaussian sample to generate diverse and tailored explanations for recommended items. Lastly, Heo \emph{et al.} used natural language for an audio-visual scene-aware dialogue generation system \cite{heo_natural-language-driven_2023}. The multimodal architecture extracts modality-specific keywords with a GPT-2 \cite{radford2019language} decoder for generating the explanations.

\section{Evaluation Criteria\label{Evaluation Criteria}}

\begin{figure}
    \centering
    \includegraphics[width=1\linewidth]{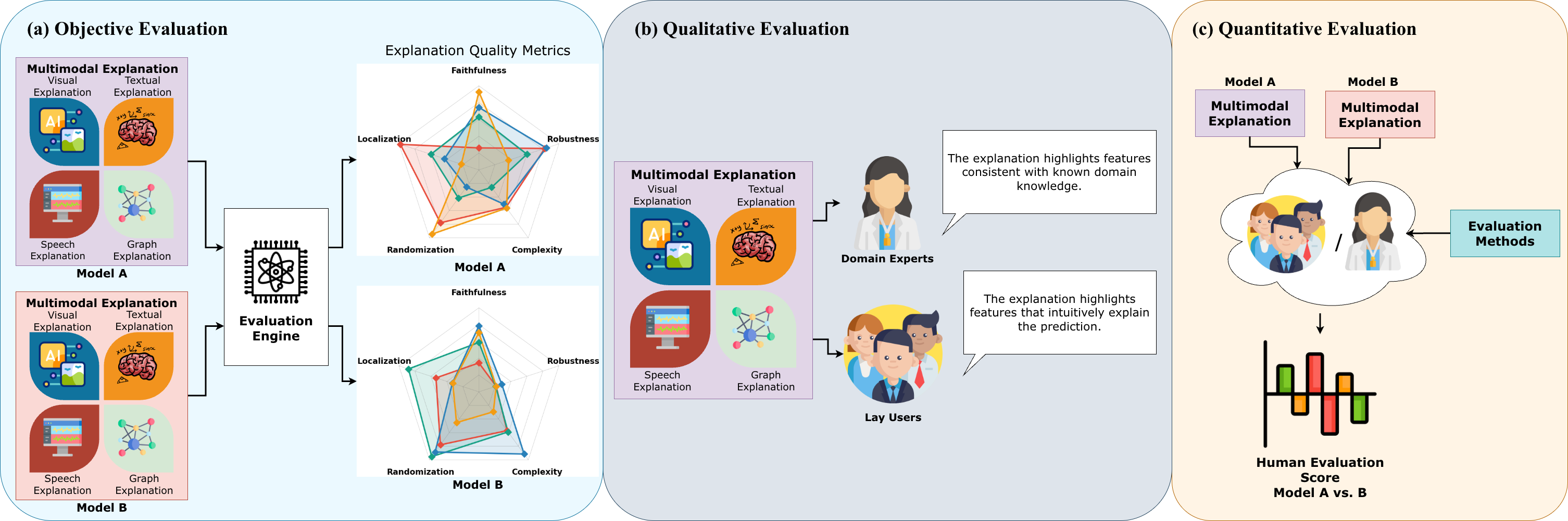}
    \caption{Evaluation paradigms for multimodal explanations.
(a) Objective evaluation assesses explanation quality using automated metrics across six different criteria: faithfulness, robustness, localization, complexity, randomization, and axiomatic.
(b) Human-centered qualitative evaluation assesses whether explanations align with domain knowledge, intuitive reasoning, or are validated using comparative measures.
(c) Human-centered quantitative evaluation compares explanations from different models through structured user studies that collect ratings or preferences.}
    \label{fig:bloc_eval}
\end{figure}
\begin{figure}
\centering
\begin{subfigure}[]{0.6\linewidth}
\centering
\resizebox{\linewidth}{!}{%
\begin{tikzpicture}[mindmap, grow cyclic, every node/.style=concept, concept color=black!10,
    level 1/.append style={level distance=4.5cm, sibling angle=180, font=\footnotesize},
    level 2/.append style={level distance=3.5cm, sibling angle=75, font=\scriptsize},
    level 3/.append style={level distance=2cm, sibling angle=45, font=\tiny},
    ]
\node {Evaluation approaches}
    child [concept color=orange!50] { node {Human-centered Metrics}
        child [concept color=green!30] { node {Qualitative Analysis}
            child { node {Intuitive} }
            child { node {Domain knowledge\\-driven} }
            child { node {Comparis-\\on-driven} }
        }
        child [concept color=red!10] { node {Quantitative Analysis}
            child { node {Human evaluation} }
        }
    }
    child [concept color=orange!50] { node {Objective Metrics}
        child [concept color=yellow!30] { node {Robustness}
            child { node {Stability} }
        }
        child [concept color=teal!40] { node {Localization}
            child { node {Segmen-\\tation Metrics} }
            child { node {Alignment Evaluation} }
        }
        child [concept color=blue!30] { node {Faithfulness}
            child { node {Fidelity} }
            child { node {Classifi-\\cation Metrics} }
            child { node {Ranking Metrics} }
            child { node {Error Metrics} }
            child { node {NLP Metrics} }
            child { node {Confid-\\ence} }
            child { node {Positive perturbation} }
            child { node {Negative perturbation} }
        }
        child [concept color=purple!50] { node {Complexity}
            child { node {Normali-\\zed Entropy} }
        }
    };
\end{tikzpicture}%
}
\caption{Categories and Criteria of Evaluation Methods}
\label{fig:eval_map}
\end{subfigure}
\begin{subfigure}[]{0.4\linewidth}
    \includegraphics[width=\textwidth]{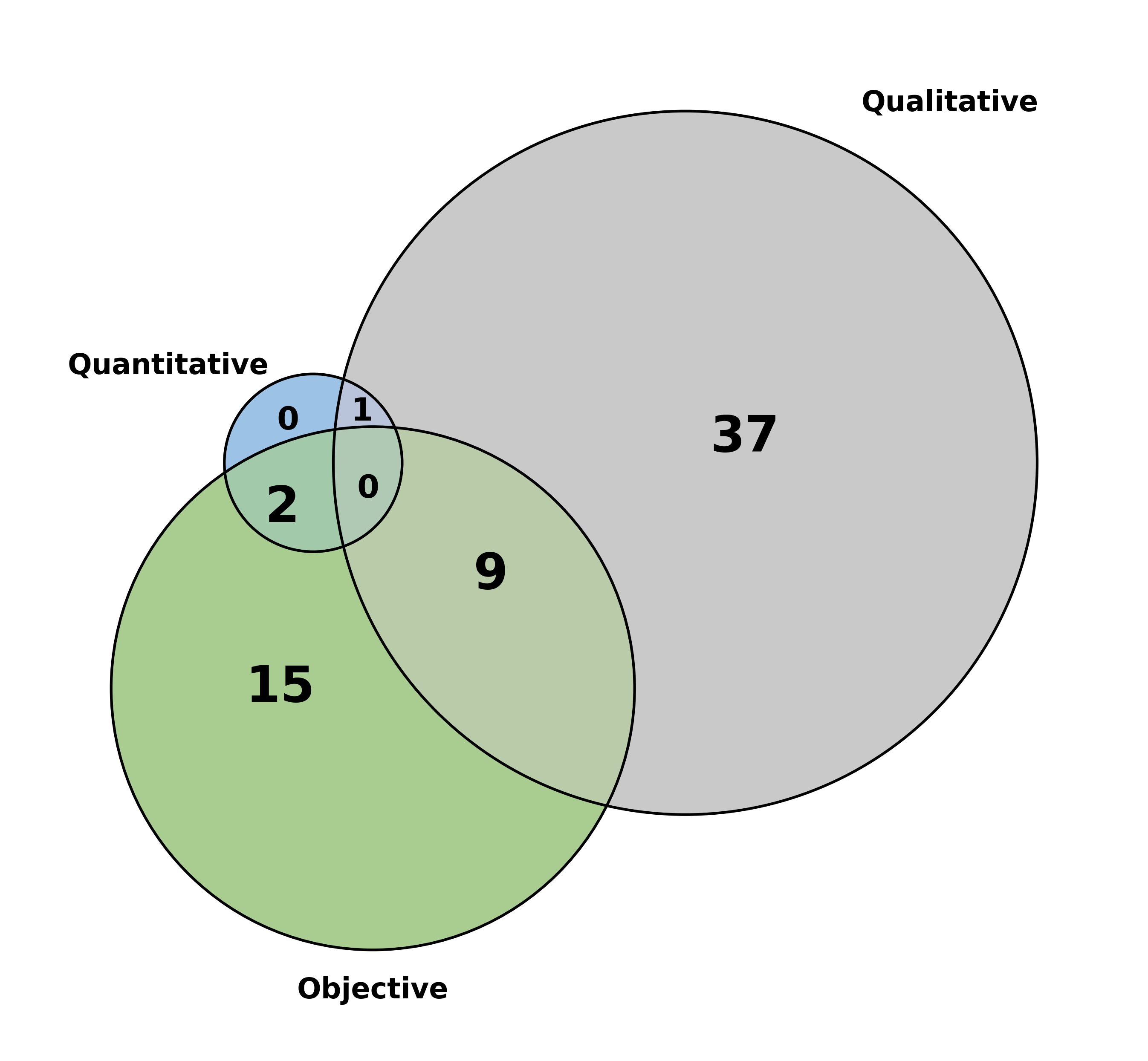}
    \caption{Distribution of Evaluation Metrics across Categories}
    \label{fig:eval_venn}
\end{subfigure}

\caption{Categories of Explanation Evaluation metrics and distribution}
\label{fig:eval}
\end{figure}

The variability in how explanations are evaluated remains a key research gap and a significant barrier to the standardization of evaluation metrics. To address this, Hedström \emph{et al.} introduced Quantus, a toolkit designed to facilitate objective evaluation of explanations \cite{hedstrom2023quantus}. This toolkit categorizes evaluation metrics into six logical groups: faithfulness, robustness, localization, complexity, randomization, and axiomatic metrics. In our framework, we incorporate these six criteria as part of the objective evaluation class within a broader classification structure. This structure is adapted from the evaluation taxonomy proposed by Vilone and Longo \cite{vilone2021notions}. We use this combined hierarchical taxonomy to guide our discussion of the evaluation metrics identified in the reviewed literature. An abstract representation of the evaluation paradigm is presented in Figure~\ref{fig:bloc_eval}. The full categorization, along with corresponding criteria, is illustrated in Figure \ref{fig:eval_map}, and the associated papers are listed in Table \ref{tab:metric}. The distribution of studies across these categories is shown in Figure \ref{fig:eval_venn}.

\begin{table}[t]
\caption{Overview of metric categories, criteria, and associated papers\\
\textbf{Legends: *} Indicates studies that use metrics from different criteria within the same or different categories.}
\label{tab:metric}
\scriptsize
\renewcommand{\arraystretch}{1.25}

\begin{tabular}{p{0.15\linewidth} p{0.11\linewidth} p{0.18\linewidth} p{0.17\linewidth} p{0.26\linewidth} p{0.02\linewidth}}
\toprule
\textbf{Category} & \textbf{Criterion} & \textbf{Definition} & \textbf{Typical Metrics} & \textbf{Representative Papers} & \textbf{\#} \\
\midrule

\multirow{4}{*}{\textbf{Objective}}

& Faithfulness
& Degree to which the explanation reflects the true reasoning of the model.
& Fidelity, Classification, Ranking, Error, Confidence, Perturbation-based
& \cite{janssens_360_2024}*, \cite{wang_exploring_2021}*, \cite{chefer_generic_2021}*, \cite{huang_generic_2023, treviso_ist-unbabel_2021}, \cite{heo_natural-language-driven_2023}*, \cite{ferrando_towards_2022}*, \cite{lin_zero-shot_2023}*, \cite{guo_explainable_2023, qiang_attcat_nodate, rigotti_attention-based_2022, sun_dfyolov5m-m2transformer_2023, chen_faster_2023}, \cite{wang_odp-transformer_2023}*, \cite{kandukuri_physical_2022}, \cite{hiemstra_using_2021}*, \cite{dong_why_2023}*, \cite{li_oscar_2020}*
& 18 \\

& Robustness
& Stability of explanations when the input is slightly perturbed.
& Stability Metric
& \cite{janssens_360_2024}*
& 1 \\

& Localization
& Ability of explanations to highlight relevant input regions or tokens.
& Segmentation, Alignment Evaluation
& \cite{ferrando_attention_2021}*, \cite{buoy_explainable_2023}, \cite{chefer_generic_2021}*, \cite{ferrando_towards_2022}*, \cite{yu_ex-vit_2023}
& 5 \\

& Complexity
& Simplicity and interpretability of explanations.
& Normalized entropy
& \cite{ilinykh_what_2021}*, \cite{boito_investigating_2020}*
& 2 \\

& Randomization
& Stability of explanations when model weights or seeds are randomly altered.
& Model Parameter Randomization Test, Random Logit Test.
& - 
& 0 \\

& Axiomatic
& Satisfaction of formal properties that explanations should theoretically fulfill.
& Completeness, Input Invariance.
& - 
& 0 \\

\midrule

\multirow{4}{*}{\textbf{Human-centered}}

& \multirow{3}{*}{Qualitative}

& Human judgment of explanation clarity and usefulness.
& Intuitiveness / interpretability assessment
& \cite{yang_explainable_2023, chiewhawan_explainable_2020, ullah_explainable_2022, zhang_explainable_2022, mohammadkhani_explaining_2023, wu_interpretable_2023, zanzotto_kermit_2020, xu_logiformer_2022, xu_multi-granular_2020, che_multiscale_2023, ukwuoma_hybrid_2023, wang_tfregnci_2023, kumar_towards_2021, naseem_vision-language_2023}, \cite{ilinykh_what_2021}*, \cite{dong_why_2023}*
& \\

& & Alignment with domain expertise or expected reasoning.
& Domain knowledge validation
& \cite{du_case-based_2023, wang_novel_2023, koyama_attention_2023, abdulkadir_augmenting_2022, meng_bidirectional_2021, ding_deepstf_2023, zhang_explainable_2022}, \cite{wang_odp-transformer_2023}*, \cite{huang_representation_2023, jha_supervised_2023, xiao_transformer_2024, naseem_vision-language_2023}, \cite{li_oscar_2020}*
& 47 \\

& & Analysis via ablation, probing, or baseline comparisons.
& Comparison-based analysis
& \cite{kumar_bert_2022, bhargava_adaptive_2020, koyama_attention_2023}, \cite{ferrando_attention_2021}*, \cite{mohammadkhani_explaining_2023, parelli_interpretable_2023}, \cite{malkiel_interpreting_2022}*, \cite{boito_investigating_2020}*, \cite{xu_multi-granular_2020, che_multiscale_2023, sun_neural_2021, ukwuoma_hybrid_2023, wang_tfregnci_2023}, \cite{ferrando_towards_2022}*, \cite{kumar_towards_2021}, \cite{hiemstra_using_2021}*, \cite{naseem_vision-language_2023}, \cite{lin_zero-shot_2023}*
& \\

\cline{2-6}

& Quantitative
& Measured user performance or ratings when using explanations.
& User studies, rating scores, task performance
& \cite{wang_exploring_2021}*, \cite{malkiel_interpreting_2022}*, \cite{heo_natural-language-driven_2023}*
& 3 \\

\midrule

\multicolumn{2}{l}{\textbf{Non-grouped Metrics}}
& \multicolumn{2}{l}{Mixed or unspecified evaluation approaches.}
& \cite{kumar_bert_2022, ferrando_attention_2021, rigotti_attention-based_2022, ilinykh_what_2021, li_oscar_2020}
& 5 \\

\bottomrule
\addlinespace[2pt]
\multicolumn{6}{l}{No papers in this review implemented Randomization or Axiomatic Criteria; listed metrics are general in the literature.}
\end{tabular}

\end{table}

\subsection{Objective Metrics\label{sec:quant}}
Let $f : \mathcal{X} \rightarrow \mathbb{R}$ denote a predictive model,
$x \in \mathcal{X}$ an input instance with $d$ features, and
$E(x) = (e_1,\dots,e_d) \in \mathbb{R}^d$ the attribution vector
produced by an explanation method $E$. Let $\mathcal{D}=\{x_i\}_{i=1}^{N}$ denote
a dataset of size $N$.

The following subsections describe objective evaluation metrics based on the
Quantus taxonomy observed in this review. The notation introduced above is used consistently across all criteria.

\subsubsection{Faithfulness}
One of the key criteria for an explanation method is how well the explanation reflects the model's decision-making. This means, for example, the important features as presented in the explanations are in fact important for the model. Many faithfulness metrics therefore
measure the agreement between feature importance scores and the change
in model output when those features are perturbed.

Let $\mathcal{P}$ denote a perturbation operator that modifies a subset of features of the input, and let $S \subseteq \{1,\dots,d\}$ be a subset of features.

A general faithfulness metric can be defined as in Equation~\ref{eq:faith}.

\begin{equation}
\centering
   FA(E) =
    \frac{1}{N}\sum_{i=1}^{N}
    \Phi\big(
    E(x),
    f(x),
    f(\mathcal{P}(x,S))
    \big)
\label{eq:faith}    
\end{equation}

where $\Phi(\cdot)$ is a comparison functional that measures agreement between
attribution scores and the change in model output induced by perturbations.

Different faithfulness metrics instantiate this formulation using different
choices of perturbation operators and comparison functions. The literature quantifies faithfulness in several ways, which can broadly be grouped into fidelity-based metrics, classification or
ranking metrics, error-based metrics, NLP metrics, confidence metrics and perturbation-based metrics.

\paragraph{Fidelity/faithfulness}: A concrete example of using faithfulness for assessing interpretability was presented by Janssens \emph{et al.} \cite{janssens_360_2024}. They trained a ridge regression surrogate model, $g$, on the interpretable representation generated by the LIME algorithm. These perturbed inputs are also used as input to the original model. Then, the coefficient of determination, $R^2$, is used to quantify the fidelity or how close the surrogate model's predictions are to the black-box model, hence the local interpretability. The average between the fidelity scores over $N_k$ observations serves as the overall fidelity, as presented in the equation \ref{eq:fid}.

\begin{equation}
   \text{Fidelity} = \frac{1}{K} \sum_{j=1}^{K} \left( \frac{\sum_{i=1}^{N_k} R_i^2}{N_k} \right)
    \label{eq:fid} 
\end{equation}

Here, $K$ represents the total number of folds of the observations.

Faithfulness was also defined as one of the target explainable properties in the Concept Transformer \cite{rigotti_attention-based_2022}. According to equation \ref{eq:ct}, the conditional probability of each output class in the Concept Transformer is dependent on the input through the average contribution of attention scores, referred to as the positive relevance scores. Hence, these positive relevance scores are essentially how the different ideas of concepts influence the output, making the model faithful.

\paragraph{Classification/ranking metrics} Classification or ranking metrics are generally metrics for evaluating the performance of models. However, when trained towards a ground-truth set of explanations, the explanations can also be evaluated using these metrics, which would then represent the faithfulness of the model. For example, in the EVAL4NLP Shared Task on QE, both in the constrained and unconstrained tracks, Treviso \emph{et al.} \cite{treviso_ist-unbabel_2021} evaluated their models using Area Under the Curve (AUC), Average Precision (AP), and Recall at Top-K (R@K). These metrics were computed against word-level ground-truth labels, restricted to the subset of tokens containing translation errors. F1-score was used in addition to other metrics for evaluating the reason-induced autonomous driving system by Dong \emph{et al.} \cite{dong_why_2023}. They report the overall F1 score and mean in-class F1 score for the reasoning text generated by the model, calculated against the ground-truth. The six selected reason labels trained for result in a very limited vocabulary; hence, the reported F1 scores can reach very high values for all the tested models.

Among the ranking metrics, Liang \emph{et al.} used the mean reciprocal rank (MRR) score \cite{voorhees1999trec} to rank evidence for their QA task \cite{hiemstra_using_2021}. For a ranked list, the MRR metric highlights how highly ranked the relevant results are for a set of queries. The formulation is as presented in equation \ref{eq:mrr}.

\begin{equation}
    MRR = \frac{1}{|Q|}\sum^{|Q|}_{i=1}{\frac{1}{rank_i}}
    \label{eq:mrr}
\end{equation}

Here, $Q$ represents the question queries.

\paragraph{Error metrics} When model parameters are inherently interpretable, the training loss reflects not only the model's ability to approximate the input-output relationship but also the degree to which the learned parameter values align with plausible representations of the underlying data. A significant training loss in such cases may indicate a mismatch between the model's learned structure and the inherent relationships within the data. An example of this is the work by Kandukuri \emph{et al.} \cite{kandukuri_physical_2022}. The differentiable physics layer is trained in both supervised and self-supervised fashion. The supervised loss is calculated from the error in pose estimation using equation \ref{eq:sl}.

\begin{equation}
L_{\text{supervised}} = \sum_i e(p_{1:N, i}^{gt}, p_{1:N, i}^{enc}) + \alpha\, e(p_{1:N, i}^{gt}, \hat{p}_{1:N, i})
\label{eq:sl}
\end{equation}

Here, $p_{1:N, i}^{enc}$ represents the inferred pose for each object $i$, which the physics engine is initiated with, and $\hat{p}_{1:N, i}$ represents the estimated pose. Both of these are compared with the ground-truth pose $p_{1:N, i}^{gt}$ using a pose estimation error function, $e$, calculated from the rotational error (quaternion geodesic norm) and translational error. The self-supervised loss, on the other hand, reconstructs the poses in images to calculate the reconstruction loss.

\paragraph{NLP Metric}
Self-explaining models are trained in a supervised fashion against ground-truth natural language explanations. Hence, commonly used metrics for natural language generation tasks can be directly applied to these models to evaluate the faithfulness. The set of the metrics used include BLEU \cite{papineni2002bleu}, ROUGE \cite{lin2004rouge}, CIDEr \cite{vedantam2015cider}, SPICE \cite{anderson2016spice} and METEOR \cite{denkowski2014meteor}. BLEU and CIDEr are n-gram-based matching techniques, with BLEU having penalization for brevity and CIDEr having additional weighing by TF-IDF values. Like BLEU, METEOR was also a metric originally proposed for measuring the quality of translations. It uses a harmonic mean of recall in addition to precision and incorporates consideration for synonyms via a thesaurus. ROUGE evaluates the recall between source and reference text, calculated based on the overlap within lexical elements. In contrast to others, SPICE is a multimodal evaluation metric used in image captioning where scene graphs are generated based on the scene and the caption, and uses a graph similarity measure to determine semantic accuracy.

In various tasks, the ROUGE metric may penalize high-quality outputs when appropriate or matching reference texts are absent or poorly aligned. Therefore, it is primarily used for text summarization-specific evaluations \cite{wang_exploring_2021}. In contrast, many studies that utilize NLP metrics tend to rely on BLEU scores, either independently \cite{ferrando_towards_2022, dong_why_2023} or in combination with other NLP metrics \cite{wang_odp-transformer_2023, heo_natural-language-driven_2023, huang_generic_2023, chen_faster_2023, sun_dfyolov5m-m2transformer_2023, guo_explainable_2023}. BLEU has become a standard metric in the field of NLP due to its efficient and fast calculation capabilities, especially for comparisons at the corpus level. However, its reliance on surface-level n-gram overlaps means it often fails to assess context or fluency effectively, leading to poor performance on individual sentences \cite{chen2015microsoft}. As a result, in tasks like image captioning, it is commonly paired with other techniques to address these limitations.

\paragraph{Confidence}
Confidence defines the exclusiveness in terms of the impact of explanations for each predicted class. Qiang \emph{et al.} defined confidence as one of the desired properties of explanation \cite{qiang_attcat_nodate}. The confidence of explanations per class was calculated using Kendall-$\tau$ correlation, which statistically measures the ordinal association between the saliency for classes. So, a lower Kendall correlation implies more confidence in explaining the predicted class. For a ranked array of per-token saliency, $S(x)$, the Kendall correlation for class $c$ is calculated using equation \ref{eq:kendall}.

\begin{equation}
    \text{Kendall correlation} = \frac{1}{N} \sum_{i=1}^{N} \text{Kendall-}\tau \left( S(x_i)_c, S(x_i)_{C \setminus c} \right)
    \label{eq:kendall}
\end{equation}
Here, $S(x_i)_c$ represents saliency scores for class $c$ and $S(x_i)_{C \setminus c}$ is the scores for classes other than $c$.

This statistical measure gives a sense of how confidently explanation methods assign saliency scores for predicting each class. Although Qiang \emph{et al.} explore confidence as a separate metric from faithfulness \cite{qiang_attcat_nodate}, confidence also describes the correlation between the accuracy of explanations and the predictive process. This is done by ensuring highly important tokens for one of the predicted classes have low importance for others.

\paragraph{Positive and Negative Perturbation}
Input perturbation based on the feature importance ranks can be used to observe how the performance of the model changes accordingly. This can be done by masking features (e.g., pixels, tokens) gradually from most important to least for positive perturbation, and observing the rate of change in performance. In case of negative perturbation, a similar process can be followed except masking features from the least important to the most important ones. 

Input perturbation based on feature-importance rankings can be viewed as a concrete instantiation of the general faithfulness formulation in Equation~\ref{eq:faith}. In this setting, the perturbation operator $\mathcal{P}(x, S)$ masks a subset of input features $S$, where $S$ is constructed according to the ranking induced by the explanation $E(x)$.

Ideally, positive perturbation should result in a sharp drop in performance, which signifies that the most important features, according to the explanations, are also important for decision-making by the model. Negative perturbation should result in stable performance. Positive perturbation can reflect the sensitivity side of faithfulness, whereas negative perturbation represents the robustness side \cite{huang_generic_2023}.

In studies by Chefer \emph{et al.} \cite{chefer_generic_2021} and Huang \emph{et al.} \cite{huang_generic_2023}, both employed positive and negative perturbation analyses for each modality in multimodal models, using the AUC as the primary evaluation metric. AUC was computed by progressively masking input tokens or image pixels in order of relevance and measuring the model's performance at each masking level. In addition to reporting accuracy-based AUC curves, Huang \emph{et al.} also monitored NLP metrics (BLEU, METEOR, CIDEr) for the captions in their image captioning task.

\subsubsection{Robustness}
Robustness of the explanations reflects how stable they are against slight changes to the model inputs, given that the output of the model remains similar. Robustness very importantly determines the consistency in explanations for similar samples, or that similar explanations result in the prediction of the same class. Expanding the metric for multimodal cases requires perturbations or counterfactual editing for all the different combinations (e.g., only one, every pair, or all) of modalities.

Let $\delta$ denote a small perturbation sampled from
a perturbation distribution (e.g., Gaussian noise) such that
$\|\delta\|_{p} \leq \epsilon$. The robustness of explanations can be
measured as the similarity between the explanations of the original and
perturbed inputs as in equation~\ref{eq:robu}.

\begin{equation}
    RO(E) =
\frac{1}{N}\sum_{i=1}^{N}
\mathrm{sim}\big(E(x_i), E(x_i + \delta_i)\big)
\label{eq:robu}
\end{equation}

where $\mathrm{sim}(\cdot,\cdot)$ is a similarity measure such as
Pearson correlation, cosine similarity, or rank correlation. Higher
values indicate that explanations remain stable under small perturbations
of the input.

\paragraph{Stability}
The term stability is often used interchangeably to represent robustness \cite{carvalho2019machine} to refer to the explanations holding up to slight perturbations to input instances. Janssens \emph{et al.} describe stability as one of the most important explanation properties and adopt it for their rumor detection task \cite{janssens_360_2024}. They propose a new metric based on the pairwise Jaccard coefficient for calculating the global stability of feature rankings across different folds in cross-fold validation. For feature ranking sets, $F_{f_1}$, and $F_{f_2}$, from folds, $f_1$, and $f_2$, respectively, the stability across $K$ folds and upto top-$k$ features are calculated as in equation \ref{eq:jaccard}.

\begin{align}
\text{Stability}_k &= \frac{\sum\limits_{f_1 < f_2}^{K} J(F_{f_1}, F_{f_2})}{\binom{K}{2}} \notag \\
&= \frac{\sum\limits_{f_1 < f_2}^{K} \frac{|F_{f_1} \cap F_{f_2}|}{|F_{f_1} \cup F_{f_2}|}}{\frac{K!}{2!(K-2)!}} 
\label{eq:jaccard}
\end{align}

A higher value would imply more agreement on the feature importance values across different folds, resulting in more stable explanations. Stability is important to promote trust in the explanations.

\subsubsection{Localization}
Localization metrics assess the degree to which explanations are confined within a specified ROI. These metrics are particularly well-suited for multidimensional input spaces, such as images, and can incorporate reference ROIs defined through bounding boxes, segmentation masks, and similar methods. For models that generate two-dimensional heatmaps, localization serves as an effective measure to evaluate the extent to which saliency scores are concentrated around the original positions of target objects.

Let $G(x) \in \{0,1\}^{d}$ denote a ground-truth relevance mask identifying
the truly relevant features. Localization can be measured using equation~\ref{eq:loca}, with a similarity function $\mathrm{sim}(\cdot,\cdot)$, such as Intersection-over-Union (IoU).

\begin{equation}
    LO(E) =
\frac{1}{N}\sum_{i=1}^{N}
\mathrm{sim}\big(E(x_i), G(x_i)\big)
\label{eq:loca}
\end{equation}

Higher values indicate better agreement between the explanation and the
ground-truth relevant regions.

\paragraph{Segmentation Metrics}
Tools and metrics commonly used in tasks such as object detection or semantic segmentation are inherently driven by localization, and follow equation~\ref{eq:loca} with varying similarity functions. Segmentation metrics found throughout this review include mean IoU, AP, and Average Recall (AR). Mean IoU scores define how well the pixels in the heatmaps are aligned to ground-truth objects and can be used in weakly supervised semantic segmentation studies \cite{yu_ex-vit_2023}. AP and AR require the confidence scores of prediction, hence, are suitable for validating object-level heatmaps \cite{chefer_generic_2021}.

\paragraph{Alignment Evaluation Metrics}
Alignment evaluation metric is similar to segmentation metrics and is used in both image and text localization problems. For scene text recognition, Buoy \emph{et al.} proposed alignment evaluation by checking for overlap between the character region as indicated by the AM, $R_k$, and the ground-truth bounding box, $G_k$ \cite{buoy_explainable_2023}. They propose the equation \ref{eq:aem} for calculating the final evaluation score for a text with length $l$.

\begin{equation}
\begin{aligned}
\text{AEM}_k &= 
\begin{cases}
1, & \text{if } R_k \cap G_k \ne 0 \\
0, & \text{otherwise}
\end{cases} \\
\text{AEM}_{\text{TEXT}} &= \frac{\sum_{k=1}^{l} \text{AEM}_k}{l}
\end{aligned}
\label{eq:aem}
\end{equation}

For comparing the more diffused cross-attention maps to the AMs, a threshold, $\beta$, is used.

Alignments are also used in a different context in NLP, specifically in NMT. Calculated similarly, Ferrando \emph{et al.} retrieved the alignments in bilingual NMT from cross-attention layers using their ALTI+ method \cite{ferrando_towards_2022}. These alignments were then validated against human-annotated gold alignments for calculating the alignment error rate (AER). AER helps evaluate how well the alignments are localized with respect to the reference.

\subsubsection{Complexity}
Complexity is the evaluation of the conciseness of explanations, which means features representative in the explanation are also the most important for prediction. A complex explanation may deem several features important, making it less accessible. This is a particularly important metric in multimodal use cases as the complexity of explanations gets aggregated.

A simple measure of explanation complexity is the number of
features receiving non-zero attribution. This can be quantified using the
$\ell_0$ norm, as in equation~\ref{eq:comp}.

\begin{equation}
    CO(E) =
\frac{1}{N}\sum_{i=1}^{N}
\|E(x_i)\|_0
\label{eq:comp}
\end{equation}

where $\|\cdot\|_0$ counts the number of non-zero elements in the attribution
vector. Lower values correspond to sparser and therefore simpler
explanations.

\paragraph{Normalized entropy}
An alternative way to measure explanation complexity is to quantify the
dispersion of importance weights using entropy. For instance, Ilinykh and Dobnik used normalized entropy to identify the level of dispersion across the different attention heads and layers \cite{ilinykh_what_2021}. Given the source object, $s_i$, and the target object, $t_j$, in the thematic clusters and the attention distribution function, $\alpha$, the entropy ($H$) is calculated as in equation \ref{eq:ne}. The value of entropy is measured for each attention head, $h$, across $\ell$ layers. 

\begin{equation}
    H_{\alpha_{\ell, h}}(t_j) = -\sum_{i=1}^{|S|} \alpha(s_i, t_j) \log\left( \alpha(s_i, t_j) \right)
    \label{eq:ne}
\end{equation}

This entropy is then normalized by dividing by the maximum possible entropy, $\log|S|$, to ensure comparability across different configurations. Higher entropy indicates a more dispersed distribution of attention weights, reflecting a higher complexity in the explanation. This suggests that the model distributes attention across multiple source-target pairs, implying a richer, more intricate relationship structure. On the other hand, lower entropy indicates saturated attention links, where a single or a few source objects dominate the attention distribution. This corresponds to a simpler, less complex explanation, as the model focuses on a narrow set of relationships.

Similarly, Boito \emph{et al.} used average normalized entropy (ANE) to quantify the soft-alignment quality in sequence-to-sequence tasks \cite{boito_investigating_2020}. Here, they calculated the entropy between the source and target sentence pairs at different granularities and averaged over the tokens in the target sequence.

\subsubsection{Other Quantus Categories}

The remaining two categories in the Quantus taxonomy—Randomization and Axiomatic—were not utilized in any of the studies included in this review. Randomization metrics measure the rate of change in explanations as the model parameters or data labels are increasingly randomized. 

Let $f_{\mathrm{rand}}$ denote a randomized version of the model, obtained
for example by reinitializing parameters or progressively randomizing layers. The corresponding explanation is denoted by $E_{f_{\mathrm{rand}}}(x)$.

A commonly used formulation measures the dissimilarity between
explanations generated from the original and randomized models as in equation~\ref{eq:rand}.

\begin{equation}
    RA(E) =
\frac{1}{N}\sum_{i=1}^{N}
\left(
1 -
\mathrm{sim}\big(
E_f(x_i),
E_{f_{\mathrm{rand}}}(x_i)
\big)
\right)
\label{eq:rand}
\end{equation}

Higher values indicate that explanations change substantially when the model is randomized, which suggests that the explanation method depends on the learned model parameters.

Different randomization-based metrics vary in the randomization strategy
(e.g., layer-wise parameter randomization or label randomization) and in
how explanation degradation is aggregated across steps.

Axiomatic methods, on the other hand, evaluate whether the explanation satisfies certain formal properties. Such properties are typically defined as mathematical axioms that explanations should fulfill, including completeness, sensitivity, implementation invariance, or symmetry.

For example, axiomatic evaluation tests satisfying the completeness property can be formulated as in equation~\ref{eq:axio}. 

Given a baseline input $x^{\mathrm{base}}$, completeness requires that the sum of feature attributions equals the difference between the model output for the input and the baseline:

\[
f(x) - f(x^{\mathrm{base}}) =
\sum_{j=1}^{d} e_j .
\]

Deviations from this equality can be measured as a representation of the axiomatic score.

\begin{equation}
    A(E) =
\frac{1}{N}\sum_{i=1}^{N}
\left|
f(x_i) - f(x_i^{\mathrm{base}})
-
\sum_{j=1}^{d} E_j(x_i)
\right|
\label{eq:axio}
\end{equation}

Lower values indicate better satisfaction of the completeness axiom.
Other axioms can be evaluated using similar formulations depending on
the property being tested.

Randomization is a complex method, requiring the selection of the target for perturbation, the rate of randomization, and metrics that can meaningfully capture degradation in explanation quality across heterogeneous modalities. Similarly, most existing axiomatic frameworks are designed for unimodal contexts and rely on assumptions that may not generalize to multimodal architectures. Their extension often depends on modality-specific factors, such as the fusion strategy and the nature of modalities involved. Consequently, while both categories represent theoretically important directions, their practical application in the multimodal domain remains underexplored. We discuss two example methods from these categories in Section \ref{sec:recent_eval}.

\subsection{Human-centered Metrics}
A very effective way to determine the quality of explanations is to compare how accurately the inference process is depicted in the mental representations of the end-users. This can be systematic user studies evaluated using quantitative methods incorporating well-defined criteria. More commonly, though, non-systematic experiments along with qualitative metrics are used.

Human-centered metrics, although not as causally reliable or robust as the objective ones, are relatively easy to use and well-suited for capturing the subjective aspects of explanatory quality. In the literature, how these methods are implemented varies based on whether they are non-systematic and require human interpretations or systematic and incorporate user studies.

\subsubsection{Qualitative Metrics}
Qualitative analysis of experiments is driven by the complications involved in producing experiments for systematic evaluation and the lack of standardization in user perspectives. These often involve generic applications analyzed from intuition or for specialized tasks analyzed by domain experts. As a result, this kind of qualitative analysis is mostly done when explanations are represented at a high level, e.g., feature importance plot or heatmaps.

\paragraph{Intuitive analysis}
High-level interpretations of general tasks, such as sentiment analysis or natural scene image classification, can be relatively accessible to lay users. For some sample inputs, the generated explanations are compared to commonly known patterns. For example, for the remote scene classification task by Yang \emph{et al.}, the visualized category activation maps for different scenes are analyzed to observe which of the input channels approximated textures better \cite{yang_explainable_2023}. Zanzotto \emph{et al.} analyzed the interpretability of the Kermit model by generating explanations for carefully chosen test samples with underlying causal patterns, which the tree-like structure of Kermit encodes much better \cite{zanzotto_kermit_2020}. Few other studies also follow a similar method of evaluation \cite{xu_logiformer_2022}.

In some cases, the underlying intuition behind model explanations is presented without adequate accompanying analysis. For instance, Wu \emph{et al.} presented the past and future features important for forecasting tourism demand across the three cities from the used dataset \cite{wu_interpretable_2023}. Although how the interpretations can be used is discussed, comprehensive validations of how features important for the model are relevant to the domain are not added. Similarly, for Thai stock market prediction, Chiewhawan and Vateekul presented the 10  most significant positive and negative words from the test data with no validation \cite{chiewhawan_explainable_2020}. Similarly, Ullah \emph{et al.} used explanations to figure out the top contributing features for malware detection \cite{ullah_explainable_2022}.

In addition to these different methods, intuitive analysis can be done on the results of some experiments to generate comparative explanations. For instance, S. Xu \emph{et al.} compare the attention matrix produced to explain the multi-granular BERT for two cases—with BERT-base by summing the attention scores for the components of the n-grams and with the same model initialized with weights from a baseline model \cite{xu_multi-granular_2020}. The results of these experiments are then analyzed based on how intuitive the matrices were, their sparsity, and the attention they produced on non-relevant words. A similar method can be found in other studies \cite{che_multiscale_2023, ukwuoma_hybrid_2023, wang_tfregnci_2023, kumar_towards_2021}.

\paragraph{Domain knowledge-driven analysis}
For more specialized domains such as medical imaging, the analysis of explanations requires deeper knowledge of the task. This might involve comparing to insights or facts established within the domain to assess the plausibility of the explanations. Such an analysis is presented by Du \emph{et al.}, where they compare the EEG topographies with insights from neuroscience on the origins of emotion as a validation method \cite{du_case-based_2023}. Similarly, Zheng \emph{et al.} \cite{wang_novel_2023} visually compared the 2D heatmap from protein sequences with the true binding site in the same protein. Commonly, analysis of explanations requires domain expertise in healthcare-related applications \cite{abdulkadir_augmenting_2022, meng_bidirectional_2021, huang_representation_2023, naseem_vision-language_2023} or applications in computational biology \cite{xiao_transformer_2024, ding_deepstf_2023, koyama_attention_2023}. One of the more general applications of this method includes the document matching work by Jha \emph{et al.} \cite{jha_supervised_2023}. The authors selected two related documents and a non-related document and analyzed explanations for matching at different levels of detail, which is information known in advance.

\paragraph{Comparison-driven analysis}
This group of studies typically uses case-based experiments to add credibility to the evaluation of explanations. While some papers in this area use objective metrics (as discussed in Section \ref{sec:quant}), they often fail to systematically measure the explanations. As a result, these approaches may rely on supplementary human-centered methods like expert reviews or intuitive analysis. For example, in their study on code models, Mohammadkhani \emph{et al.} split the input code into different types of tokens to better understand the average attention scores across all model layers for each group \cite{mohammadkhani_explaining_2023}. They compared model predictions to gold references to classify how complex the code is, looking for patterns in specific code metrics and comparing them to trends in the dataset, such as the number of tokens or variables. Studies that use probing and ablation methods for interpretability often use this kind of analysis. In the study by Sun \emph{et al.}, different statistical analyses and comparisons were used for determining the relative behavior of different DSM models \cite{sun_neural_2021}. For example, to explain the neural encoding with DSMs, they find the correlation between linguistic features and encoding accuracy for different regions of interest (ROI) from large-scale brain networks. They do further analysis on the pairwise matching accuracies, accuracy of the model in matching brain responses to the corresponding stimuli, for the same scenario with different ROI topics. Similarly, to understand how adaptive mechanisms reflect on the learning of multimodal representations, Bhargava evaluated different configurations of their sparse attention-based method against the baseline through test set (test-dev and test-std) accuracy \cite{bhargava_adaptive_2020}. These methods observe model behaviors and establish the constraints under which these emerge.

Such techniques are also used in novel studies revolved around explainability, such as identifying the implications of training self-explaining models. For instance, Parelli \emph{et al.} quantitatively compared the test set accuracy of their VQA models before and after aligning to the reasoning \cite{parelli_interpretable_2023}. In addition, they also used ablation accuracies to observe the drop in accuracy after masking key visual features in both cases. In their image retrieval work, Lin \emph{et al.} used experiments like finding key matching-feature pairs by observing the maximum drop in relation score in individual matches \cite{lin_zero-shot_2023}.

\subsubsection{Quantitative metrics - Human evaluations/user studies}
Rooted in qualitative judgments, quantitative human evaluations address both the subjectivity of explanations and provide a systematic framework for comparing different models based on that. These incorporate the two categories of evaluation approaches proposed by Doshi-Velez and Kim—application-guided and human-guided evaluation \cite{doshi2017towards}. With certain bias, such studies can be used to compare explanations across different modalities, various forms of explanations (e.g., high-level vs. low-level), and across multiple criteria \cite{mohseni2021multidisciplinary}. However, one very clear issue with these is the lack of a standard protocol followed for such studies \cite{mohseni2021multidisciplinary, vilone2021classification}. Although a commonly used form of experiment, in this review, only three publications report the use of human evaluations. 

For their abstractive summarization work, H. Wang \emph{et al.} proposed human evaluation for two different tasks \cite{wang_exploring_2021}. The subjects are lay users from the Amazon Mechanical Turk platform. In one experiment, question answering evaluations were designed so that each user answers questions devised from the gold summary while having access to the model-generated summary only. The responses are evaluated on a 3-level scale (correct, partially correct, incorrect). The other evaluation performed was to assess the generation process on four criteria: informativeness, novelty, relevance, and fluency. Users evaluate and rank summaries generated by different models for each category, ordering them from best to worst after reviewing the original document. Only the selection of the best cases or the worst cases is considered when calculating the final scores for each criterion-candidate model pair. Paired-T tests are then conducted to compare to the baseline. Although the human evaluation experiments in this study are mainly focused on the performance rather than the explainability part of the solution, for evaluating self-explaining models, these methods can be adopted.

A more explainability-oriented human evaluation was conducted by Malkiel \emph{et al.} for their work on text similarity matching \cite{malkiel_interpreting_2022}. To compare the proposed solution to baseline explainability methods, they used a 5-point mean opinion score from novice users. The users were presented with randomly shuffled explanations of test set samples from the candidate methods. The users evaluate the same number of samples from each candidate from 1 (poor) to 5 (excellent), and the mean and the standard deviation of the experiment serve as the interpretation score. Similarly, a 5-point Likert scale is used for evaluation in the DSTC10 challenge \cite{yoshino2023overview}, where the study by Heo \emph{et al.} \cite{heo_natural-language-driven_2023} on multimodal scene-aware dialogue systems was one of the submissions. In this challenge, five humans evaluate each answer based on their own view of the correctness while also considering the relevance, fluency, and informativeness of the answer. The final score is the average of each criterion. This experiment is again suitable for high-level explanations, such as natural language or natural images.

\subsection{Other Metrics}
In addition to the two established evaluation methods with standard categories, some implementations for assessing explanations do not fit within these frameworks. For instance, plausibility is a well-established metric in XAI literature, used to evaluate how well explanations align with human understanding of a task, irrespective of their faithfulness \cite{carvalho2019machine}. By nature, plausibility necessitates human-centered studies, as it depends on subjective judgments grounded in domain knowledge. However, Rigotti \emph{et al.} addressed this by incorporating plausibility into their Concept Transformer as an integral part of the model's design, alongside faithfulness \cite{rigotti_attention-based_2022}. This was achieved by integrating human-interpretable concepts during training. Formally, given an expected attention distribution $H$ (aligned with domain knowledge) and attention weights $A$, they introduced a specialized explanation loss term $\mathcal{L}_{\text{expln}}$ in addition to the classification loss. This loss is defined as in equation \ref{eq:plausibility}.

\begin{equation}
    \mathcal{L}_{\text{expln}} = ||A - H||^2_F
    \label{eq:plausibility} 
\end{equation}

Here, $||\cdot||_F$ denotes the Frobenius norm. By minimizing this loss, the model adapts its weights to incorporate prior domain knowledge, thereby learning plausible, interpretable concepts.

Beyond these metrics, other approaches employing comparison-driven analyses use metrics that are not traditionally tied to explainability. For instance, studies often generate clusters of internal feature representations to establish a foundation for explanation, analyzing them via class-based grouping \cite{kumar_bert_2022} or thematic tags \cite{ilinykh_what_2021, li_oscar_2020}. These analyses leverage related metrics such as average cosine similarity \cite{ferrando_attention_2021, ilinykh_what_2021}, inter- and intra-cluster distances \cite{kumar_bert_2022}, or visual inspection \cite{li_oscar_2020}.

Notably, Feucht \emph{et al.} introduced an explanation method for ICD-9 label classification, but it is the only study in the review to not include any specific evaluation \cite{feucht_description-based_2021}.

\subsection{Discussion}

In summary, a total of 81 instances of evaluation methods were identified across the 52 studies included in the review. Five of these instances could not be classified within the proposed classification framework due to non-standard implementation or limited suitability as formal evaluation metrics. Nonetheless, they may still offer comparative insights, particularly when analyzing internal model components such as latent vectors. As shown in Figure \ref{fig:eval_venn}, the majority of metric instances are qualitative. This prevalence is likely due to the relative ease of implementation, even though such methods often yield subjective or partial insights. Authors may prefer these methods because they allow for intuitive explanations of a model's decision-making process without the need for rigorous experimental design. Human-centered quantitative studies are rare in the context of multimodality, with only the three instances identified. This scarcity may be attributed to the high cost of conducting such studies comprehensively and the lack of standardized study design guidelines. Among all the evaluation instances, 40.6\% employed objective metrics, and 42.3\% were found in studies that also used either qualitative or quantitative methods. Despite their presence, the use of objective metrics is highly concentrated with 15 out of the 26 instances relying solely on \textbf{faithfulness} as the metric. Moreover, none of the identified objective methods explicitly quantify inter-modal interactions or assess multimodality in a holistic manner. This indicates a narrow focus within the category and highlights the need for greater diversity in evaluation metrics along with dedicated methods for future multimodal research.

When the studies are aggregated by explanation method, clearer relationships between the employed XAI algorithm categories and their corresponding evaluation strategies emerge, as summarized in Table~\ref{tab:xai_methods_comparison}. Attention-based approaches, particularly those leveraging attention weights as explanatory signals, represent the most prevalent category in the reviewed literature. These methods are evaluated using a heterogeneous set of strategies, spanning both human-centered assessments and objective metrics. However, qualitative human-centered evaluation remains the most prevalent approach within this category. 

Moreover, with the exception of ante-hoc methods, nearly all other XAI categories are consistently accompanied by qualitative validation. This trend suggests a widespread reliance on human interpretability and subjective assessment, even when objective evaluation techniques are available. In contrast, ante-hoc methods tend to emphasize on faithfulness- or localization-oriented metrics rather than post-hoc qualitative evaluation. This distinction underscores a divergence in evaluation practices between explanation techniques, and further highlights the need for more standardized and comprehensive evaluation frameworks across XAI categories.

\begin{table*}[]
\centering
\scriptsize
\caption{
Comparison of explanation methods across categories with respect to the applied evaluation criteria: Faithfulness (Faith.), Localization (Loca.), Robustness (Robu.), Complexity (Compl.), Qualitative (Qual.), and Quantitative (Quan.) metrics. A checkmark indicates that a criterion is commonly addressed in the literature, while double checkmarks denote particularly strong qualitative performance. \textit{Italicized} entries refer to specific named methods, whereas non-italic entries denote broader methodological families.
}
\label{tab:xai_methods_comparison}
\begin{tabular}{lp{140pt}|cccc|cc}
\toprule
\textbf{Category} & \textbf{Method} &
\textbf{Faith.} & \textbf{Loca.} & \textbf{Robu.} &
\textbf{Compl.} & \textbf{Qual.} & \textbf{Quan.} \\
\midrule

\multirow{4}{*}{Ante-hoc}
 & Interpretable parameters              & \checkmark & \checkmark &  &  &  &  \\
 & Association maps                      &            & \checkmark &  &  &  &  \\
 & Feature importance                    &            &            &  &  &  & \checkmark \\
 & Differentiable physics models         & \checkmark &            &  &  &  &  \\

\midrule
\multirow{2}{*}{Model-agnostic}
 & \textit{LIME}                                  & \checkmark &  & \checkmark &  & \checkmark &  \\
 & \textit{SHAP}                                  &            &  &  &  & \checkmark &  \\

\midrule
\multirow{5}{*}{Attention-based}
 & Attention scores/weights                     & \checkmark & \checkmark &  &  & \checkmark\checkmark & \checkmark \\
 & Representation clustering             & \checkmark &  &  & \checkmark & \checkmark &  \\
 & \textit{ALTI+}                                 &            &            &  &  & \checkmark &  \\
 & Soft alignment                        &            &  &  & \checkmark & \checkmark &  \\
 & Similarity scoring                    &            &  &  &  & \checkmark &  \\

\midrule
\multirow{2}{*}{Gradient-based}
 & \textit{Grad-CAM}                              &  &  &  &  & \checkmark &  \\
 & \textit{Integrated Gradients}                  &  &  &  &  & \checkmark &  \\

\midrule
\multirow{4}{*}{Attention composite}
 & \textit{Transformer attribution}               &  &  &  &  & \checkmark &  \\
 & Gradient-weighted attention           & \checkmark & \checkmark &  &  &  &  \\
 & Perturbation-based attention analysis &  & \checkmark &  &  & \checkmark &  \\
 & \textit{BERT interpretations}             &  &  &  &  & \checkmark & \checkmark \\

\midrule
\multirow{2}{*}{Probing \& ablation}
 & Probing classifiers                   & \checkmark &  &  &  & \checkmark &  \\
 & Adaptive probing and ablation         &  &  &  &  & \checkmark &  \\

\midrule
\multirow{6}{*}{Hybrid approaches}
 & Attention + \textit{LIME} + perturbation + gradients 
                                         & \checkmark &  &  &  &  &  \\
 & Attention + occlusion                 & \checkmark &  &  &  & \checkmark &  \\
 & Attention + \textit{LRP}                       &  &  &  &  & \checkmark &  \\
 & Attention + \textit{Grad-CAM}                  &  &  &  &  & \checkmark &  \\
 & \textit{Grad-CAM} + \textit{SHAP}                       &  &  &  &  & \checkmark &  \\
 & Attention + \textit{Grad-RAM}                  &  &  &  &  & \checkmark &  \\

\midrule
Self-explaining models
 & Explanation-aware learning objectives & \checkmark &  &  &  & \checkmark & \checkmark \\

\bottomrule
\end{tabular}
\end{table*}

\section{Explanation Interfaces\label{Explanation Interfaces}}

Another critical aspect of explainability is its presentation. Explanations often lack the cognitive engagement necessary to foster trust and reliability. Consequently, an emerging area of research focuses on the intersection of human-computer interaction (HCI) and XAI to make AI decision-making processes more accessible \cite{gilpin2018explaining, mohseni2021multidisciplinary}. In this review, beyond studies that explore explainability as a primary or secondary objective, we highlight three works that specifically address the presentation and interface of explanations \cite{sarti_inseq_2023, katz_visit_2023, aflalo_vl-interpret_2022}.

Sarti \emph{et al.} introduce Inseq, a Python toolkit that adapts various perturbation-, gradient-, and model-internal attribution methods for both encoder–decoder and decoder-only sequence generation models \cite{sarti_inseq_2023}. In addition to a lightweight visualization framework, the toolkit supports different aggregation strategies for token-level attributions (e.g., subword-, neuron-, or layer-based grouping). The interface is demonstrated through multiple case studies. One application involves detecting gender bias in NMT by analyzing the use of occupations and pronouns to uncover spurious correlations across gendered translations. Qualitative validation is also conducted using attribution scores for sample sentence pairs. Another case study focuses on locating factual knowledge in GPT-2 using contrastive attribution tracing, which identifies layer-wise and token-wise contributions to factual content. The toolkit is well-suited for generating explanations in a range of generation tasks, including NMT, code synthesis, and dialogue generation.

\begin{figure}[h]
    \centering
    \includegraphics[width=0.8\linewidth]{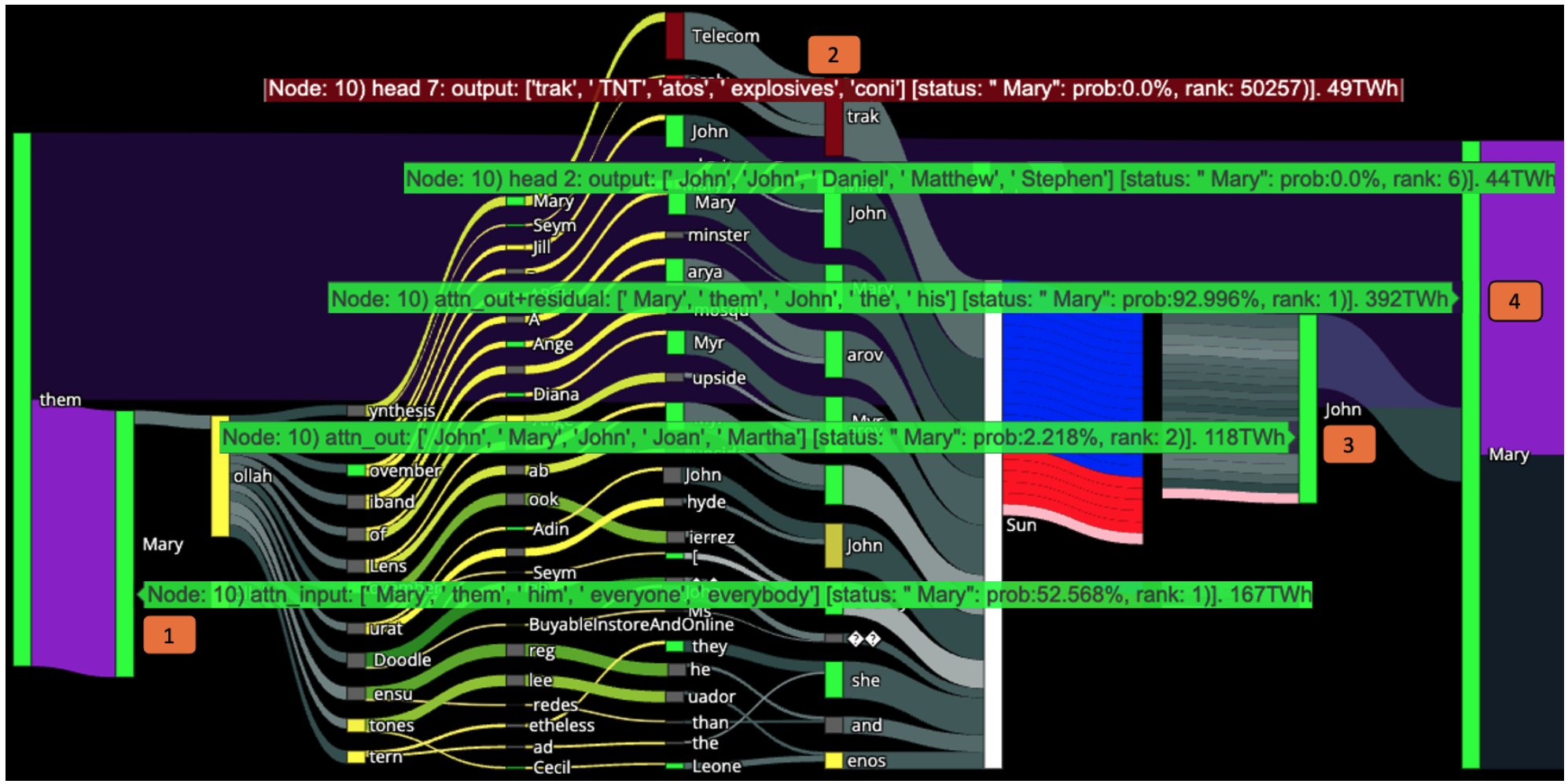}
    \caption{Flow-graph for layer 10 attention from GPT-2 small in VISIT for an IOI task \cite{katz_visit_2023}.}
    \label{fig:katz_visit}
\end{figure}

In another study, Katz and Belinkov present VISIT, a tool designed to visualize semantic patterns in the information flow of transformer-based, GPT-style models \cite{katz_visit_2023}. VISIT projects attention weights and hidden states into the vocabulary space, rendering them as forward-flow graphs that reflect attention head activations and dynamic memory states during prompt completion in GPT-2. The tool also enables probing of other model components to better understand decision-making processes. Human-centered workflows are supported in two key ways: first, through intuition-driven analysis of an indirect object identification (IOI) task, which highlights the role of specific transformer components (as seen in Figure \ref{fig:katz_visit}); and second, through comparison-driven analysis to examine the influence of layer normalization and regularization across different model configurations.

Finally, Aflalo \emph{et al.} propose VL-InterpreT, an interface for interpreting VL multimodal transformer models \cite{aflalo_vl-interpret_2022}. VL-InterpreT provides visualization of both cross-modal and intra-modal attention. It comprises two main components: attention maps across all heads and layers, and hidden state plots that track the evolution of token embeddings. The interface is validated through case studies involving the KD-VLP model \cite{liu2021kd}, analyzed on two benchmarks—VCR and WebQA—using attention-based metrics, cluster analysis, and heatmaps. The analysis covers both successful and failure cases, offering a bottom-up, visually guided approach to model interpretation for end users.

Taken together, these interface-focused tools transform raw model internals into intuitive and interactive experiences. They enable users to explore attention flow, detect biases, trace factual retrieval, and compare architectural variants—effectively bridging the gap between complex transformer operations and meaningful user insights.

\section{Beyond the Systematic Search: Emerging and Complementary Directions\label{sec:Beyond}}

While our systematic review provides a view of the current explainability practices in the research scape of multimodal attention-based models, the rapidly growing field warrants broader reflection. To contextualize our findings and acknowledge promising developments beyond the scope of the systematic search, this section provides a narrative discussion of emerging approaches and, in the case of evaluation, influential classic methods that continue to shape the area.

We structure this discussion using the same three key dimensions as the primary analysis: multimodal architectures, explanation algorithms, and evaluation criteria. Unlike the SLR, the works highlighted here may focus on individual dimensions rather than addressing all three simultaneously. For each, we highlight notable developments beyond the search scope, focusing on emerging techniques and, where relevant, classic evaluation frameworks that offer opportunities for adaptation in multimodal XAI.


\subsection{Emerging architectures}

From a modeling perspective, growth in the multimodal domain has been rapid. Especially since the introduction of transformers, there have been considerable advances in architectures capable of processing image and text modalities, with extensions to additional modalities steadily catching up \cite{yin2024survey, wadekar2024evolution}. The most powerful of these are foundation LLMs and large VL models, often aligned to tokens originating from diverse modalities in addition to text. Beyond leveraging large amounts of injected knowledge, these larger multimodal models demonstrate several emergent capabilities on top of traditional tasks, such as mathematical reasoning from images without requiring optical character recognition or visual storytelling \cite{yin2024survey}. Since this class of models builds on LLM capabilities, the text modality is often involved either as the sole generation modality or in both the input prompt and output.

Modern multimodal LLMs (MLLMs) are characterized by parameter counts often in the billions. Beyond scale, their integration of multimodal data involves varying architectural choices. For instance, \textbf{Early Concatenation}-based MLLM architectures typically use LLMs as the core model with regular textual tokens. Representations for other modalities are either obtained by linearly projecting each modality's embeddings into the text space (e.g., LLaVa \cite{liu2023visual}, DeepSeek-VL \cite{lu2024deepseek}) or through specialized components such as Q-former (e.g., BLIP-2 \cite{li2023blip}, Video-LLaMA \cite{zhang2023video}) or attention pooling (e.g., Qwen-VL \cite{bai2023qwen}). Due to extensive pretraining, \textbf{Single-stream generative} or decoder-only models are also common. In these cases, a separate tokenizer is learned—for example, for image tokens—used alongside text tokens as input to the LLM decoder. Generated tokens are then detokenized using modality-specific tokenizers (e.g., LaViT \cite{jin2023unified}, VL-GPT \cite{zhu2023vl}) or unified frameworks such as MiniGPT-4 \cite{zhu2023minigpt}.

More commonly, \textbf{Cross-attention variants} are used to scale LLMs to additional modalities. Wadekar \emph{et al.} classify cross-attention variants in MLLMs into two types: standard cross-attention and custom cross-attention layers \cite{wadekar2024evolution}. The first category includes models trained with a next-token prediction objective where multimodal data is injected into the language backbone using standard (or slightly modified) cross-attention mechanisms. These models may use early concatenation between text and multimodal representations and follow either an encoder–decoder design (e.g., VL-BART and VL-T5 \cite{cho2021unifying}, PaLI-X \cite{chen2023pali}) or employ cross-attention layers to inject multimodal information into decoder-only LMs (e.g., Flamingo with gated cross-attention \cite{alayrac2022flamingo}). The second category uses specialized layers to integrate cross-modal information into pretrained LMs. For example, LLaMA-Adapter-V2 introduces feedforward bottleneck layers (or adapters) and learnable bias terms to the attention and feedforward outputs of an otherwise frozen LM \cite{gao2023llama}. The model can then be trained for multimodal tasks by prepending projected embeddings from other modalities to text tokens under an autoregressive objective. Similar adapter-like designs have been explored for encoder–decoder models as well, for instance, VL-Adapter employs lightweight MLP or attention-based modules for multimodal learning \cite{sung2022vl}. In addition to these adapter-based methods, modern MLLMs also adopt parameter-efficient techniques such as low-rank adaptation (LoRA) \cite{hu2022lora}, weight freezing, quantization, shared parameters, and prefix tuning \cite{wu2023multimodal, jin2024efficient, yin2024survey}.

Another powerful parameter-efficient technique, known as the mixture of experts (MoE), has also seen increased adoption in language modeling and computer vision. Originally introduced as an adaptive gating mechanism for expert multilayer networks of subtasks, alongside a hierarchical tree-structured variant \cite{jacobs1991adaptive, jordan1994hierarchical}, MoE has undergone several evolutionary steps leading to sparse MoE architectures for transformer-based LLMs \cite{lepikhin2020gshard, cai2025survey}. Modern sparse MoE models incorporate task-specific expert units that are activated only when a routing mechanism deems an input suitable for that expert. This enables efficient integration of multiple experts within a single architecture. Sparse MoEs are increasingly used in multimodal learning, where expert networks correspond to each modality. Beyond efficiency, multimodal MoE (MMoE) architectures help address the common challenge of poor task generalization in multimodal models \cite{shen2024mome}. One of the earliest MMoE models is LIMoE by Mustafa \emph{et al.} \cite{mustafa2022multimodal}. Trained with a multimodal contrastive loss for VL tasks, this single-tower transformer employs several MoE layers that route input tokens to expert feedforward networks in a modality-agnostic manner. Tokens are then pooled and linearly aggregated based on learned gating weights. To ensure training stability and balanced expert utilization, auxiliary loss functions and entropy-based regularization is employed. LIMoE achieves competitive accuracy compared to dense baselines while being more efficient, and the analysis of its routing confidences reveals modality-specific and multimodal-specialist experts, even without explicit supervision. McKinzie \emph{et al.} examine the fundamental aspects of sparse MoE architectures for VL tasks through ablations of key components: the image encoder, VL connector, pretraining strategy, and language decoder \cite{mckinzie2024mm1}. They scale the decoder using sparse MoE by replacing every few dense layers with expert FFNs and a top-\textit{k} gating mechanism. Building on these principles, they introduce the MM1 model, which achieves performance comparable to supervised fine-tuned models across several benchmarks. Similar design choices have been explored in other studies such as DeepSeek-VL2 \cite{wu2024deepseek}, MoE-LLaVA \cite{lin2024moe}, and Uni-MoE \cite{li2025uni}. Dense variants of MoE have also been proposed, such as dense mixtures with soft gating using LoRA-based experts \cite{wu2024mixture}.

Many MLLMs are further refined through alignment techniques to better control their generation behavior. These may include reinforcement learning from human feedback (RLHF) \cite{ouyang2022training}, direct policy optimization (DPO) \cite{rafailov2023direct}, or emerging multimodal preference optimization methods \cite{wang2024enhancing}. As models become larger and their internal decision processes more complex, alignment procedures introduce additional challenges for explainability.

Overall, the massive scale of parameters in MLLMs and foundation VL models diffuses attributions across layers and modalities. Other major challenges involve unclear semantics of emergent features and complex cross-modal alignments \cite{kazmierczak2025explainability}. Consequently, explaining these models becomes increasingly difficult, both in isolating the parameters responsible for individual inferences and in understanding the contribution of intermediate steps.

\subsection{Emerging Practices in Explanation Algorithms}

While modeling practices have advanced rapidly, corresponding explanation methods have not achieved comparable success, especially for MLLMs. One widely studied direction is natural-language explanation via self-explaining models, which offer several practical advantages (see Section \ref{sec:nle}) but whose faithfulness remains uncertain. Another promising direction that has emerged, especially for these larger models, is mechanistic interpretability. Mechanistic interpretability is an ante-hoc model-specific mechanism for decomposing and analyzing models structurally at algorithmic level. Below we discuss these and other trending practices for explanation in extension to the ones listed in our systematic review.

\subsubsection{Multimodal Reasoning and NLE}
NLE have gotten much more effective and powerful with LLMs, especially after the observation of CoT \cite{wei2022chain} and zero-shot reasoning capabilities \cite{kojima2022large}. Natural language inference datasets with human annotated justifications such as the e-SNLI \cite{camburu2018snli} have enabled the systematic evaluation of these explanations. This area has also been expanded to MLLMs and VL tasks extensively \cite{wang2025multimodal}. One of the earliest of such approach, Interactive Prompting Visual Reasoner (IPVR), introduced by Z.~Chen \emph{et al.}, is a knowledge-based visual reasoning method \cite{chen2023see}. They achieve visual reasoning in VQA through contextualizing an LLM with global and local description of key objects in the image required for the answer and corresponding evidence-based rationale. For MLLMs, Zhang \emph{et al.} introduced a two-stage VL CoT mechanism which is able to generate textual rationale from both modalities in VQA tasks \cite{zhang2023multimodal}.

In addition to inherent or externally trained deep reasoning solutions, RL-based optimization of reasoning trajectories have been explored in various studies such as VisualThinker-R1-Zero \cite{zhou2025r1}, MM-Eureka \cite{meng2025mm}. Recent research has also emphasized test-time scaling, where larger LLMs or ensembles are leveraged during inference to improve reasoning quality without additional training, often enhancing the reliability of generated justifications. The Tree of Thoughts framework further extends this idea by structuring reasoning as a search over multiple candidate thought paths, allowing models to explore diverse reasoning trajectories before committing to an answer \cite{yao2023tree}.

The introduction of multimodal CoT benchmark datasets such as e-SNLI-VE \cite{do2020snli}, e-VIL \cite{kayser2021vil}, Visual-CoT \cite{shao2024visual} has empowered the evaluation of such systems. Several other multimodal CoT based experiments have since been introduced leveraging different modalities in providing justifications all while contributing to the primary task objectives, comprehensively discussed in survey articles such as the one by Y.~Wang~\emph{et al.} \cite{wang2025multimodal}.

\subsubsection{Mechanistic Interpretabilty}

Mechanistic interpretability aims to explain model behavior by identifying the internal algorithms implemented by neural networks. This is typically achieved through structural analyses, such as causal interventions, circuit tracing, and ablation studies, that attempt to isolate functional subcomponents of a model. As a form of global explainability, mechanistic interpretability seeks not only to characterize model outputs but to uncover how computation is distributed internally. However, these methods become increasingly complex as model size grows, and the difficulty is amplified in multimodal architectures, where interactions between modalities can obscure the underlying computational pathways.

One of the earliest efforts to extend mechanistic analysis to multimodal systems is the work of Goh~\emph{et al.}~\cite{goh2021multimodal}. Studying CLIP, a dual-encoder vision–language model trained with a contrastive objective, they combined sparse linear probes with feature visualization techniques to identify neurons responding to shared high-level concepts across image and text modalities. Their results suggested that multimodal alignment in CLIP is partially mediated by units that encode modality-invariant semantic features.

Mechanistic analyses of transformers in the language domain have also provided insights relevant for multimodal models. Olsson~\emph{et al.}~\cite{olsson2022context} used causal interventions to identify ``induction heads" or attention heads that learn to copy and continue token sequences. Their circuit-level analysis showed that these heads are central to in-context learning in language models. Although this work focused on unimodal text models, induction-style mechanisms have since been hypothesized to contribute to pattern continuation and cross-modal grounding in multimodal transformers as well. Verifying this, however, remains an open research direction.

Mechanistic interpretability can also reveal how pretraining and fine-tuning shape the internal structure of multimodal models. Salin~\emph{et al.} \cite{salin2022vision} introduced a suite of controlled probing tasks, including unimodal probes (e.g., part-of-speech tagging, object counting) and multimodal probes (e.g., color association, adversarial captions), to assess how different pretraining strategies contribute to downstream vision–language capabilities. Their framework enables quantitative comparisons of representational structure across training regimes.

Causal Mediation Analysis (CMA) has emerged as another influential tool for mechanistic interpretability. CMA decomposes model behavior into latent causal pathways and has been widely applied in LLMs to assess the contribution of specific states or components to factual recall \cite{meng2022locating}. Adaptations to multimodal transformers typically involve activation patching or structured perturbations to track how information flows across modalities. For example, Palit~\emph{et al.} examined the BLIP model by comparing the causal influence of intermediate activations for clean versus corrupted visual inputs \cite{palit2023towards}. Golovanevsky~\emph{et al.} extended these methods to BLIP and LLaVA, using symmetric token replacement and paired image corruption to trace contributions from both cross-attention and self-attention pathways \cite{golovanevsky2024vlms}. Similarly, Neo~\emph{et al.} \cite{neo2024towards} used CMA-style interventions to show that visual tokens in LLaVA progressively adopt language-like internal structure during inference, supporting the view that multimodal transformers converge toward shared representational semantics, even in the absence of targeted pretraining.

\begin{figure}
    \centering
    \includegraphics[width=0.3\linewidth, height=0.3\linewidth, trim = 0 65 0 100, clip]{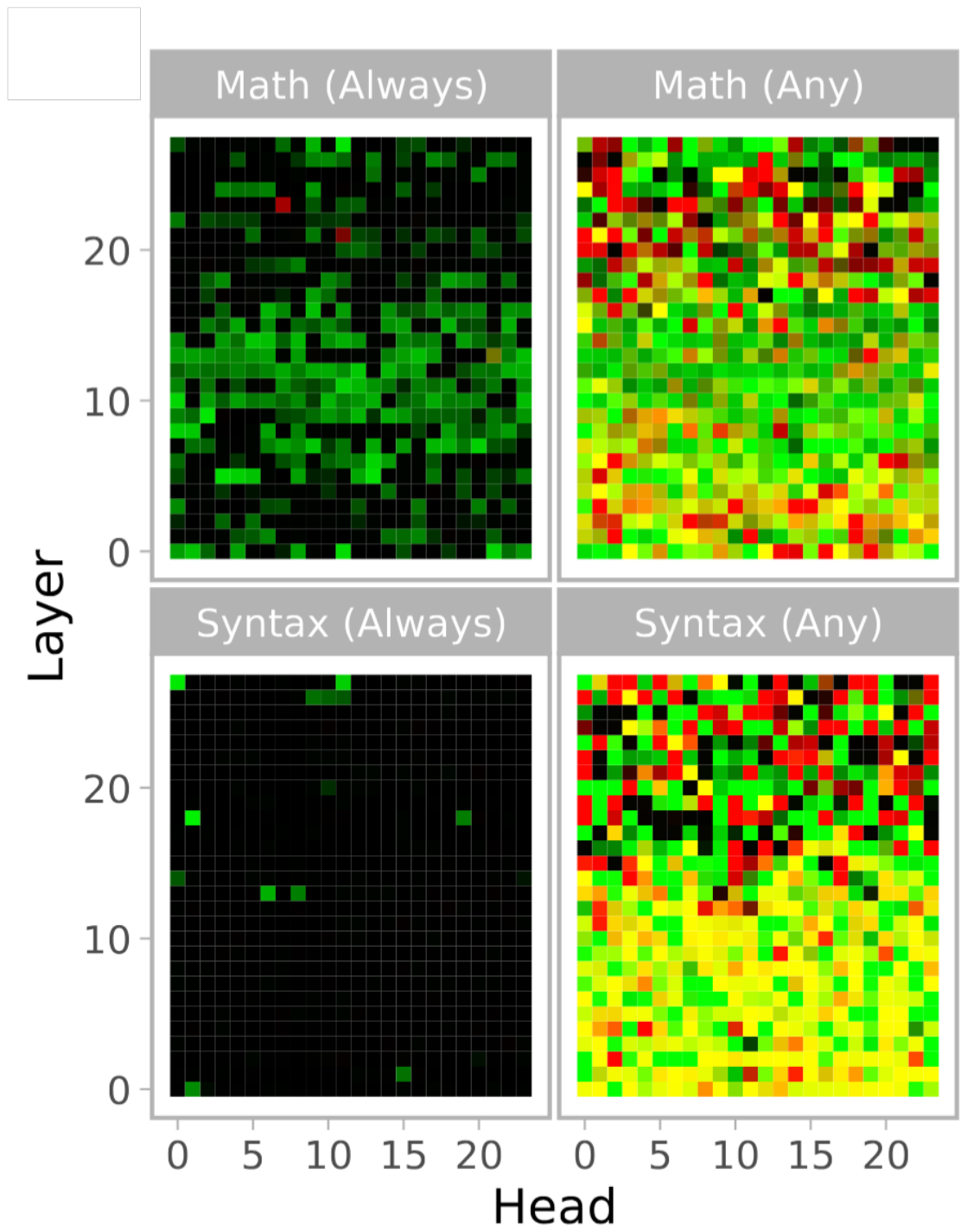}
    \caption{CHG scores aggregated by two methods (``Always" vs ``Any") for the instruction-tuned Llama-3 model (L3.2-3BI) across mathematical and syntactic reasoning tasks. Red denotes negative (interfering) contribution, green denotes positive (facilitating) contribution, yellow indicates overlap (both interference and facilitation), and black denotes irrelevance. \cite{nam2025causal}. }
    \label{fig:chg}
\end{figure}

Sparse methods also have recently gained traction as a scalable approach to identifying human-interpretable circuits in large models. Sparse autoencoders, in particular, can disentangle superposed features by learning high-dimensional sparse codes of dense activations. Gao~\emph{et al.} \cite{gao2025weight} demonstrated that combining sparsity-inducing objectives with model pruning can reveal minimal circuits sufficient for fine-grained tasks, thereby mitigating the ``superposition problem." Yang~\emph{et al.} \cite{nam2025causal} integrated sparse learning with causal analysis by introducing Causal Head Gating (CHG), a soft-gating mechanism applied to attention heads. With appropriate regularization, CHG estimates the positive, negative, or neutral causal contribution of each head to specific tasks (example in Figure \ref{fig:chg}). Unlike many mechanistic interpretability methods, CHG is computationally scalable, does not rely on curated diagnostic datasets, and is directly applicable to multimodal architectures.

Additional tools for mechanistic analysis include the logit lens, which traces the evolution of logits across layers \cite{gandelsman2024interpreting, huo2024mmneuron}, as well as synthetic circuit-level benchmarks designed to test mechanistic generalization \cite{rai2024practical}. Recent surveys provide broader syntheses of mechanistic interpretability advances in multimodal settings \cite{lin2025survey, liu2025mechanistic}.

\subsubsection{Other Notable Algorithms}

In parallel to these more recently popularised developments, there has been a steady adoption of explainability practices tailored to multimodal use cases. One particularly notable direction is the use of neurosymbolic reasoning in MLLMs. Instead of relying solely on natural language CoT, intermediate reasoning steps can be constrained to logical or symbolic rules, thereby providing an interpretable bridge between neural predictions and final outputs. Such approaches have been explored in tasks including code generation \cite{gao2023pal}, logical reasoning and deduction \cite{olausson2023linc, ye2023satlm}, and visually grounded reasoning \cite{hu2024visual, menon2024whiteboard, wang2025vgr}. A more advanced line of research aims to ground this reasoning in embodied cognitive schema, aligning neural–symbolic inference more closely with human cognitive processes \cite{pmlr-v284-olivier25a}. Although integrating symbolic formalisms with neural architectures remains technically challenging, neurosymbolic pipelines offer a path toward more robust and interpretable multimodal systems, with growing implications for symbolic–neural integration in multimodal reasoning \cite{bhuyan2024neuro}.

Beyond the attention-centric multimodal explanation algorithms discussed in Section~\ref{sec:att-cen}, several efforts have attempted to generalize model-agnostic interpretability methods to the multimodal setting. A representative example is the disentangled multimodal explanations (DIME) framework, which extends the LIME methodology by decomposing explanations into unimodal contributions and multimodal interaction terms. This disentanglement enables clearer attribution of model behavior, and additional optimization techniques reduce the computational overhead associated with perturbation-based analysis. More recently, Wenderoth~\emph{et al.}~\cite{wenderoth2025measuring} introduced an alternative decomposition grounded in the Shapley Interaction Index (SII), allowing pairwise modality interactions to be isolated and aggregated. Their final cross-modal interaction measure, termed \textit{InterSHAP}, is defined as in equation \ref{eq:intershap}.

\begin{align}
    \textit{InterSHAP} &= \frac{\textit{Interactions}}{\textit{Behavior}}\label{eq:intershap}\\[4pt]
    where, \textit{Interactions} &= \sum_{\substack{i,j=1 \\ i \ne j}}^{M} \Phi_{ij}, 
    \textit{Behavior} = \sum_{i,j=1}^{M} \Phi_{ij}\notag
\end{align}

Here $\Phi_{ij}$ denotes the SII between modalities $i$ and $j$ for a given model $f$ and sample. InterSHAP quantifies the proportion of a model's behavior attributable specifically to cross-modal interactions and generalizes naturally to any number of modalities. A similar SII-based explanation method, MultiSHAP, was introduced by Z.~Wang~\emph{et al.}, offering fine-grained analysis of interactions between visual patches and text tokens \cite{wang2025multishap}. While the computational cost increases exponentially with the number of modalities, an inherent limitation of Shapley-based methods, these approaches remain promising due to their model-agnostic nature, applicability to both global and local explanations, and capacity to yield interpretable interaction structures in complex multimodal systems.

\subsection{Explainability Evaluation: Insights from the Past and Present\label{sec:prom_eval}}
Evaluating explanations remains challenging due to the subjective and context-dependent nature of interpretability. While the SLR captures evaluation approaches reported in the selected studies, the broader XAI literature provides additional perspectives on established and emerging methods for assessing multimodal explanations. This section therefore discusses complementary evaluation approaches, covering both objective and human-centered methods.

\subsubsection{Cross-modal Objective Metrics\label{sec:recent_eval}}
As noted earlier, objective evaluation methods in multimodal explainability remain relatively shallow in scope, primarily due to the structural and semantic heterogeneity across modalities. Beyond the techniques identified in the SLR, prior work has proposed a range of classic and emerging approaches that aim to improve the rigor and comparability of multimodal XAI evaluations.

\textbf{Randomization}-based approaches, such as the sanity checks with the Model Parameter Randomization Test (MPRT) and Data Randomization Test proposed by Adebayo~\emph{et al.}~\cite{adebayo2018sanity}, offer a fast and effective means of assessing whether explanations are truly sensitive to model parameters and data statistics. These tests, which involve randomizing model weights or labels, are largely model-agnostic and can be naturally extended to multimodal architectures by additionally shuffling or permuting cross-modal pairings. Their generality makes them particularly useful for assessing whether multimodal attention maps or alignment scores reflect genuine learned structure rather than architectural or dataset priors.

Cross-modal consistency, introduced by X.~Zhang~\emph{et al.}~\cite{zhang2024cross}, represents a more recent \textbf{axiomatic} framework tailored to multimodal settings. Their task-consistency score quantifies the degree to which explanations or predictions across modalities preserve semantically equivalent information. This metric enables the detection of modality imbalance or over-reliance on a dominant modality, a commonly observed issue in large VL models. Although initially demonstrated on a custom dataset for comparative analysis, the principle of cross-modal agreement generalizes broadly and complements existing unimodal consistency tests in XAI. Other multimodal-specific evaluation methods include measures of modality heterogeneity~\cite{liang2022high}, multimodal matching scores~\cite{jeong2024multimodal}, and recently proposed notions of synergistic faithfulness and unified stability~\cite{agarwal2025rethinking}, each capturing complementary dimensions of multimodal coherence and alignment.

Classic algorithmic frameworks from the broader XAI literature also continue to play an important role, though they require careful adaptation to multimodal architectures. For example, the ROAR framework by Hooker~\emph{et al.}~\cite{hooker2019benchmark} extends perturbation-based evaluation by retraining models after removing the most salient features identified by an explanation method. This allows direct causal linkage between feature importance and predictive performance. For multimodal systems, however, feature removal must be designed to handle heterogeneous representations (e.g., pixels, tokens, acoustic frames), and retraining must account for interactions across modalities. Despite the increased computational cost, ROAR remains a valuable causal validation tool when multimodal perturbations are implemented thoughtfully. Other widely used quantitative methods, including meaningful perturbation, deletion and preservation tests, minimal sufficient explanations~\cite{fong2017interpretable}, local Lipschitz-style robustness analyses~\cite{alvarez2018robustness}, and counterfactual fidelity metrics~\cite{wachter2017counterfactual, white2019measurable}, can similarly provide strong quantitative foundations when extended to multimodal settings.

A further important consideration is the quantitative characterization of tradeoffs between distinct objective criteria. The infidelity–sensitivity metrics of Yeh~\emph{et al.}~\cite{yeh2019fidelity} formalize a fundamental tension between \textbf{faithfulness} (alignment with the true predictive mechanism) and \textbf{robustness} (stability under small perturbations). Deletion and insertion-based metrics similarly expose relationships linking \textbf{faithfulness}, \textbf{robustness}, and \textbf{complexity} of explanations~\cite{samek2016evaluating, petsiuk2018rise}. Similarly, analyses comparing sparsity against faithfulness demonstrate that overly sparse explanations may compromise fidelity to the underlying model~\cite{pmlr-v238-sun24b}. Systematically quantifying and resolving such tradeoffs is essential for developing standardized and reliable evaluation frameworks for multimodal explainability.

\subsubsection{Multimodal Human-centered Metrics}

Human-centered evaluation has become an important component of explainability assessment, particularly as AI systems move from unimodal to multimodal settings. Recent studies emphasize that explanations should not only reflect model behavior but also support human understanding, trust, and decision-making \cite{lopes2022xai, bobek2025user}. As a result, human-centered evaluation practices are gradually shifting from ad-hoc qualitative inspection toward more structured approaches that combine qualitative analysis, human studies, and domain-informed validation~\cite{kim2024human,naveed2024overview}. In survey literature, explanation quality in human-centered evaluation is typically assessed through dimensions such as interpretability, usefulness in human–AI interaction, and the impact of explanations on human task performance~\cite{kim2024human}. However, standardized methodologies remain limited, and evaluation protocols are often tailored to specific domains or multimodal tasks.

\paragraph{Qualitative Evaluation Across Domains} Qualitative methods such as intuitive analysis and domain-driven evaluation are increasingly extended to multimodal systems. For example, in VL models, explanation analysis often combines visual attention maps with textual rationales to reveal cross-modal interactions~\cite{agarwal2025rethinking}. These multimodal probing approaches allow researchers to inspect whether highlighted visual regions correspond to the reasoning expressed in generated explanations. Here, while intuitive and comparison-driven analyses remain general tools for interpretability inspection, domain-driven approaches are particularly important in high-stakes applications where explanations must align with expert knowledge and operational practices.

In autonomous driving, explanations are frequently validated against traffic rules, scene semantics, and driver behavior patterns~\cite{atakishiyev2024explainable}. For instance, saliency maps can be overlaid on driving scenes to verify whether highlighted regions correspond to decision-critical objects such as pedestrians or traffic signals. Datasets such as BDD-X provide annotated driving scenarios and textual rationales that enable explanation validation grounded in human driving expertise~\cite{kim2018textual}. Recent multimodal models further integrate video, images, and text to generate natural-language explanations for driving actions, which are often assessed through expert feedback or alignment with known driving strategies~\cite{atakishiyev2024explainable,kuznietsov2024explainable}. 

More advanced approaches incorporate domain knowledge directly into the modeling process. Tang \emph{et al.} propose a grounded explanation framework in which expert knowledge about traffic interactions is encoded into reward functions for a Grounded Relational Inference (GRI) model~\cite{tang2024grounded}. The resulting interaction graphs represent relationships among traffic participants, providing semantically meaningful explanations that allow users to monitor and assess model reasoning.

Human-centered evaluation is also widely applied in finance, where explanations must support regulatory compliance and expert auditing. In tasks such as fraud detection or credit risk prediction, explanations, often generated using model-agnostic post-hoc methods, are compared against known transaction patterns or financial risk indicators~\cite{weber2024applications}. Domain experts may audit feature importance scores to ensure that explanations align with established credit risk factors and regulatory guidelines~\cite{talaat2024toward}. Although many current financial applications operate on unimodal data, the domain-driven evaluation protocols used in these settings provide a foundation for assessing explanations in emerging multimodal financial systems.

\begin{figure}[]
\centering
\includegraphics[width=\linewidth]{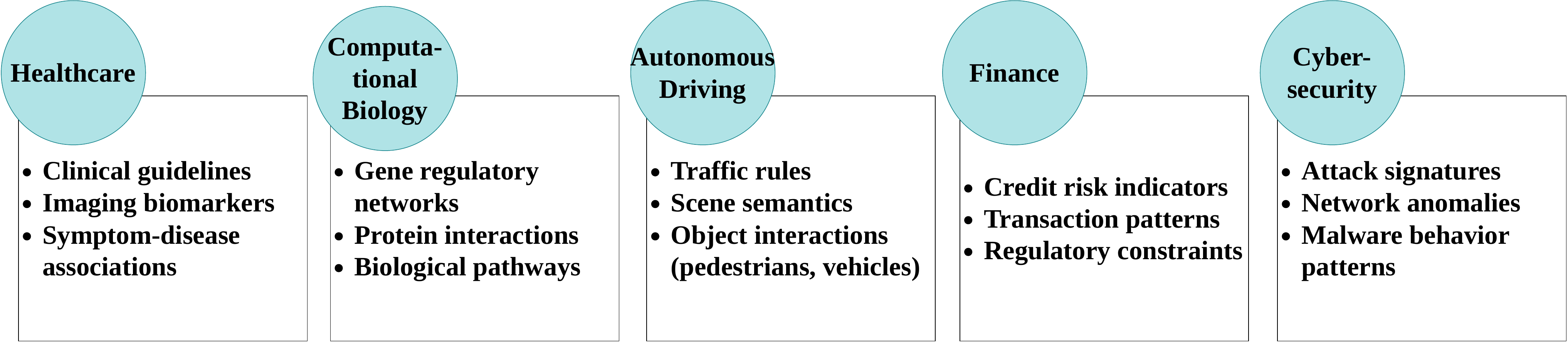}
\caption{Examples of domain-specific concepts used for human-centered validation of AI explanations in high-stake domains.}
\label{fig:dd}
\end{figure}

Examples of domain-specific insights used to validate explanations in multimodal applications across different domains are illustrated in Figure~\ref{fig:dd}.

\paragraph{Quantitative Human-centered Evaluation}

While user studies based on surveys, Likert-scale questionnaires, and point-based question answering remain common practices, recent literature emphasizes that quantitative evaluation should also incorporate objective, and task-oriented metrics that capture how explanations influence human understanding and interaction with AI systems. Reviews of explainability evaluation frameworks indicate that quantitative human-centered methods typically assess aspects such as user trust, task performance, cognitive load, and explanation usefulness through measurable indicators rather than relying solely on subjective feedback~\cite{lopes2022xai}.

Beyond qualitative interviews and subjective surveys, Lopes \emph{et al.} discuss a broader set of human-centered evaluation methods used across different assessment dimensions~\cite{lopes2022xai}. For instance, in the human–AI task performance dimension, quantitative evaluation examines whether the presence of explanations improves users' ability to work with or interact with ML models. These studies often involve controlled experiments in which participants complete tasks with or without explanations, and performance metrics such as task accuracy, decision time, or error rates are measured. Trust and reliance can be assessed through measures such as trust evolution over time (e.g., trust curves) or compliance rates indicating how frequently users follow system recommendations. Similarly, understandability may be evaluated by testing whether users can predict model outputs from explanations or detect potential model failures. Although subjective instruments such as Likert-scale questionnaires and usability surveys remain widely used, these approaches illustrate how objective behavioral outcomes can be combined with subjective feedback to assess whether explanations genuinely improve human–AI collaboration rather than merely appearing plausible to users. Objective metrics may further complement these evaluations by including computationally derived indicators in a reproducible manner.

In the context of multimodal AI systems, quantitative human-centered evaluation can extend these principles to account for cross-modal reasoning. Metrics may assess how effectively explanations capture interactions between modalities, such as the alignment between visual evidence and generated textual rationales in VL models. Controlled user studies can also evaluate whether multimodal explanations improve human understanding of model decisions in complex tasks, for example by comparing user performance when explanations incorporate a single modality versus multiple modalities.








\section{Recommendation and Future Directions}\label{Recommendation and Future Directions}

Based on our analysis of the current state of explainability in multimodal attention-based models, we identify several key areas that warrant further attention. Building on insights from our SLR and recent advances in explainable multimodal learning, we outline the following recommendations aimed at guiding future progress toward principled, robust, and cognitively grounded multimodal explainability.

\subsection{Streamlining Multimodal Architecture}

As deep neural network architectures continue to evolve, their design and training pipelines have grown increasingly complex, particularly in multimodal settings. This trend is evident even within the narrower scope of attention-based models reviewed here. The wide variety of architectural configurations, especially fusion mechanisms, poses significant challenges for explainability. Architectural choices are often influenced by baseline models within specific subfields, leading to clusters of similar designs that are evaluated in isolation. Consequently, identical tasks in different domains may employ fundamentally different model structures, complicating systematic comparison.

To promote generalizable explainability, architectural choices should be streamlined across tasks and domains. This includes the adoption of representational benchmarks for evaluating diverse fusion strategies—early fusion, hierarchical fusion, cross-attention, modular fusion, and MoE across modalities or tasks. Benchmarking across these standardized categories ensures that token-level and cross-modal interaction pathways are captured consistently and provides an empirical basis for selecting appropriate fusion strategies in specific contexts. Open, controlled experimental models with known modality interaction patterns can serve as essential baselines for evaluating explainability methods. Such models, analogous to controlled circuits in mechanistic interpretability, enable precise analysis of how explainers behave under predefined interaction structures.

Mechanistic tracing of multimodal circuits offers a promising direction for uncovering modality-specific contributions. To fully leverage such analyses, abstractions of multimodal pipelines (e.g., modality-specific encoders, cross-modal connectors, and generative decoders) should be formalized to support consistent evaluation across architectures. Interpretability-aware design choices, such as structurally sparse layers, disentangled fusion modules, or white-box adapter components, may further facilitate attribution, mechanistic tracing, and controlled intervention.

\subsection{XAI Algorithms for Modeling Multimodal Interactions\label{sec:mm_interactions}}

Despite substantial progress in the development of XAI techniques, most existing algorithms remain limited in their ability to capture the complex interactions that characterize multimodal models. This is particularly evident in the case of attention-based methods, by far the most common explanation strategy observed in this review (34.6\% of the total algorithm papers), whose validity as causal or faithful attribution tools remains contested \cite{jain2019attention, serrano2019attention, bai2021attentions}.

Multimodal interactions typically occur at three levels: (1) between tokens within individual modalities, (2) between token pairs from different modalities, and (3) between grouped interactions involving multiple tokens and modalities. Current attention-based and composite methods can only effectively capture the first two types, while model-agnostic methods tend to model only intra-modality dynamics at a higher cost of computation. This algorithmic limitations are amplified with MLLMs, where inter-modality reasoning spans several layers, heads and expert modules. 

Emerging approaches such as self-explanations, particularly those enabled by MLLMs, offer high-level natural language justifications. However, as these are also generated from opaque, black-box systems, they lack the transparency and reliability in tracing model-internal decision paths. Mechanistic interpretability methods, though promising, also suffer from issues with scaling the circuit analyses that often do not generalize, remapping tokens often deeply fused within internal circuitry back to human concepts, and in general designing causal intervention that work for multimodal problems \cite{lin2025survey}. 

Recent advances in model-agnostic methods, including Shapley-based interaction measures such as InterSHAP and MultiSHAP, and disentangled multimodal attribution methods such as DIME, highlight promising directions. These techniques enable decomposition of unimodal effects, cross-modal interactions, and higher-order synergy terms without requiring explicit labels for interactions. Continued development of optimization-aware implementations is essential for enabling these methods to scale to both local (per-sample) and global (dataset-level) explanations in wider applications. 

Mechanistic approaches offer complementary benefits. Techniques such as CMA, attention-head gating, activation patching, structured ablations, and probing can be adapted more systematically to multimodal settings. Furthermore, neurosymbolic reasoning and NLEs can be integrated with mechanistic insights to generate both human-interpretable and computationally faithful explanations. Overall, effective multimodal XAI must move toward holistic, multi-level modeling of interactions, from token-to-token dynamics to cross-modal circuits.

Although interpretability requires more than any single analytical lens, attention-based tools remain valuable for exploratory analysis. They support rapid hypothesis formation regarding representational structure and are often computationally efficient. Moreover, in some settings, learned attention patterns have been shown to correlate with linguistically or semantically meaningful reasoning \cite{clark2019does, wiegreffe2019attention}. As with human attentional focus \cite{posner1980orienting}, machine attention can offer insight even when it is not strictly causal, provided it is interpreted with methodological caution.

\subsection{Cognition-Aware and Domain-Aware Modality Fusion, User-Centric Explanations}

Existing multimodal XAI algorithms often assume equal contribution from all input modalities, an assumption that contradicts well-established principles of human cognition. Research in neuroscience and psychology has shown that different sensory modalities contribute unequally to human decision-making depending on context and individual differences \cite{sheppard2013dynamic}. Furthermore, the criteria for interpretability are known to vary significantly across application domains \cite{carvalho2019machine}.

Future work should focus on developing cognition- and domain-aware fusion strategies that account for such variability. Weighted or conditional fusion methods, such as the frameworks introduced by Huang \emph{et al.} \cite{huang_generic_2023} and He \emph{et al.} \cite{he2024efficient}, demonstrate how modality contributions can be modulated based on context-dependent reliability. These frameworks may be extended to incorporate cognitive priors from multisensory perception, for example, reliability weighting, cross-modal bias, or temporal alignment constraints. To address domain variability, Islam \emph{et al.} introduced a model-agnostic,
weighted framework for quantifying explanations that can be applied across different domains \cite{islam2020towards}. Although promising, both these classes of techniques are currently very limited to specific use-cases. Significant research should be carried out in this area for cognition-, task- and domain-aware modality fusion for generating meaningful and context-sensitive explanations. Incorporating inductive biases informed by human perception or expert reasoning, combined with application-grounded interpretability protocols, represents a promising direction.

Another critical yet often overlooked consideration is the trustworthiness and usefulness of explanations \cite{nauta2023anecdotal, carvalho2019machine}. Neurosymbolic and structured reasoning pipelines support alignment of model justifications with human reasoning patterns while preserving model flexibility. Likewise, interactive and personalized explanation interfaces that allow users to probe, compare, and validate explanations across modalities can strengthen trust, enhance understanding, and support decision-making \cite{dang2024explainable}. Multimodal explainability interfaces, such as MultiViz \cite{liang2022multiviz}, exemplify this approach.

\subsection{Explainability as a Core Design Objective}

Despite advancements in both algorithmic and evaluation methods, explainability is still often treated as an afterthought rather than a fundamental design criterion. Foundational works have emphasized the value of model-agnostic frameworks \cite{carvalho2019machine} and application-grounded evaluations \cite{doshi2017towards} for achieving robust explainability. Yet, adoption of these recommendations remains sparse, particularly in multimodal contexts.

Algorithmically, many studies limit their scope to partial explanations, either covering a subset of modalities or analyzing each modality in isolation, typically using attention-guided, model-specific techniques. Only a few approaches address inter-modality interactions explicitly \cite{huang_generic_2023, chefer_generic_2021}. From an evaluation standpoint, most works rely on ad hoc qualitative analysis, with only a minority incorporating both human-centered and objective assessments (only 17.19\% of the total in this review, as seen in Figure \ref{fig:eval_venn}).

We emphasize that any system labeled as ``explainable" must undergo extensive experimentation for both algorithmic and evaluation aspects of explainability, at the very least. As Doshi-Velez and Kim argue, multiple dimensions of explainability, such as target audience, formulation of interpretability, evaluation level, task-related factors, and XAI interface, should be clearly documented \cite{doshi2017towards}. Ideally, a standardized reporting guideline specific to explainability should be developed, drawing inspiration from frameworks such as CONSORT-AI \cite{liu2020reporting}, to promote reliable and reproducible adoption of XAI practices.

From a technical perspective, training methods can be be adjusted to align to better rationales, for example, through differentiable objectives for regularizing explanations, alignment losses or synergy regularizers. For self-explaining model, complementary loops for verifying first and then reporting the explanations can be adopted. In each such experimentation, however, appropriate ablation studies should be reported to comprehensively analyze the nuanced relationships between explanations and modeling design choices.

\subsection{Towards Deeper and More Systematic Evaluation}

Evaluating explanations remains a challenging task due to the inherently subjective nature of interpretability and the diversity of applications and domains in which multimodal models are deployed. While several comprehensive evaluation frameworks exist \cite{nauta2023anecdotal, carvalho2019machine, hedstrom2023quantus, doshi2017towards}, very few studies apply more than one objective criterion and none address multimodal interaction metrics specifically. For example, the Quantus toolkit organizes existing objective evaluation methods into six logical categories to facilitate reproducible and accessible assessments \cite{hedstrom2023quantus}. However, in this review, only four of those categories were used, and evaluation methods outside the \textbf{faithfulness} class were rarely applied (only 8 out of 52 studies, or approximately 15.38\%).

The problem is amplified in emerging areas such as NLE generation for LLMs, where evaluation remains inconsistent and often disconnected from model performance \cite{cambria2023survey, cambria2024xai, bilal2025llms, palikhe2025towards}. Similarly, mechanistic interpretability in multimodal models suffers from the absence of standardized benchmarks or ground-truth circuit structures, leading to reliance on qualitative claims rather than systematic intervention-based validation \cite{lin2025survey}.
We encourage the integration of methodologies from adjacent disciplines such as cognitive science \cite{taylor2021artificial}, HCI \cite{lawless2019artificial, wang2019designing}, and statistics \cite{rohe2015cortical} to support deeper, more grounded evaluation of multimodal models. This can involve creating benchmark datasets expressly designed for evaluating multimodal reasoning and interaction-dependent explanations, which can support various explanation methods. Furthermore, evaluations should adopt multi-axis reporting across objective categories \cite{kazmierczak2025explainability}. Toolkits such as Quantus facilitate several of these tests and should be extended with multimodal primitives. In addition, cross-category metrics, as discussed before, may be very useful in multidimensional evaluation, giving the ability to create balance between often contrasting aspects of explanations.

Finally, human-centered evaluations remain largely unsystematic. Most rely on informal qualitative feedback, while those that involve quantitative studies often lack adherence to established protocols. Given the crucial role of human users in the interpretability loop, we echo Vilone and Longo \cite{vilone2021notions} in stressing the importance of standardized, domain-sensitive human evaluations that account for application context and modality-specific cognitive load. Following Doshi-Velez and Kim's framework \cite{doshi2017towards}, human studies should combine objective and subjective measures and draw from emerging best-practices from areas such as VL model trustworthiness \cite{vatsa2024adventures} or LLM rationale evaluation \cite{kunz2024properties} suitably adapted to multimodal settings.

\section{Conclusion}\label{Conclusion}
In this work, we have provided a comprehensive overview of explainability in multimodal attention-based models. This overview was motivated by the growing gap between explainability and the rapid advancements in modern AI applications. First, we systematically collected relevant publications in the area and analyzed them based on application tasks, domains, and evaluation datasets. We then identified the architectural variants used in the multimodal literature, with a particular focus on the fusion methods employed for combining input modalities. Based on this, we classified the models into four fusion categories and outlined the prevailing trends and limitations in architectural choices. Subsequently, we conducted an in-depth discussion of the selected publications along two key dimensions of explainability: explanation algorithms and evaluation methods. We adopted and extended a taxonomy grounded in prior research to categorize studies and highlight their distribution across these dimensions. Additionally, we briefly reviewed studies that proposed interfaces for explainability. Finally, we complement the discussion with advances across key axes of analysis beyond the performed systematic review and provide detailed recommendations along with outlined future research directions based on our findings.

Our analysis reveals that, while there has been a significant body of work on multimodal explainability, the adoption of tools and methods—many of which stem from an already incomplete field of explainability—still requires considerable refinement for effective use in multimodal scenarios. This observation holds true across all examined dimensions: application-architecture, explanation algorithm, and evaluation strategy. The widespread but inconsistent use of the term ``explainable solution'' has further hindered progress toward standardization in the field.

As this review emphasizes, explainability must be treated as a rigorous scientific objective rather than an optional afterthought. Achieving trustworthy multimodal AI requires standardized architectural abstractions, explicit modeling of inter-modality interactions, scalable and faithful explanation algorithms, and systematic evaluation protocols grounded in both objective metrics and human-centered studies. We encourage the community to work toward a unified multimodal taxonomy encompassing architecture patterns, interaction structures, explanation modalities, and evaluation frameworks. Ultimately, as multimodal models continue to grow in capability and societal impact, explainability should be integrated as a core design principle. We hope that the insights and recommendations presented in this work will contribute to more transparent, interpretable, and reliable multimodal AI systems.

\section{CRediT authorship contribution statement}
\textbf{Md Raisul Kibria:} Conceptualization, Methodology, Investigation, Visualization, Writing - original draft. \textbf{Sébastien Lafond:} Supervision, Conceptualization, Methodology, Writing
- review \& editing. \textbf{Janan Arslan:} Supervision, Methodology, Writing - review \& editing.

\section{Declaration of Competing Interest}
The authors declare that they have no known competing financial interests or personal relationships that could have appeared to influence the work reported in this paper.

\section{Declaration of Generative AI and AI-assisted technologies in the
writing process}

During the preparation of the manuscript, the authors utilized generative AI tools, such as ChatGPT, to enhance the language and readability of selected sections. The process was undertaken in accordance with established policies, under the authors' strict guidance (through review and editing), and the authors bear full responsibility for the content of the publication.



\bibliographystyle{elsarticle-num} 
%
\bibliography{revised_bib}

@inproceedings{lin2014microsoft,
  title={Microsoft coco: Common objects in context},
  author={Lin, Tsung-Yi and Maire, Michael and Belongie, Serge and Hays, James and Perona, Pietro and Ramanan, Deva and Doll{\'a}r, Piotr and Zitnick, C Lawrence},
  booktitle={Computer Vision--ECCV 2014: 13th European Conference, Zurich, Switzerland, September 6-12, 2014, Proceedings, Part V 13},
  pages={740--755},
  year={2014},
  organization={Springer}
}

@inproceedings{wang-etal-2018-glue,
    title = "{GLUE}: A Multi-Task Benchmark and Analysis Platform for Natural Language Understanding",
    author = "Wang, Alex  and
      Singh, Amanpreet  and
      Michael, Julian  and
      Hill, Felix  and
      Levy, Omer  and
      Bowman, Samuel",
    editor = "Linzen, Tal  and
      Chrupa{\l}a, Grzegorz  and
      Alishahi, Afra",
    booktitle = "Proceedings of the 2018 {EMNLP} Workshop {B}lackbox{NLP}: Analyzing and Interpreting Neural Networks for {NLP}",
    month = nov,
    year = "2018",
    address = "Brussels, Belgium",
    publisher = "Association for Computational Linguistics",
    url = "https://aclanthology.org/W18-5446",
    doi = "10.18653/v1/W18-5446",
    pages = "353--355",
}

@inproceedings{pelka2018radiology,
  title={Radiology objects in context (roco): a multimodal image dataset},
  author={Pelka, Obioma and Koitka, Sven and R{\"u}ckert, Johannes and Nensa, Felix and Friedrich, Christoph M},
  booktitle={7th Joint International Workshop, CVII-STENT 2018 and Third International Workshop, LABELS 2018, Held in Conjunction with MICCAI 2018, Granada, Spain, September 16, 2018, Proceedings 3},
  pages={180--189},
  year={2018},
  organization={Springer}
}

@inproceedings{antol2015vqa,
  title={Vqa: Visual question answering},
  author={Antol, Stanislaw and Agrawal, Aishwarya and Lu, Jiasen and Mitchell, Margaret and Batra, Dhruv and Zitnick, C Lawrence and Parikh, Devi},
  booktitle={Proceedings of the IEEE international conference on computer vision},
  pages={2425--2433},
  year={2015}
}

@article{krishna2017visual,
  title={Visual genome: Connecting language and vision using crowdsourced dense image annotations},
  author={Krishna, Ranjay and Zhu, Yuke and Groth, Oliver and Johnson, Justin and Hata, Kenji and Kravitz, Joshua and Chen, Stephanie and Kalantidis, Yannis and Li, Li-Jia and Shamma, David A and others},
  journal={International journal of computer vision},
  volume={123},
  pages={32--73},
  year={2017},
  publisher={Springer}
}

@article{johnson2016mimic,
  title={MIMIC-III, a freely accessible critical care database},
  author={Johnson, Alistair EW and Pollard, Tom J and Shen, Lu and Lehman, Li-wei H and Feng, Mengling and Ghassemi, Mohammad and Moody, Benjamin and Szolovits, Peter and Anthony Celi, Leo and Mark, Roger G},
  journal={Scientific data},
  volume={3},
  number={1},
  pages={1--9},
  year={2016},
  publisher={Nature Publishing Group}
}

@article{vaswani2017attention,
  title={Attention is all you need},
  author={Vaswani, A},
  journal={Advances in Neural Information Processing Systems},
  year={2017}
}

@article{dosovitskiy2020image,
  title={An image is worth 16x16 words: Transformers for image recognition at scale},
  author={Dosovitskiy, Alexey},
  journal={arXiv preprint arXiv:2010.11929},
  year={2020}
}

@inproceedings{liu2022video,
  title={Video swin transformer},
  author={Liu, Ze and Ning, Jia and Cao, Yue and Wei, Yixuan and Zhang, Zheng and Lin, Stephen and Hu, Han},
  booktitle={Proceedings of the IEEE/CVF conference on computer vision and pattern recognition},
  pages={3202--3211},
  year={2022}
}

@article{xu2023multimodal,
  title={Multimodal learning with transformers: A survey},
  author={Xu, Peng and Zhu, Xiatian and Clifton, David A},
  journal={IEEE Transactions on Pattern Analysis and Machine Intelligence},
  volume={45},
  number={10},
  pages={12113--12132},
  year={2023},
  publisher={IEEE}
}

@article{sengar2024generative,
  title={Generative Artificial Intelligence: A Systematic Review and Applications},
  author={Sengar, Sandeep Singh and Hasan, Affan Bin and Kumar, Sanjay and Carroll, Fiona},
  journal={arXiv preprint arXiv:2405.11029},
  year={2024}
}

@article{abnar2020quantifying,
  title={Quantifying attention flow in transformers},
  author={Abnar, Samira and Zuidema, Willem},
  journal={arXiv preprint arXiv:2005.00928},
  year={2020}
}

@article{nauta2023anecdotal,
  title={From anecdotal evidence to quantitative evaluation methods: A systematic review on evaluating explainable ai},
  author={Nauta, Meike and Trienes, Jan and Pathak, Shreyasi and Nguyen, Elisa and Peters, Michelle and Schmitt, Yasmin and Schl{\"o}tterer, J{\"o}rg and Van Keulen, Maurice and Seifert, Christin},
  journal={ACM Computing Surveys},
  volume={55},
  number={13s},
  pages={1--42},
  year={2023},
  publisher={ACM New York, NY}
}

@article{rodis2024multimodal,
  title={Multimodal explainable artificial intelligence: A comprehensive review of methodological advances and future research directions},
  author={Rodis, Nikolaos and Sardianos, Christos and Radoglou-Grammatikis, Panagiotis and Sarigiannidis, Panagiotis and Varlamis, Iraklis and Papadopoulos, Georgios Th},
  journal={IEEE Access},
  year={2024},
  publisher={IEEE}
}

@inproceedings{islam2020towards,
  title={Towards quantification of explainability in explainable artificial intelligence methods},
  author={Islam, Sheikh Rabiul and Eberle, William and Ghafoor, Sheikh K},
  booktitle={The thirty-third international flairs conference},
  year={2020}
}

@article{hedstrom2023quantus,
  title={Quantus: An explainable ai toolkit for responsible evaluation of neural network explanations and beyond},
  author={Hedstr{\"o}m, Anna and Weber, Leander and Krakowczyk, Daniel and Bareeva, Dilyara and Motzkus, Franz and Samek, Wojciech and Lapuschkin, Sebastian and H{\"o}hne, Marina M-C},
  journal={Journal of Machine Learning Research},
  volume={24},
  number={34},
  pages={1--11},
  year={2023}
}

@inproceedings{chefer2021transformer,
  title={Transformer interpretability beyond attention visualization},
  author={Chefer, Hila and Gur, Shir and Wolf, Lior},
  booktitle={Proceedings of the IEEE/CVF conference on computer vision and pattern recognition},
  pages={782--791},
  year={2021}
}

@article{alzubaidi2021review,
  title={Review of deep learning: concepts, CNN architectures, challenges, applications, future directions},
  author={Alzubaidi, Laith and Zhang, Jinglan and Humaidi, Amjad J and Al-Dujaili, Ayad and Duan, Ye and Al-Shamma, Omran and Santamar{\'\i}a, Jos{\'e} and Fadhel, Mohammed A and Al-Amidie, Muthana and Farhan, Laith},
  journal={Journal of big Data},
  volume={8},
  pages={1--74},
  year={2021},
  publisher={Springer}
}

@article{liu2022combining,
  title={Combining context-relevant features with multi-stage attention network for short text classification},
  author={Liu, Yingying and Li, Peipei and Hu, Xuegang},
  journal={Computer Speech \& Language},
  volume={71},
  pages={101268},
  year={2022},
  publisher={Elsevier}
}

@article{bahdanau2014neural,
  title={Neural machine translation by jointly learning to align and translate},
  author={Bahdanau, Dzmitry and Cho, Kyunghyun and Bengio, Yoshua},
  journal={arXiv preprint arXiv:1409.0473},
  year={2014}
}

@article{yang2022unbox,
  title={Unbox the black-box for the medical explainable AI via multi-modal and multi-centre data fusion: A mini-review, two showcases and beyond},
  author={Yang, Guang and Ye, Qinghao and Xia, Jun},
  journal={Information Fusion},
  volume={77},
  pages={29--52},
  year={2022},
  publisher={Elsevier}
}

@article{arrieta2020explainable,
  title={Explainable Artificial Intelligence (XAI): Concepts, taxonomies, opportunities and challenges toward responsible AI},
  author={Arrieta, Alejandro Barredo and D{\'\i}az-Rodr{\'\i}guez, Natalia and Del Ser, Javier and Bennetot, Adrien and Tabik, Siham and Barbado, Alberto and Garc{\'\i}a, Salvador and Gil-L{\'o}pez, Sergio and Molina, Daniel and Benjamins, Richard and others},
  journal={Information fusion},
  volume={58},
  pages={82--115},
  year={2020},
  publisher={Elsevier}
}

@inproceedings{gilpin2018explaining,
  title={Explaining explanations: An overview of interpretability of machine learning},
  author={Gilpin, Leilani H and Bau, David and Yuan, Ben Z and Bajwa, Ayesha and Specter, Michael and Kagal, Lalana},
  booktitle={2018 IEEE 5th International Conference on data science and advanced analytics (DSAA)},
  pages={80--89},
  year={2018},
  organization={IEEE}
}

@inproceedings{nannini2023explainability,
  title={Explainability in AI policies: A critical review of communications, reports, regulations, and standards in the EU, US, and UK},
  author={Nannini, Luca and Balayn, Agathe and Smith, Adam Leon},
  booktitle={Proceedings of the 2023 ACM conference on fairness, accountability, and transparency},
  pages={1198--1212},
  year={2023}
}

@article{guidotti2018survey,
  title={A survey of methods for explaining black box models},
  author={Guidotti, Riccardo and Monreale, Anna and Ruggieri, Salvatore and Turini, Franco and Giannotti, Fosca and Pedreschi, Dino},
  journal={ACM computing surveys (CSUR)},
  volume={51},
  number={5},
  pages={1--42},
  year={2018},
  publisher={ACM New York, NY, USA}
}

@article{saxena2021generative,
  title={Generative adversarial networks (GANs) challenges, solutions, and future directions},
  author={Saxena, Divya and Cao, Jiannong},
  journal={ACM Computing Surveys (CSUR)},
  volume={54},
  number={3},
  pages={1--42},
  year={2021},
  publisher={ACM New York, NY, USA}
}

@article{parcalabescu2022mm,
  title={Mm-shap: A performance-agnostic metric for measuring multimodal contributions in vision and language models \& tasks},
  author={Parcalabescu, Letitia and Frank, Anette},
  journal={arXiv preprint arXiv:2212.08158},
  year={2022}
}

@article{lawless2019artificial,
  title={Artificial intelligence, autonomy, and human-machine teams—interdependence, context, and explainable AI},
  author={Lawless, William F and Mittu, Ranjeev and Sofge, Don and Hiatt, Laura},
  journal={Ai Magazine},
  volume={40},
  number={3},
  pages={5--13},
  year={2019}
}

@inproceedings{wang2019designing,
  title={Designing theory-driven user-centric explainable AI},
  author={Wang, Danding and Yang, Qian and Abdul, Ashraf and Lim, Brian Y},
  booktitle={Proceedings of the 2019 CHI conference on human factors in computing systems},
  pages={1--15},
  year={2019}
}

@article{rohe2015cortical,
  title={Cortical hierarchies perform Bayesian causal inference in multisensory perception},
  author={Rohe, Tim and Noppeney, Uta},
  journal={PLoS biology},
  volume={13},
  number={2},
  pages={e1002073},
  year={2015},
  publisher={Public Library of Science San Francisco, CA USA}
}

@article{sheppard2013dynamic,
  title={Dynamic weighting of multisensory stimuli shapes decision-making in rats and humans},
  author={Sheppard, John P and Raposo, David and Churchland, Anne K},
  journal={Journal of vision},
  volume={13},
  number={6},
  pages={4--4},
  year={2013},
  publisher={The Association for Research in Vision and Ophthalmology}
}

@article{taylor2021artificial,
  title={Artificial cognition: How experimental psychology can help generate explainable artificial intelligence},
  author={Taylor, J Eric T and Taylor, Graham W},
  journal={Psychonomic Bulletin \& Review},
  volume={28},
  number={2},
  pages={454--475},
  year={2021},
  publisher={Springer}
}

@article{serrano2019attention,
  title={Is attention interpretable?},
  author={Serrano, Sofia and Smith, Noah A},
  journal={arXiv preprint arXiv:1906.03731},
  year={2019}
}

@inproceedings{bai2021attentions,
  title={Why attentions may not be interpretable?},
  author={Bai, Bing and Liang, Jian and Zhang, Guanhua and Li, Hao and Bai, Kun and Wang, Fei},
  booktitle={Proceedings of the 27th ACM SIGKDD conference on knowledge discovery \& data mining},
  pages={25--34},
  year={2021}
}

@article{liu2020reporting,
  title={Reporting guidelines for clinical trial reports for interventions involving artificial intelligence: the CONSORT-AI extension},
  author={Liu, Xiaoxuan and Rivera, Samantha Cruz and Moher, David and Calvert, Melanie J and Denniston, Alastair K and Ashrafian, Hutan and Beam, Andrew L and Chan, An-Wen and Collins, Gary S and Deeks, Ara DarziJonathan J and others},
  journal={The Lancet Digital Health},
  volume={2},
  number={10},
  pages={e537--e548},
  year={2020},
  publisher={Elsevier}
}

@inproceedings{anderson2016spice,
  title={Spice: Semantic propositional image caption evaluation},
  author={Anderson, Peter and Fernando, Basura and Johnson, Mark and Gould, Stephen},
  booktitle={Computer Vision--ECCV 2016: 14th European Conference, Amsterdam, The Netherlands, October 11-14, 2016, Proceedings, Part V 14},
  pages={382--398},
  year={2016},
  organization={Springer}
}

@inproceedings{lin2004rouge,
  title={Rouge: A package for automatic evaluation of summaries},
  author={Lin, Chin-Yew},
  booktitle={Text summarization branches out},
  pages={74--81},
  year={2004}
}

@inproceedings{papineni2002bleu,
  title={Bleu: a method for automatic evaluation of machine translation},
  author={Papineni, Kishore and Roukos, Salim and Ward, Todd and Zhu, Wei-Jing},
  booktitle={Proceedings of the 40th annual meeting of the Association for Computational Linguistics},
  pages={311--318},
  year={2002}
}

@inproceedings{denkowski2014meteor,
  title={Meteor universal: Language specific translation evaluation for any target language},
  author={Denkowski, Michael and Lavie, Alon},
  booktitle={Proceedings of the ninth workshop on statistical machine translation},
  pages={376--380},
  year={2014}
}

@inproceedings{vedantam2015cider,
  title={Cider: Consensus-based image description evaluation},
  author={Vedantam, Ramakrishna and Lawrence Zitnick, C and Parikh, Devi},
  booktitle={Proceedings of the IEEE conference on computer vision and pattern recognition},
  pages={4566--4575},
  year={2015}
}

@inproceedings{wohlin2014guidelines,
  title={Guidelines for snowballing in systematic literature studies and a replication in software engineering},
  author={Wohlin, Claes},
  booktitle={Proceedings of the 18th international conference on evaluation and assessment in software engineering},
  pages={1--10},
  year={2014}
}

@article{chen2015microsoft,
  title={Microsoft coco captions: Data collection and evaluation server},
  author={Chen, Xinlei and Fang, Hao and Lin, Tsung-Yi and Vedantam, Ramakrishna and Gupta, Saurabh and Doll{\'a}r, Piotr and Zitnick, C Lawrence},
  journal={arXiv preprint arXiv:1504.00325},
  year={2015}
}

@inproceedings{wenderoth2025measuring,
  title={Measuring cross-modal interactions in multimodal models},
  author={Wenderoth, Laura and Hemker, Konstantin and Simidjievski, Nikola and Jamnik, Mateja},
  booktitle={Proceedings of the AAAI Conference on Artificial Intelligence},
  volume={39},
  number={20},
  pages={21501--21509},
  year={2025}
}

@article{liang2022high,
  title={High-modality multimodal transformer: Quantifying modality \& interaction heterogeneity for high-modality representation learning},
  author={Liang, Paul Pu and Lyu, Yiwei and Fan, Xiang and Tsaw, Jeffrey and Liu, Yudong and Mo, Shentong and Yogatama, Dani and Morency, Louis-Philippe and Salakhutdinov, Ruslan},
  journal={arXiv preprint arXiv:2203.01311},
  year={2022}
}

@inproceedings{jeong2024multimodal,
  title={Multimodal image-text matching improves retrieval-based chest x-ray report generation},
  author={Jeong, Jaehwan and Tian, Katherine and Li, Andrew and Hartung, Sina and Adithan, Subathra and Behzadi, Fardad and Calle, Juan and Osayande, David and Pohlen, Michael and Rajpurkar, Pranav},
  booktitle={Medical Imaging with Deep Learning},
  pages={978--990},
  year={2024},
  organization={PMLR}
}

@article{agarwal2025rethinking,
  title={Rethinking Explainability in the Era of Multimodal AI},
  author={Agarwal, Chirag},
  journal={arXiv preprint arXiv:2506.13060},
  year={2025}
}

@article{kitchenham2004procedures,
  title={Procedures for performing systematic reviews},
  author={Kitchenham, Barbara},
  journal={Keele, UK, Keele University},
  volume={33},
  number={2004},
  pages={1--26},
  year={2004},
  publisher={Citeseer}
}

@article{moher2009preferred,
  title={Preferred reporting items for systematic reviews and meta-analyses: the PRISMA statement},
  author={Moher, David and Liberati, Alessandro and Tetzlaff, Jennifer and Altman, Douglas G},
  journal={Bmj},
  volume={339},
  year={2009},
  publisher={British Medical Journal Publishing Group}
}

@article{altmae2023artificial,
  title={Artificial intelligence in scientific writing: a friend or a foe?},
  author={Altm{\"a}e, Signe and Sola-Leyva, Alberto and Salumets, Andres},
  journal={Reproductive BioMedicine Online},
  volume={47},
  number={1},
  pages={3--9},
  year={2023},
  publisher={Elsevier}
}

@article{dagdelen2024structured,
  title={Structured information extraction from scientific text with large language models},
  author={Dagdelen, John and Dunn, Alexander and Lee, Sanghoon and Walker, Nicholas and Rosen, Andrew S and Ceder, Gerbrand and Persson, Kristin A and Jain, Anubhav},
  journal={Nature Communications},
  volume={15},
  number={1},
  pages={1418},
  year={2024},
  publisher={Nature Publishing Group UK London}
}

@inproceedings{huotala2024promise,
  title={The promise and challenges of using LLMs to accelerate the screening process of systematic reviews},
  author={Huotala, Aleksi and Kuutila, Miikka and Ralph, Paul and M{\"a}ntyl{\"a}, Mika},
  booktitle={Proceedings of the 28th International Conference on Evaluation and Assessment in Software Engineering},
  pages={262--271},
  year={2024}
}

@inproceedings{felizardo2024chatgpt,
  title={ChatGPT application in Systematic Literature Reviews in Software Engineering: an evaluation of its accuracy to support the selection activity},
  author={Felizardo, Katia Romero and Lima, M{\'a}rcia Sampaio and Deizepe, Anderson and Conte, Tayana Uch{\^o}a and Steinmacher, Igor},
  booktitle={Proceedings of the 18th ACM/IEEE International Symposium on Empirical Software Engineering and Measurement},
  pages={25--36},
  year={2024}
}

@article{grattafiori2024llama,
  title={The llama 3 herd of models},
  author={Grattafiori, Aaron and Dubey, Abhimanyu and Jauhri, Abhinav and Pandey, Abhinav and Kadian, Abhishek and Al-Dahle, Ahmad and Letman, Aiesha and Mathur, Akhil and Schelten, Alan and Vaughan, Alex and others},
  journal={arXiv preprint arXiv:2407.21783},
  year={2024}
}

@article{wei2022chain,
  title={Chain-of-thought prompting elicits reasoning in large language models},
  author={Wei, Jason and Wang, Xuezhi and Schuurmans, Dale and Bosma, Maarten and Xia, Fei and Chi, Ed and Le, Quoc V and Zhou, Denny and others},
  journal={Advances in neural information processing systems},
  volume={35},
  pages={24824--24837},
  year={2022}
}

@article{janssens_360_2024,
	title = {360 {Degrees} rumor detection: {When} explanations got some explaining to do},
	volume = {317},
	issn = {03772217},
	shorttitle = {360 {Degrees} rumor detection},
	url = {https://linkinghub.elsevier.com/retrieve/pii/S0377221723004769},
	doi = {10.1016/j.ejor.2023.06.024},
	language = {en},
	number = {2},
	urldate = {2025-01-29},
	journal = {European Journal of Operational Research},
	author = {Janssens, Bram and Schetgen, Lisa and Bogaert, Matthias and Meire, Matthijs and Van Den Poel, Dirk},
	month = sep,
	year = {2024},
	pages = {366--381},
}

@article{kumar_bert_2022,
	title = {A {BERT} based dual-channel explainable text emotion recognition system},
	volume = {150},
	issn = {08936080},
	url = {https://linkinghub.elsevier.com/retrieve/pii/S0893608022000958},
	doi = {10.1016/j.neunet.2022.03.017},
	language = {en},
	urldate = {2025-01-29},
	journal = {Neural Networks},
	author = {Kumar, Puneet and Raman, Balasubramanian},
	month = jun,
	year = {2022},
	pages = {392--407},
}

@inproceedings{du_case-based_2023,
	address = {Chenzhou China},
	title = {A {Case}-based {Channel} {Selection} {Method} for {EEG} {Emotion} {Recognition} {Using} {Interpretable} {Transformer} {Networks}},
	isbn = {9798400708701},
	doi = {10.1145/3627341.3630372},
	language = {en},
	urldate = {2025-01-29},
	booktitle = {Proceedings of the 2023 {International} {Conference} on {Computer}, {Vision} and {Intelligent} {Technology}},
	publisher = {ACM},
	author = {Du, Yang and Guan, Zijing and Huang, Weichen and Zhang, Xichun and Huang, Qiyun},
	month = aug,
	year = {2023},
	pages = {1--5},
}

@incollection{wang_novel_2023,
	address = {Cham},
	title = {A {Novel} {Deep} {Learning} {Framework} for {Interpretable} {Drug}-{Target} {Interaction} {Prediction} with {Attention} and {Multi}-task {Mechanism}},
	volume = {13946},
	isbn = {978-3-031-30677-8 978-3-031-30678-5},
	url = {https://link.springer.com/10.1007/978-3-031-30678-5_26},
	language = {en},
	urldate = {2025-01-29},
	booktitle = {Database {Systems} for {Advanced} {Applications}},
	publisher = {Springer Nature Switzerland},
	author = {Zheng, Yubin and Tang, Peng and Qiu, Weidong and Wang, Hao and Guo, Jie and Huang, Zheng},
	editor = {Wang, Xin and Sapino, Maria Luisa and Han, Wook-Shin and El Abbadi, Amr and Dobbie, Gill and Feng, Zhiyong and Shao, Yingxiao and Yin, Hongzhi},
	year = {2023},
	doi = {10.1007/978-3-031-30678-5_26},
	note = {Series Title: Lecture Notes in Computer Science},
	pages = {336--352},
}

@misc{bhargava_adaptive_2020,
	title = {Adaptive {Transformers} for {Learning} {Multimodal} {Representations}},
	url = {http://arxiv.org/abs/2005.07486},
	doi = {10.48550/arXiv.2005.07486},
	language = {en},
	urldate = {2025-01-29},
	publisher = {arXiv},
	author = {Bhargava, Prajjwal},
	month = jul,
	year = {2020},
	note = {arXiv:2005.07486 [cs]},
}

@inproceedings{guo_explainable_2023,
	address = {Haikou, China},
	title = {An {Explainable} {Recommendation} {Method} based on {Diffusion} {Model}},
	copyright = {https://doi.org/10.15223/policy-029},
	isbn = {9798350330076},
	url = {https://ieeexplore.ieee.org/document/10429319/},
	doi = {10.1109/BigDIA60676.2023.10429319},
	language = {en},
	urldate = {2025-01-29},
	booktitle = {2023 9th {International} {Conference} on {Big} {Data} and {Information} {Analytics} ({BigDIA})},
	publisher = {IEEE},
	author = {Guo, Yupu and Cai, Fei and Chen, Honghui and Chen, Chonghao and Zhang, Xin and Zhang, Menxi},
	month = dec,
	year = {2023},
	pages = {802--806},
}

@article{yang_explainable_2023,
	title = {An {Explainable} {Spatial}–{Frequency} {Multiscale} {Transformer} for {Remote} {Sensing} {Scene} {Classification}},
	volume = {61},
	copyright = {https://ieeexplore.ieee.org/Xplorehelp/downloads/license-information/IEEE.html},
	issn = {0196-2892, 1558-0644},
	url = {https://ieeexplore.ieee.org/document/10097579/},
	doi = {10.1109/TGRS.2023.3265361},
	language = {en},
	urldate = {2025-01-29},
	journal = {IEEE Transactions on Geoscience and Remote Sensing},
	author = {Yang, Yuting and Jiao, Licheng and Liu, Fang and Liu, Xu and Li, Lingling and Chen, Puhua and Yang, Shuyuan},
	year = {2023},
	pages = {1--15},
}

@article{qiang_attcat_nodate,
	title = {{AttCAT}: {Explaining} {Transformers} via {Attentive} {Class} {Activation} {Tokens}},
	language = {en},
	author = {Qiang, Yao and Pan, Deng and Li, Chengyin and Li, Xin and Jang, Rhongho and Zhu, Dongxiao},
}

@article{koyama_attention_2023,
	title = {Attention network for predicting {T}-cell receptor–peptide binding can associate attention with interpretable protein structural properties},
	volume = {3},
	issn = {2673-7647},
	url = {https://www.frontiersin.org/articles/10.3389/fbinf.2023.1274599/full},
	doi = {10.3389/fbinf.2023.1274599},
	language = {en},
	urldate = {2025-01-29},
	journal = {Frontiers in Bioinformatics},
	author = {Koyama, Kyohei and Hashimoto, Kosuke and Nagao, Chioko and Mizuguchi, Kenji},
	month = dec,
	year = {2023},
	pages = {1274599},
}

@misc{ferrando_attention_2021,
	title = {Attention {Weights} in {Transformer} {NMT} {Fail} {Aligning} {Words} {Between} {Sequences} but {Largely} {Explain} {Model} {Predictions}},
	url = {http://arxiv.org/abs/2109.05853},
	doi = {10.48550/arXiv.2109.05853},
	language = {en},
	urldate = {2025-01-29},
	publisher = {arXiv},
	author = {Ferrando, Javier and Costa-jussà, Marta R.},
	month = sep,
	year = {2021},
	note = {arXiv:2109.05853 [cs]},
}

@article{rigotti_attention-based_2022,
	title = {{ATTENTION}-{BASED} {INTERPRETABILITY} {WITH} {CONCEPT} {TRANSFORMERS}},
	language = {en},
	author = {Rigotti, Mattia and Miksovic, Christoph and Giurgiu, Ioana and Gschwind, Thomas and Scotton, Paolo},
	year = {2022},
}

@incollection{abdulkadir_augmenting_2022,
	address = {Cham},
	title = {Augmenting {Magnetic} {Resonance} {Imaging} with {Tabular} {Features} for {Enhanced} and {Interpretable} {Medial} {Temporal} {Lobe} {Atrophy} {Prediction}},
	volume = {13596},
	isbn = {978-3-031-17898-6 978-3-031-17899-3},
	url = {https://link.springer.com/10.1007/978-3-031-17899-3_13},
	language = {en},
	urldate = {2025-01-29},
	booktitle = {Machine {Learning} in {Clinical} {Neuroimaging}},
	publisher = {Springer Nature Switzerland},
	author = {Lee, Dongsoo and Suh, Chong Hyun and Kim, Jinyoung and Jung, Wooseok and Park, Changhyun and Jung, Kyu-Hwan and Kong, Seo Taek and Shim, Woo Hyun and Heo, Hwon and Kim, Sang Joon},
	editor = {Abdulkadir, Ahmed and Bathula, Deepti R. and Dvornek, Nicha C. and Habes, Mohamad and Kia, Seyed Mostafa and Kumar, Vinod and Wolfers, Thomas},
	year = {2022},
	doi = {10.1007/978-3-031-17899-3_13},
	note = {Series Title: Lecture Notes in Computer Science},
	pages = {125--134},
}

@article{meng_bidirectional_2021,
	title = {Bidirectional {Representation} {Learning} {From} {Transformers} {Using} {Multimodal} {Electronic} {Health} {Record} {Data} to {Predict} {Depression}},
	volume = {25},
	copyright = {https://ieeexplore.ieee.org/Xplorehelp/downloads/license-information/IEEE.html},
	issn = {2168-2194, 2168-2208},
	url = {https://ieeexplore.ieee.org/document/9369833/},
	doi = {10.1109/JBHI.2021.3063721},
	language = {en},
	number = {8},
	urldate = {2025-01-29},
	journal = {IEEE Journal of Biomedical and Health Informatics},
	author = {Meng, Yiwen and Speier, William and Ong, Michael K. and Arnold, Corey W.},
	month = aug,
	year = {2021},
	pages = {3121--3129},
}

@article{ding_deepstf_2023,
	title = {{DeepSTF}: predicting transcription factor binding sites by interpretable deep neural networks combining sequence and shape},
	volume = {24},
	copyright = {https://academic.oup.com/pages/standard-publication-reuse-rights},
	issn = {1467-5463, 1477-4054},
	shorttitle = {{DeepSTF}},
	url = {https://academic.oup.com/bib/article/doi/10.1093/bib/bbad231/7199560},
	doi = {10.1093/bib/bbad231},
	language = {en},
	number = {4},
	urldate = {2025-01-29},
	journal = {Briefings in Bioinformatics},
	author = {Ding, Pengju and Wang, Yifei and Zhang, Xinyu and Gao, Xin and Liu, Guozhu and Yu, Bin},
	month = jul,
	year = {2023},
	pages = {bbad231},
}

@misc{feucht_description-based_2021,
	title = {Description-based {Label} {Attention} {Classifier} for {Explainable} {ICD}-9 {Classification}},
	url = {http://arxiv.org/abs/2109.12026},
	doi = {10.48550/arXiv.2109.12026},
	language = {en},
	urldate = {2025-01-29},
	publisher = {arXiv},
	author = {Feucht, Malte and Wu, Zhiliang and Althammer, Sophia and Tresp, Volker},
	month = sep,
	year = {2021},
	note = {arXiv:2109.12026 [cs]},
}

@article{sun_dfyolov5m-m2transformer_2023,
	title = {{DFYOLOv5m}-{M2transformer}: {Interpretation} of vegetable disease recognition results using image dense captioning techniques},
	volume = {215},
	issn = {01681699},
	shorttitle = {{DFYOLOv5m}-{M2transformer}},
	url = {https://linkinghub.elsevier.com/retrieve/pii/S0168169923008487},
	doi = {10.1016/j.compag.2023.108460},
	language = {en},
	urldate = {2025-01-29},
	journal = {Computers and Electronics in Agriculture},
	author = {Sun, Wei and Wang, Chunshan and Wu, Huarui and Miao, Yisheng and Zhu, Huaji and Guo, Wang and Li, Jiuxi},
	month = dec,
	year = {2023},
	pages = {108460},
}

@article{yu_ex-vit_2023,
	title = {{eX}-{ViT}: {A} {Novel} explainable vision transformer for weakly supervised semantic segmentation},
	volume = {142},
	issn = {00313203},
	shorttitle = {{eX}-{ViT}},
	url = {https://linkinghub.elsevier.com/retrieve/pii/S0031320323003679},
	doi = {10.1016/j.patcog.2023.109666},
	language = {en},
	urldate = {2025-01-29},
	journal = {Pattern Recognition},
	author = {Yu, Lu and Xiang, Wei and Fang, Juan and Chen, Yi-Ping Phoebe and Chi, Lianhua},
	month = oct,
	year = {2023},
	pages = {109666},
}

@article{buoy_explainable_2023,
	title = {Explainable {Connectionist}-{Temporal}-{Classification}-{Based} {Scene} {Text} {Recognition}},
	volume = {9},
	copyright = {https://creativecommons.org/licenses/by/4.0/},
	issn = {2313-433X},
	url = {https://www.mdpi.com/2313-433X/9/11/248},
	doi = {10.3390/jimaging9110248},
	language = {en},
	number = {11},
	urldate = {2025-01-29},
	journal = {Journal of Imaging},
	author = {Buoy, Rina and Iwamura, Masakazu and Srun, Sovila and Kise, Koichi},
	month = nov,
	year = {2023},
	pages = {248},
}

@inproceedings{chiewhawan_explainable_2020,
	address = {Singapore Singapore},
	title = {Explainable {Deep} {Learning} for {Thai} {Stock} {Market} {Prediction} {Using} {Textual} {Representation} and {Technical} {Indicators}},
	isbn = {978-1-4503-8766-8},
	url = {https://dl.acm.org/doi/10.1145/3411174.3411191},
	doi = {10.1145/3411174.3411191},
	language = {en},
	urldate = {2025-01-29},
	booktitle = {Proceedings of the 8th {International} {Conference} on {Computer} and {Communications} {Management}},
	publisher = {ACM},
	author = {Chiewhawan, Tanawat and Vateekul, Peerapon},
	month = jul,
	year = {2020},
	pages = {19--23},
}

@article{ullah_explainable_2022,
	title = {Explainable {Malware} {Detection} {System} {Using} {Transformers}-{Based} {Transfer} {Learning} and {Multi}-{Model} {Visual} {Representation}},
	volume = {22},
	copyright = {https://creativecommons.org/licenses/by/4.0/},
	issn = {1424-8220},
	url = {https://www.mdpi.com/1424-8220/22/18/6766},
	doi = {10.3390/s22186766},
	language = {en},
	number = {18},
	urldate = {2025-01-29},
	journal = {Sensors},
	author = {Ullah, Farhan and Alsirhani, Amjad and Alshahrani, Mohammed Mujib and Alomari, Abdullah and Naeem, Hamad and Shah, Syed Aziz},
	month = sep,
	year = {2022},
	pages = {6766},
}

@article{zhang_explainable_2022,
	title = {Explainable multimodal trajectory prediction using attention models},
	volume = {143},
	issn = {0968090X},
	url = {https://linkinghub.elsevier.com/retrieve/pii/S0968090X22002509},
	doi = {10.1016/j.trc.2022.103829},
	language = {en},
	urldate = {2025-01-29},
	journal = {Transportation Research Part C: Emerging Technologies},
	author = {Zhang, Kunpeng and Li, Li},
	month = oct,
	year = {2022},
	pages = {103829},
}

@inproceedings{mohammadkhani_explaining_2023,
	address = {Bogotá, Colombia},
	title = {Explaining {Transformer}-based {Code} {Models}: {What} {Do} {They} {Learn}? {When} {They} {Do} {Not} {Work}?},
	copyright = {https://doi.org/10.15223/policy-029},
	isbn = {9798350305067},
	shorttitle = {Explaining {Transformer}-based {Code} {Models}},
	url = {https://ieeexplore.ieee.org/document/10356671/},
	doi = {10.1109/SCAM59687.2023.00020},
	language = {en},
	urldate = {2025-01-29},
	booktitle = {2023 {IEEE} 23rd {International} {Working} {Conference} on {Source} {Code} {Analysis} and {Manipulation} ({SCAM})},
	publisher = {IEEE},
	author = {Mohammadkhani, Ahmad Haji and Tantithamthavorn, Chakkrit and Hemmatif, Hadi},
	month = oct,
	year = {2023},
	pages = {96--106},
}

@article{wang_exploring_2021,
	title = {Exploring {Explainable} {Selection} to {Control} {Abstractive} {Summarization}},
	volume = {35},
	issn = {2374-3468, 2159-5399},
	url = {https://ojs.aaai.org/index.php/AAAI/article/view/17641},
	doi = {10.1609/aaai.v35i15.17641},
	language = {en},
	number = {15},
	urldate = {2025-01-29},
	journal = {Proceedings of the AAAI Conference on Artificial Intelligence},
	author = {Wang, Haonan and Gao, Yang and Bai, Yu and Lapata, Mirella and Huang, Heyan},
	month = may,
	year = {2021},
	pages = {13933--13941},
}

@article{chen_faster_2023,
	title = {Faster, {Stronger}, and {More} {Interpretable}: {Massive} {Transformer} {Architectures} for {Vision}-{Language} {Tasks}},
	volume = {03},
	issn = {25829793},
	shorttitle = {Faster, {Stronger}, and {More} {Interpretable}},
	url = {https://www.oajaiml.com/uploads/archivepdf/50081181.pdf},
	doi = {10.54364/AAIML.2023.1181},
	language = {en},
	number = {03},
	urldate = {2025-01-29},
	journal = {Advances in Artificial Intelligence and Machine Learning},
	author = {Chen, Tong and Liu, Sicong and Chen, Zhiran and Hu, Wenyan and Chen, Dachi and Wang, Yuanxin and Lyu, Qi and Le, Cindy X. and Wang, Wenping},
	year = {2023},
	pages = {1369--1388},
}

@inproceedings{huang_generic_2023,
	address = {Tainan Taiwan},
	title = {Generic {Attention}-model {Explainability} by {Weighted} {Relevance} {Accumulation}},
	isbn = {9798400702051},
	url = {https://dl.acm.org/doi/10.1145/3595916.3626437},
	doi = {10.1145/3595916.3626437},
	language = {en},
	urldate = {2025-01-29},
	booktitle = {{ACM} {Multimedia} {Asia} 2023},
	publisher = {ACM},
	author = {Huang, Yiming and Jia, Aozhe and Zhang, Xiaodan and Zhang, Jiawei},
	month = dec,
	year = {2023},
	pages = {1--7},
}

@inproceedings{chefer_generic_2021,
	address = {Montreal, QC, Canada},
	title = {Generic {Attention}-model {Explainability} for {Interpreting} {Bi}-{Modal} and {Encoder}-{Decoder} {Transformers}},
	copyright = {https://doi.org/10.15223/policy-029},
	isbn = {978-1-66542-812-5},
	url = {https://ieeexplore.ieee.org/document/9710570/},
	doi = {10.1109/ICCV48922.2021.00045},
	language = {en},
	urldate = {2025-01-29},
	booktitle = {2021 {IEEE}/{CVF} {International} {Conference} on {Computer} {Vision} ({ICCV})},
	publisher = {IEEE},
	author = {Chefer, Hila and Gur, Shir and Wolf, Lior},
	month = oct,
	year = {2021},
	pages = {387--396},
}

@inproceedings{sarti_inseq_2023,
	title = {Inseq: {An} {Interpretability} {Toolkit} for {Sequence} {Generation} {Models}},
	shorttitle = {Inseq},
	url = {http://arxiv.org/abs/2302.13942},
	doi = {10.18653/v1/2023.acl-demo.40},
	language = {en},
	urldate = {2025-01-29},
	booktitle = {Proceedings of the 61st {Annual} {Meeting} of the {Association} for {Computational} {Linguistics} ({Volume} 3: {System} {Demonstrations})},
	author = {Sarti, Gabriele and Feldhus, Nils and Sickert, Ludwig and Wal, Oskar van der and Nissim, Malvina and Bisazza, Arianna},
	year = {2023},
	note = {arXiv:2302.13942 [cs]},
	pages = {421--435},
}

@article{wu_interpretable_2023,
	title = {Interpretable tourism demand forecasting with temporal fusion transformers amid {COVID}-19},
	volume = {53},
	issn = {0924-669X, 1573-7497},
	url = {https://link.springer.com/10.1007/s10489-022-04254-0},
	doi = {10.1007/s10489-022-04254-0},
	language = {en},
	number = {11},
	urldate = {2025-01-29},
	journal = {Applied Intelligence},
	author = {Wu, Binrong and Wang, Lin and Zeng, Yu-Rong},
	month = jun,
	year = {2023},
	pages = {14493--14514},
}

@misc{parelli_interpretable_2023,
	title = {Interpretable {Visual} {Question} {Answering} via {Reasoning} {Supervision}},
	url = {http://arxiv.org/abs/2309.03726},
	doi = {10.48550/arXiv.2309.03726},
	language = {en},
	urldate = {2025-01-29},
	publisher = {arXiv},
	author = {Parelli, Maria and Mallis, Dimitrios and Diomataris, Markos and Pitsikalis, Vassilis},
	month = sep,
	year = {2023},
	note = {arXiv:2309.03726 [cs]},
}

@misc{malkiel_interpreting_2022,
	title = {Interpreting {BERT}-based {Text} {Similarity} via {Activation} and {Saliency} {Maps}},
	url = {http://arxiv.org/abs/2208.06612},
	doi = {10.48550/arXiv.2208.06612},
	language = {en},
	urldate = {2025-01-29},
	publisher = {arXiv},
	author = {Malkiel, Itzik and Ginzburg, Dvir and Barkan, Oren and Caciularu, Avi and Weill, Jonathan and Koenigstein, Noam},
	month = aug,
	year = {2022},
	note = {arXiv:2208.06612 [cs]},
}

@article{boito_investigating_2020,
	title = {Investigating alignment interpretability for low-resource {NMT}},
	volume = {34},
	issn = {0922-6567, 1573-0573},
	url = {http://link.springer.com/10.1007/s10590-020-09254-w},
	doi = {10.1007/s10590-020-09254-w},
	language = {en},
	number = {4},
	urldate = {2025-01-29},
	journal = {Machine Translation},
	author = {Boito, Marcely Zanon and Villavicencio, Aline and Besacier, Laurent},
	month = dec,
	year = {2020},
	pages = {305--323},
}

@inproceedings{treviso_ist-unbabel_2021,
	address = {Punta Cana, Dominican Republic},
	title = {{IST}-{Unbabel} 2021 {Submission} for the {Explainable} {Quality} {Estimation} {Shared} {Task}},
	url = {https://aclanthology.org/2021.eval4nlp-1.14},
	doi = {10.18653/v1/2021.eval4nlp-1.14},
	language = {en},
	urldate = {2025-01-29},
	booktitle = {Proceedings of the 2nd {Workshop} on {Evaluation} and {Comparison} of {NLP} {Systems}},
	publisher = {Association for Computational Linguistics},
	author = {Treviso, Marcos and Guerreiro, Nuno M. and Rei, Ricardo and Martins, André F. T.},
	year = {2021},
	pages = {133--145},
}

@inproceedings{zanzotto_kermit_2020,
	address = {Online},
	title = {{KERMIT}: {Complementing} {Transformer} {Architectures} with {Encoders} of {Explicit} {Syntactic} {Interpretations}},
	shorttitle = {{KERMIT}},
	url = {https://www.aclweb.org/anthology/2020.emnlp-main.18},
	doi = {10.18653/v1/2020.emnlp-main.18},
	language = {en},
	urldate = {2025-01-29},
	booktitle = {Proceedings of the 2020 {Conference} on {Empirical} {Methods} in {Natural} {Language} {Processing} ({EMNLP})},
	publisher = {Association for Computational Linguistics},
	author = {Zanzotto, Fabio Massimo and Santilli, Andrea and Ranaldi, Leonardo and Onorati, Dario and Tommasino, Pierfrancesco and Fallucchi, Francesca},
	year = {2020},
	pages = {256--267},
}

@inproceedings{xu_logiformer_2022,
	address = {Madrid Spain},
	title = {Logiformer: {A} {Two}-{Branch} {Graph} {Transformer} {Network} for {Interpretable} {Logical} {Reasoning}},
	isbn = {978-1-4503-8732-3},
	shorttitle = {Logiformer},
	url = {https://dl.acm.org/doi/10.1145/3477495.3532016},
	doi = {10.1145/3477495.3532016},
	language = {en},
	urldate = {2025-01-29},
	booktitle = {Proceedings of the 45th {International} {ACM} {SIGIR} {Conference} on {Research} and {Development} in {Information} {Retrieval}},
	publisher = {ACM},
	author = {Xu, Fangzhi and Liu, Jun and Lin, Qika and Pan, Yudai and Zhang, Lingling},
	month = jul,
	year = {2022},
	pages = {1055--1065},
}

@inproceedings{xu_multi-granular_2020,
	address = {Sydney, NSW, Australia},
	title = {Multi-{Granular} {BERT}: {An} {Interpretable} {Model} {Applicable} to {Internet}-of-{Thing} devices},
	copyright = {https://ieeexplore.ieee.org/Xplorehelp/downloads/license-information/IEEE.html},
	isbn = {978-0-7381-0500-0},
	shorttitle = {Multi-{Granular} {BERT}},
	url = {https://ieeexplore.ieee.org/document/9270262/},
	doi = {10.1109/ICEI49372.2020.00032},
	language = {en},
	urldate = {2025-01-29},
	booktitle = {2020 {IEEE} {International} {Conference} on {Energy} {Internet} ({ICEI})},
	publisher = {IEEE},
	author = {Xu, Sihao and Zhang, Wei and Zhang, Fan},
	month = aug,
	year = {2020},
	pages = {134--139},
}

@article{che_multiscale_2023,
	title = {Multiscale {Time}-{Frequency} {Sparse} {Transformer} {Based} on {Partly} {Interpretable} {Method} for {Bearing} {Fault} {Diagnosis}},
	volume = {2023},
	copyright = {https://creativecommons.org/licenses/by/4.0/},
	issn = {1875-9203, 1070-9622},
	url = {https://www.hindawi.com/journals/sv/2023/1639287/},
	doi = {10.1155/2023/1639287},
	language = {en},
	urldate = {2025-01-29},
	journal = {Shock and Vibration},
	author = {Che, Shouquan and Lu, Jianfeng and Bao, Congwang and Zhang, Caihong and Liu, Yongzhi},
	editor = {Jiang, Xingxing},
	month = aug,
	year = {2023},
	pages = {1--21},
}

@article{heo_natural-language-driven_2023,
	title = {Natural-{Language}-{Driven} {Multimodal} {Representation} {Learning} for {Audio}-{Visual} {Scene}-{Aware} {Dialog} {System}},
	volume = {23},
	copyright = {https://creativecommons.org/licenses/by/4.0/},
	issn = {1424-8220},
	url = {https://www.mdpi.com/1424-8220/23/18/7875},
	doi = {10.3390/s23187875},
	language = {en},
	number = {18},
	urldate = {2025-01-29},
	journal = {Sensors},
	author = {Heo, Yoonseok and Kang, Sangwoo and Seo, Jungyun},
	month = sep,
	year = {2023},
	pages = {7875},
}

@article{sun_neural_2021,
	title = {Neural {Encoding} and {Decoding} {With} {Distributed} {Sentence} {Representations}},
	volume = {32},
	copyright = {https://ieeexplore.ieee.org/Xplorehelp/downloads/license-information/IEEE.html},
	issn = {2162-237X, 2162-2388},
	url = {https://ieeexplore.ieee.org/document/9223750/},
	doi = {10.1109/TNNLS.2020.3027595},
	language = {en},
	number = {2},
	urldate = {2025-01-29},
	journal = {IEEE Transactions on Neural Networks and Learning Systems},
	author = {Sun, Jingyuan and Wang, Shaonan and Zhang, Jiajun and Zong, Chengqing},
	month = feb,
	year = {2021},
	pages = {589--603},
}

@article{wang_odp-transformer_2023,
	title = {{ODP}-{Transformer}: {Interpretation} of pest classification results using image caption generation techniques},
	volume = {209},
	issn = {01681699},
	shorttitle = {{ODP}-{Transformer}},
	url = {https://linkinghub.elsevier.com/retrieve/pii/S016816992300251X},
	doi = {10.1016/j.compag.2023.107863},
	language = {en},
	urldate = {2025-01-29},
	journal = {Computers and Electronics in Agriculture},
	author = {Wang, Shansong and Zeng, Qingtian and Ni, Weijian and Cheng, Cheng and Wang, Yanxue},
	month = jun,
	year = {2023},
	pages = {107863},
}

@article{ukwuoma_hybrid_2023,
	title = {A hybrid explainable ensemble transformer encoder for pneumonia identification from chest {X}-ray images},
	volume = {48},
	issn = {20901232},
	url = {https://linkinghub.elsevier.com/retrieve/pii/S2090123222002028},
	doi = {10.1016/j.jare.2022.08.021},
	language = {en},
	urldate = {2025-01-29},
	journal = {Journal of Advanced Research},
	author = {Ukwuoma, Chiagoziem C. and Qin, Zhiguang and Belal Bin Heyat, Md and Akhtar, Faijan and Bamisile, Olusola and Muaad, Abdullah Y. and Addo, Daniel and Al-antari, Mugahed A.},
	month = jun,
	year = {2023},
	pages = {191--211},
}

@misc{li_oscar_2020,
	title = {Oscar: {Object}-{Semantics} {Aligned} {Pre}-training for {Vision}-{Language} {Tasks}},
	shorttitle = {Oscar},
	url = {http://arxiv.org/abs/2004.06165},
	doi = {10.48550/arXiv.2004.06165},
	language = {en},
	urldate = {2025-01-29},
	publisher = {arXiv},
	author = {Li, Xiujun and Yin, Xi and Li, Chunyuan and Zhang, Pengchuan and Hu, Xiaowei and Zhang, Lei and Wang, Lijuan and Hu, Houdong and Dong, Li and Wei, Furu and Choi, Yejin and Gao, Jianfeng},
	month = jul,
	year = {2020},
	note = {arXiv:2004.06165 [cs]},
}

@article{kandukuri_physical_2022,
	title = {Physical {Representation} {Learning} and {Parameter} {Identification} from {Video} {Using} {Differentiable} {Physics}},
	volume = {130},
	issn = {0920-5691, 1573-1405},
	url = {https://link.springer.com/10.1007/s11263-021-01493-5},
	doi = {10.1007/s11263-021-01493-5},
	language = {en},
	number = {1},
	urldate = {2025-01-29},
	journal = {International Journal of Computer Vision},
	author = {Kandukuri, Rama Krishna and Achterhold, Jan and Moeller, Michael and Stueckler, Joerg},
	month = jan,
	year = {2022},
	pages = {3--16},
}

@article{huang_representation_2023,
	title = {Representation of time-varying and time-invariant {EMR} data and its application in modeling outcome prediction for heart failure patients},
	volume = {143},
	issn = {15320464},
	url = {https://linkinghub.elsevier.com/retrieve/pii/S153204642300148X},
	doi = {10.1016/j.jbi.2023.104427},
	language = {en},
	urldate = {2025-01-29},
	journal = {Journal of Biomedical Informatics},
	author = {Huang, Yanqun and Wang, Muyu and Zheng, Zhimin and Ma, Moxuan and Fei, Xiaolu and Wei, Lan and Chen, Hui},
	month = jul,
	year = {2023},
	pages = {104427},
}

@article{jha_supervised_2023,
	title = {Supervised {Contrastive} {Learning} for {Interpretable} {Long}-{Form} {Document} {Matching}},
	volume = {17},
	issn = {1556-4681, 1556-472X},
	url = {https://dl.acm.org/doi/10.1145/3542822},
	doi = {10.1145/3542822},
	language = {en},
	number = {2},
	urldate = {2025-01-29},
	journal = {ACM Transactions on Knowledge Discovery from Data},
	author = {Jha, Akshita and Rakesh, Vineeth and Chandrashekar, Jaideep and Samavedhi, Adithya and Reddy, Chandan K.},
	month = apr,
	year = {2023},
	pages = {1--17},
}

@article{wang_tfregnci_2023,
	title = {{TFRegNCI}: {Interpretable} {Noncovalent} {Interaction} {Correction} {Multimodal} {Based} on {Transformer} {Encoder} {Fusion}},
	volume = {63},
	copyright = {https://doi.org/10.15223/policy-029},
	issn = {1549-9596, 1549-960X},
	shorttitle = {{TFRegNCI}},
	url = {https://pubs.acs.org/doi/10.1021/acs.jcim.2c01283},
	doi = {10.1021/acs.jcim.2c01283},
	language = {en},
	number = {3},
	urldate = {2025-01-29},
	journal = {Journal of Chemical Information and Modeling},
	author = {Wang, Donghan and Li, Wenze and Dong, Xu and Li, Hongzhi and Hu, LiHong},
	month = feb,
	year = {2023},
	pages = {782--793},
}

@misc{ferrando_towards_2022,
	title = {Towards {Opening} the {Black} {Box} of {Neural} {Machine} {Translation}: {Source} and {Target} {Interpretations} of the {Transformer}},
	shorttitle = {Towards {Opening} the {Black} {Box} of {Neural} {Machine} {Translation}},
	url = {http://arxiv.org/abs/2205.11631},
	doi = {10.48550/arXiv.2205.11631},
	language = {en},
	urldate = {2025-01-29},
	publisher = {arXiv},
	author = {Ferrando, Javier and Gállego, Gerard I. and Alastruey, Belen and Escolano, Carlos and Costa-jussà, Marta R.},
	month = nov,
	year = {2022},
	note = {arXiv:2205.11631 [cs]},
}

@inproceedings{kumar_towards_2021,
	title = {Towards the {Explainability} of {Multimodal} {Speech} {Emotion} {Recognition}},
	url = {https://www.isca-archive.org/interspeech_2021/kumar21d_interspeech.html},
	doi = {10.21437/Interspeech.2021-1718},
	language = {en},
	urldate = {2025-01-29},
	booktitle = {Interspeech 2021},
	publisher = {ISCA},
	author = {Kumar, Puneet and Kaushik, Vishesh and Raman, Balasubramanian},
	month = aug,
	year = {2021},
	pages = {1748--1752},
}

@article{xiao_transformer_2024,
	title = {Transformer with convolution and graph-node co-embedding: {An} accurate and interpretable vision backbone for predicting gene expressions from local histopathological image},
	volume = {91},
	issn = {13618415},
	shorttitle = {Transformer with convolution and graph-node co-embedding},
	url = {https://linkinghub.elsevier.com/retrieve/pii/S1361841523003006},
	doi = {10.1016/j.media.2023.103040},
	language = {en},
	urldate = {2025-01-29},
	journal = {Medical Image Analysis},
	author = {Xiao, Xiao and Kong, Yan and Li, Ronghan and Wang, Zuoheng and Lu, Hui},
	month = jan,
	year = {2024},
	pages = {103040},
}

@incollection{hiemstra_using_2021,
	address = {Cham},
	title = {Using the {Hammer} only on {Nails}: {A} {Hybrid} {Method} for {Representation}-{Based} {Evidence} {Retrieval} for {Question} {Answering}},
	volume = {12656},
	isbn = {978-3-030-72112-1 978-3-030-72113-8},
	shorttitle = {Using the {Hammer} only on {Nails}},
	url = {https://link.springer.com/10.1007/978-3-030-72113-8_22},
	language = {en},
	urldate = {2025-01-29},
	booktitle = {Advances in {Information} {Retrieval}},
	publisher = {Springer International Publishing},
	author = {Liang, Zhengzhong and Zhao, Yiyun and Surdeanu, Mihai},
	editor = {Hiemstra, Djoerd and Moens, Marie-Francine and Mothe, Josiane and Perego, Raffaele and Potthast, Martin and Sebastiani, Fabrizio},
	year = {2021},
	doi = {10.1007/978-3-030-72113-8_22},
	note = {Series Title: Lecture Notes in Computer Science},
	pages = {327--341},
}

@article{naseem_vision-language_2023,
	title = {Vision-{Language} {Transformer} for {Interpretable} {Pathology} {Visual} {Question} {Answering}},
	volume = {27},
	copyright = {https://ieeexplore.ieee.org/Xplorehelp/downloads/license-information/IEEE.html},
	issn = {2168-2194, 2168-2208},
	url = {https://ieeexplore.ieee.org/document/9745795/},
	doi = {10.1109/JBHI.2022.3163751},
	language = {en},
	number = {4},
	urldate = {2025-01-29},
	journal = {IEEE Journal of Biomedical and Health Informatics},
	author = {Naseem, Usman and Khushi, Matloob and Kim, Jinman},
	month = apr,
	year = {2023},
	pages = {1681--1690},
}

@misc{katz_visit_2023,
	title = {{VISIT}: {Visualizing} and {Interpreting} the {Semantic} {Information} {Flow} of {Transformers}},
	shorttitle = {{VISIT}},
	url = {http://arxiv.org/abs/2305.13417},
	doi = {10.48550/arXiv.2305.13417},
	language = {en},
	urldate = {2025-01-29},
	publisher = {arXiv},
	author = {Katz, Shahar and Belinkov, Yonatan},
	month = nov,
	year = {2023},
	note = {arXiv:2305.13417 [cs]},
}

@misc{aflalo_vl-interpret_2022,
	title = {{VL}-{InterpreT}: {An} {Interactive} {Visualization} {Tool} for {Interpreting} {Vision}-{Language} {Transformers}},
	shorttitle = {{VL}-{InterpreT}},
	url = {http://arxiv.org/abs/2203.17247},
	doi = {10.48550/arXiv.2203.17247},
	language = {en},
	urldate = {2025-01-29},
	publisher = {arXiv},
	author = {Aflalo, Estelle and Du, Meng and Tseng, Shao-Yen and Liu, Yongfei and Wu, Chenfei and Duan, Nan and Lal, Vasudev},
	month = aug,
	year = {2022},
	note = {arXiv:2203.17247 [cs]},
}

@article{ilinykh_what_2021,
	title = {What {Does} a {Language}-{And}-{Vision} {Transformer} {See}: {The} {Impact} of {Semantic} {Information} on {Visual} {Representations}},
	volume = {4},
	issn = {2624-8212},
	shorttitle = {What {Does} a {Language}-{And}-{Vision} {Transformer} {See}},
	url = {https://www.frontiersin.org/articles/10.3389/frai.2021.767971/full},
	doi = {10.3389/frai.2021.767971},
	language = {en},
	urldate = {2025-01-29},
	journal = {Frontiers in Artificial Intelligence},
	author = {Ilinykh, Nikolai and Dobnik, Simon},
	month = dec,
	year = {2021},
	pages = {767971},
}

@article{dong_why_2023,
	title = {Why did the {AI} make that decision? {Towards} an explainable artificial intelligence ({XAI}) for autonomous driving systems},
	volume = {156},
	issn = {0968090X},
	shorttitle = {Why did the {AI} make that decision?},
	url = {https://linkinghub.elsevier.com/retrieve/pii/S0968090X23003480},
	doi = {10.1016/j.trc.2023.104358},
	language = {en},
	urldate = {2025-01-29},
	journal = {Transportation Research Part C: Emerging Technologies},
	author = {Dong, Jiqian and Chen, Sikai and Miralinaghi, Mohammad and Chen, Tiantian and Li, Pei and Labi, Samuel},
	month = nov,
	year = {2023},
	pages = {104358},
}

@misc{lin_zero-shot_2023,
	title = {Zero-{Shot} {Everything} {Sketch}-{Based} {Image} {Retrieval}, and in {Explainable} {Style}},
	url = {http://arxiv.org/abs/2303.14348},
	doi = {10.48550/arXiv.2303.14348},
	language = {en},
	urldate = {2025-01-29},
	publisher = {arXiv},
	author = {Lin, Fengyin and Li, Mingkang and Li, Da and Hospedales, Timothy and Song, Yi-Zhe and Qi, Yonggang},
	month = mar,
	year = {2023},
	note = {arXiv:2303.14348 [cs]},
}

@article{fantozzi2024explainability,
  title={The Explainability of Transformers: Current Status and Directions},
  author={Fantozzi, Paolo and Naldi, Maurizio},
  journal={Computers},
  volume={13},
  number={4},
  pages={92},
  year={2024},
  publisher={MDPI}
}

@article{chun2024comparative,
  title={Comparative Global AI Regulation: Policy Perspectives from the EU, China, and the US},
  author={Chun, Jon and de Witt, Christian Schroeder and Elkins, Katherine},
  journal={arXiv preprint arXiv:2410.21279},
  year={2024}
}

@article{burkart2021survey,
  title={A survey on the explainability of supervised machine learning},
  author={Burkart, Nadia and Huber, Marco F},
  journal={Journal of Artificial Intelligence Research},
  volume={70},
  pages={245--317},
  year={2021}
}

@article{jaitly2016online,
  title={An online sequence-to-sequence model using partial conditioning},
  author={Jaitly, Navdeep and Le, Quoc V and Vinyals, Oriol and Sutskever, Ilya and Sussillo, David and Bengio, Samy},
  journal={Advances in neural information processing systems},
  volume={29},
  year={2016}
}

@inproceedings{prabhavalkar2017analysis,
  title={An Analysis of" Attention" in Sequence-to-Sequence Models.},
  author={Prabhavalkar, Rohit and Sainath, Tara N and Li, Bo and Rao, Kanishka and Jaitly, Navdeep},
  booktitle={Interspeech},
  pages={3702--3706},
  year={2017}
}

@article{tan2019lxmert,
  title={Lxmert: Learning cross-modality encoder representations from transformers},
  author={Tan, Hao and Bansal, Mohit},
  journal={arXiv preprint arXiv:1908.07490},
  year={2019}
}

@article{vilone2021classification,
  title={Classification of explainable artificial intelligence methods through their output formats},
  author={Vilone, Giulia and Longo, Luca},
  journal={Machine Learning and Knowledge Extraction},
  volume={3},
  number={3},
  pages={615--661},
  year={2021},
  publisher={MDPI}
}

@inproceedings{xu2019explainable,
  title={Explainable AI: A brief survey on history, research areas, approaches and challenges},
  author={Xu, Feiyu and Uszkoreit, Hans and Du, Yangzhou and Fan, Wei and Zhao, Dongyan and Zhu, Jun},
  booktitle={Natural language processing and Chinese computing: 8th cCF international conference, NLPCC 2019, dunhuang, China, October 9--14, 2019, proceedings, part II 8},
  pages={563--574},
  year={2019},
  organization={Springer}
}

@article{jain2019attention,
  title={Attention is not explanation},
  author={Jain, Sarthak and Wallace, Byron C},
  journal={arXiv preprint arXiv:1902.10186},
  year={2019}
}

@inproceedings{vig-2019-multiscale,
    title = "A Multiscale Visualization of Attention in the Transformer Model",
    author = "Vig, Jesse",
    booktitle = "Proceedings of the 57th Annual Meeting of the Association for Computational Linguistics: System Demonstrations",
    month = jul,
    year = "2019",
    address = "Florence, Italy",
    publisher = "Association for Computational Linguistics",
    url = "https://www.aclweb.org/anthology/P19-3007",
    doi = "10.18653/v1/P19-3007",
    pages = "37--42",
}

@inproceedings{selvaraju2017grad,
  title={Grad-cam: Visual explanations from deep networks via gradient-based localization},
  author={Selvaraju, Ramprasaath R and Cogswell, Michael and Das, Abhishek and Vedantam, Ramakrishna and Parikh, Devi and Batra, Dhruv},
  booktitle={Proceedings of the IEEE international conference on computer vision},
  pages={618--626},
  year={2017}
}

@article{hartmann2022survey,
  title={A survey on improving NLP models with human explanations},
  author={Hartmann, Mareike and Sonntag, Daniel},
  journal={arXiv preprint arXiv:2204.08892},
  year={2022}
}

@article{radford2019language,
  title={Language models are unsupervised multitask learners},
  author={Radford, Alec and Wu, Jeffrey and Child, Rewon and Luan, David and Amodei, Dario and Sutskever, Ilya and others},
  journal={OpenAI blog},
  volume={1},
  number={8},
  pages={9},
  year={2019}
}

@inproceedings{voorhees1999trec,
  title={The trec-8 question answering track report.},
  author={Voorhees, Ellen M and others},
  booktitle={Trec},
  volume={99},
  pages={77--82},
  year={1999}
}

@article{doshi2017towards,
  title={Towards a rigorous science of interpretable machine learning},
  author={Doshi-Velez, Finale and Kim, Been},
  journal={arXiv preprint arXiv:1702.08608},
  year={2017}
}

@article{correia2019adaptively,
  title={Adaptively sparse transformers},
  author={Correia, Gon{\c{c}}alo M and Niculae, Vlad and Martins, Andr{\'e} FT},
  journal={arXiv preprint arXiv:1909.00015},
  year={2019}
}

@article{fan2019reducing,
  title={Reducing transformer depth on demand with structured dropout},
  author={Fan, Angela and Grave, Edouard and Joulin, Armand},
  journal={arXiv preprint arXiv:1909.11556},
  year={2019}
}

@article{mohseni2021multidisciplinary,
  title={A multidisciplinary survey and framework for design and evaluation of explainable AI systems},
  author={Mohseni, Sina and Zarei, Niloofar and Ragan, Eric D},
  journal={ACM Transactions on Interactive Intelligent Systems (TiiS)},
  volume={11},
  number={3-4},
  pages={1--45},
  year={2021},
  publisher={ACM New York, NY}
}

@article{vilone2021notions,
  title={Notions of explainability and evaluation approaches for explainable artificial intelligence},
  author={Vilone, Giulia and Longo, Luca},
  journal={Information Fusion},
  volume={76},
  pages={89--106},
  year={2021},
  publisher={Elsevier}
}

@article{yoshino2023overview,
  title={Overview of the Tenth Dialog System Technology Challenge: DSTC10},
  author={Yoshino, Koichiro and Chen, Yun-Nung and Crook, Paul and Kottur, Satwik and Li, Jinchao and Hedayatnia, Behnam and Moon, Seungwhan and Fei, Zhengcong and Li, Zekang and Zhang, Jinchao and others},
  journal={IEEE/ACM Transactions on Audio, Speech, and Language Processing},
  volume={32},
  pages={765--778},
  year={2023},
  publisher={IEEE}
}

@article{carvalho2019machine,
  title={Machine learning interpretability: A survey on methods and metrics},
  author={Carvalho, Diogo V and Pereira, Eduardo M and Cardoso, Jaime S},
  journal={Electronics},
  volume={8},
  number={8},
  pages={832},
  year={2019},
  publisher={Multidisciplinary Digital Publishing Institute}
}

@article{ferrando2022measuring,
  title={Measuring the mixing of contextual information in the transformer},
  author={Ferrando, Javier and G{\'a}llego, Gerard I and Costa-Juss{\`a}, Marta R},
  journal={arXiv preprint arXiv:2203.04212},
  year={2022}
}

@article{liu2021kd,
  title={Kd-vlp: Improving end-to-end vision-and-language pretraining with object knowledge distillation},
  author={Liu, Yongfei and Wu, Chenfei and Tseng, Shao-yen and Lal, Vasudev and He, Xuming and Duan, Nan},
  journal={arXiv preprint arXiv:2109.10504},
  year={2021}
}

@article{wadekar2024evolution,
  title={The evolution of multimodal model architectures},
  author={Wadekar, Shakti N and Chaurasia, Abhishek and Chadha, Aman and Culurciello, Eugenio},
  journal={arXiv preprint arXiv:2405.17927},
  year={2024}
}

@article{yin2024survey,
  title={A survey on multimodal large language models},
  author={Yin, Shukang and Fu, Chaoyou and Zhao, Sirui and Li, Ke and Sun, Xing and Xu, Tong and Chen, Enhong},
  journal={National Science Review},
  volume={11},
  number={12},
  pages={nwae403},
  year={2024},
  publisher={Oxford University Press}
}

@inproceedings{wu2023multimodal,
  title={Multimodal large language models: A survey},
  author={Wu, Jiayang and Gan, Wensheng and Chen, Zefeng and Wan, Shicheng and Yu, Philip S},
  booktitle={2023 IEEE International Conference on Big Data (BigData)},
  pages={2247--2256},
  year={2023},
  organization={IEEE}
}

@article{jin2024efficient,
  title={Efficient multimodal large language models: A survey},
  author={Jin, Yizhang and Li, Jian and Liu, Yexin and Gu, Tianjun and Wu, Kai and Jiang, Zhengkai and He, Muyang and Zhao, Bo and Tan, Xin and Gan, Zhenye and others},
  journal={arXiv preprint arXiv:2405.10739},
  year={2024}
}

@article{liu2023visual,
  title={Visual instruction tuning},
  author={Liu, Haotian and Li, Chunyuan and Wu, Qingyang and Lee, Yong Jae},
  journal={Advances in neural information processing systems},
  volume={36},
  pages={34892--34916},
  year={2023}
}

@article{lu2024deepseek,
  title={Deepseek-vl: towards real-world vision-language understanding},
  author={Lu, Haoyu and Liu, Wen and Zhang, Bo and Wang, Bingxuan and Dong, Kai and Liu, Bo and Sun, Jingxiang and Ren, Tongzheng and Li, Zhuoshu and Yang, Hao and others},
  journal={arXiv preprint arXiv:2403.05525},
  year={2024}
}

@inproceedings{li2023blip,
  title={Blip-2: Bootstrapping language-image pre-training with frozen image encoders and large language models},
  author={Li, Junnan and Li, Dongxu and Savarese, Silvio and Hoi, Steven},
  booktitle={International conference on machine learning},
  pages={19730--19742},
  year={2023},
  organization={PMLR}
}

@article{zhang2023video,
  title={Video-llama: An instruction-tuned audio-visual language model for video understanding},
  author={Zhang, Hang and Li, Xin and Bing, Lidong},
  journal={arXiv preprint arXiv:2306.02858},
  year={2023}
}

@article{bai2023qwen,
  title={Qwen-vl: A frontier large vision-language model with versatile abilities},
  author={Bai, Jinze and Bai, Shuai and Yang, Shusheng and Wang, Shijie and Tan, Sinan and Wang, Peng and Lin, Junyang and Zhou, Chang and Zhou, Jingren},
  journal={arXiv preprint arXiv:2308.12966},
  volume={1},
  number={2},
  pages={3},
  year={2023}
}

@article{jin2023unified,
  title={Unified language-vision pretraining in llm with dynamic discrete visual tokenization},
  author={Jin, Yang and Xu, Kun and Chen, Liwei and Liao, Chao and Tan, Jianchao and Huang, Quzhe and Chen, Bin and Lei, Chenyi and Liu, An and Song, Chengru and others},
  journal={arXiv preprint arXiv:2309.04669},
  year={2023}
}

@article{zhu2023vl,
  title={Vl-gpt: A generative pre-trained transformer for vision and language understanding and generation},
  author={Zhu, Jinguo and Ding, Xiaohan and Ge, Yixiao and Ge, Yuying and Zhao, Sijie and Zhao, Hengshuang and Wang, Xiaohua and Shan, Ying},
  journal={arXiv preprint arXiv:2312.09251},
  year={2023}
}

@inproceedings{cho2021unifying,
  title={Unifying vision-and-language tasks via text generation},
  author={Cho, Jaemin and Lei, Jie and Tan, Hao and Bansal, Mohit},
  booktitle={International Conference on Machine Learning},
  pages={1931--1942},
  year={2021},
  organization={PMLR}
}

@inproceedings{sung2022vl,
  title={Vl-adapter: Parameter-efficient transfer learning for vision-and-language tasks},
  author={Sung, Yi-Lin and Cho, Jaemin and Bansal, Mohit},
  booktitle={Proceedings of the IEEE/CVF conference on computer vision and pattern recognition},
  pages={5227--5237},
  year={2022}
}

@article{chen2023pali,
  title={Pali-x: On scaling up a multilingual vision and language model},
  author={Chen, Xi and Djolonga, Josip and Padlewski, Piotr and Mustafa, Basil and Changpinyo, Soravit and Wu, Jialin and Ruiz, Carlos Riquelme and Goodman, Sebastian and Wang, Xiao and Tay, Yi and others},
  journal={arXiv preprint arXiv:2305.18565},
  year={2023}
}

@article{alayrac2022flamingo,
  title={Flamingo: a visual language model for few-shot learning},
  author={Alayrac, Jean-Baptiste and Donahue, Jeff and Luc, Pauline and Miech, Antoine and Barr, Iain and Hasson, Yana and Lenc, Karel and Mensch, Arthur and Millican, Katherine and Reynolds, Malcolm and others},
  journal={Advances in neural information processing systems},
  volume={35},
  pages={23716--23736},
  year={2022}
}

@article{gao2023llama,
  title={Llama-adapter v2: Parameter-efficient visual instruction model},
  author={Gao, Peng and Han, Jiaming and Zhang, Renrui and Lin, Ziyi and Geng, Shijie and Zhou, Aojun and Zhang, Wei and Lu, Pan and He, Conghui and Yue, Xiangyu and others},
  journal={arXiv preprint arXiv:2304.15010},
  year={2023}
}

@article{shen2024mome,
  title={Mome: Mixture of multimodal experts for generalist multimodal large language models},
  author={Shen, Leyang and Chen, Gongwei and Shao, Rui and Guan, Weili and Nie, Liqiang},
  journal={Advances in neural information processing systems},
  volume={37},
  pages={42048--42070},
  year={2024}
}

@article{cai2025survey,
  title={A survey on mixture of experts in large language models},
  author={Cai, Weilin and Jiang, Juyong and Wang, Fan and Tang, Jing and Kim, Sunghun and Huang, Jiayi},
  journal={IEEE Transactions on Knowledge and Data Engineering},
  year={2025},
  publisher={IEEE}
}

@article{jordan1994hierarchical,
  title={Hierarchical mixtures of experts and the EM algorithm},
  author={Jordan, Michael I and Jacobs, Robert A},
  journal={Neural computation},
  volume={6},
  number={2},
  pages={181--214},
  year={1994},
  publisher={MIT Press}
}

@article{jacobs1991adaptive,
  title={Adaptive mixtures of local experts},
  author={Jacobs, Robert A and Jordan, Michael I and Nowlan, Steven J and Hinton, Geoffrey E},
  journal={Neural computation},
  volume={3},
  number={1},
  pages={79--87},
  year={1991},
  publisher={MIT Press}
}

@article{mustafa2022multimodal,
  title={Multimodal contrastive learning with limoe: the language-image mixture of experts},
  author={Mustafa, Basil and Riquelme, Carlos and Puigcerver, Joan and Jenatton, Rodolphe and Houlsby, Neil},
  journal={Advances in Neural Information Processing Systems},
  volume={35},
  pages={9564--9576},
  year={2022}
}

@article{wu2024deepseek,
  title={Deepseek-vl2: Mixture-of-experts vision-language models for advanced multimodal understanding},
  author={Wu, Zhiyu and Chen, Xiaokang and Pan, Zizheng and Liu, Xingchao and Liu, Wen and Dai, Damai and Gao, Huazuo and Ma, Yiyang and Wu, Chengyue and Wang, Bingxuan and others},
  journal={arXiv preprint arXiv:2412.10302},
  year={2024}
}

@article{lepikhin2020gshard,
  title={Gshard: Scaling giant models with conditional computation and automatic sharding},
  author={Lepikhin, Dmitry and Lee, HyoukJoong and Xu, Yuanzhong and Chen, Dehao and Firat, Orhan and Huang, Yanping and Krikun, Maxim and Shazeer, Noam and Chen, Zhifeng},
  journal={arXiv preprint arXiv:2006.16668},
  year={2020}
}

@inproceedings{mckinzie2024mm1,
  title={Mm1: methods, analysis and insights from multimodal llm pre-training},
  author={McKinzie, Brandon and Gan, Zhe and Fauconnier, Jean-Philippe and Dodge, Sam and Zhang, Bowen and Dufter, Philipp and Shah, Dhruti and Du, Xianzhi and Peng, Futang and Belyi, Anton and others},
  booktitle={European Conference on Computer Vision},
  pages={304--323},
  year={2024},
  organization={Springer}
}

@article{lin2024moe,
  title={Moe-llava: Mixture of experts for large vision-language models},
  author={Lin, Bin and Tang, Zhenyu and Ye, Yang and Cui, Jiaxi and Zhu, Bin and Jin, Peng and Huang, Jinfa and Zhang, Junwu and Pang, Yatian and Ning, Munan and others},
  journal={arXiv preprint arXiv:2401.15947},
  year={2024}
}

@article{li2025uni,
  title={Uni-moe: Scaling unified multimodal llms with mixture of experts},
  author={Li, Yunxin and Jiang, Shenyuan and Hu, Baotian and Wang, Longyue and Zhong, Wanqi and Luo, Wenhan and Ma, Lin and Zhang, Min},
  journal={IEEE Transactions on Pattern Analysis and Machine Intelligence},
  year={2025},
  publisher={IEEE}
}

@article{wu2024mixture,
  title={Mixture of lora experts},
  author={Wu, Xun and Huang, Shaohan and Wei, Furu},
  journal={arXiv preprint arXiv:2404.13628},
  year={2024}
}

@article{ouyang2022training,
  title={Training language models to follow instructions with human feedback},
  author={Ouyang, Long and Wu, Jeffrey and Jiang, Xu and Almeida, Diogo and Wainwright, Carroll and Mishkin, Pamela and Zhang, Chong and Agarwal, Sandhini and Slama, Katarina and Ray, Alex and others},
  journal={Advances in neural information processing systems},
  volume={35},
  pages={27730--27744},
  year={2022}
}

@article{rafailov2023direct,
  title={Direct preference optimization: Your language model is secretly a reward model},
  author={Rafailov, Rafael and Sharma, Archit and Mitchell, Eric and Manning, Christopher D and Ermon, Stefano and Finn, Chelsea},
  journal={Advances in neural information processing systems},
  volume={36},
  pages={53728--53741},
  year={2023}
}

@article{kojima2022large,
  title={Large language models are zero-shot reasoners},
  author={Kojima, Takeshi and Gu, Shixiang Shane and Reid, Machel and Matsuo, Yutaka and Iwasawa, Yusuke},
  journal={Advances in neural information processing systems},
  volume={35},
  pages={22199--22213},
  year={2022}
}

@article{camburu2018snli,
  title={e-snli: Natural language inference with natural language explanations},
  author={Camburu, Oana-Maria and Rockt{\"a}schel, Tim and Lukasiewicz, Thomas and Blunsom, Phil},
  journal={Advances in Neural Information Processing Systems},
  volume={31},
  year={2018}
}

@article{zhang2023multimodal,
  title={Multimodal chain-of-thought reasoning in language models},
  author={Zhang, Zhuosheng and Zhang, Aston and Li, Mu and Zhao, Hai and Karypis, George and Smola, Alex},
  journal={arXiv preprint arXiv:2302.00923},
  year={2023}
}

@article{wang2025multimodal,
  title={Multimodal chain-of-thought reasoning: A comprehensive survey},
  author={Wang, Yaoting and Wu, Shengqiong and Zhang, Yuecheng and Yan, Shuicheng and Liu, Ziwei and Luo, Jiebo and Fei, Hao},
  journal={arXiv preprint arXiv:2503.12605},
  year={2025}
}

@article{chen2023see,
  title={See, think, confirm: Interactive prompting between vision and language models for knowledge-based visual reasoning},
  author={Chen, Zhenfang and Zhou, Qinhong and Shen, Yikang and Hong, Yining and Zhang, Hao and Gan, Chuang},
  journal={arXiv preprint arXiv:2301.05226},
  year={2023}
}

@article{zhou2025r1,
  title={R1-Zero's" Aha Moment" in Visual Reasoning on a 2B Non-SFT Model},
  author={Zhou, Hengguang and Li, Xirui and Wang, Ruochen and Cheng, Minhao and Zhou, Tianyi and Hsieh, Cho-Jui},
  journal={arXiv preprint arXiv:2503.05132},
  year={2025}
}

@article{meng2025mm,
  title={Mm-eureka: Exploring the frontiers of multimodal reasoning with rule-based reinforcement learning},
  author={Meng, Fanqing and Du, Lingxiao and Liu, Zongkai and Zhou, Zhixiang and Lu, Quanfeng and Fu, Daocheng and Han, Tiancheng and Shi, Botian and Wang, Wenhai and He, Junjun and others},
  journal={arXiv preprint arXiv:2503.07365},
  year={2025}
}

@article{shao2024visual,
  title={Visual cot: Advancing multi-modal language models with a comprehensive dataset and benchmark for chain-of-thought reasoning},
  author={Shao, Hao and Qian, Shengju and Xiao, Han and Song, Guanglu and Zong, Zhuofan and Wang, Letian and Liu, Yu and Li, Hongsheng},
  journal={Advances in Neural Information Processing Systems},
  volume={37},
  pages={8612--8642},
  year={2024}
}

@article{do2020snli,
  title={e-snli-ve: Corrected visual-textual entailment with natural language explanations},
  author={Do, Virginie and Camburu, Oana-Maria and Akata, Zeynep and Lukasiewicz, Thomas},
  journal={arXiv preprint arXiv:2004.03744},
  year={2020}
}

@inproceedings{kayser2021vil,
  title={e-vil: A dataset and benchmark for natural language explanations in vision-language tasks},
  author={Kayser, Maxime and Camburu, Oana-Maria and Salewski, Leonard and Emde, Cornelius and Do, Virginie and Akata, Zeynep and Lukasiewicz, Thomas},
  booktitle={Proceedings of the IEEE/CVF international conference on computer vision},
  pages={1244--1254},
  year={2021}
}

@article{goh2021multimodal,
  title={Multimodal neurons in artificial neural networks},
  author={Goh, Gabriel and Cammarata, Nick and Voss, Chelsea and Carter, Shan and Petrov, Michael and Schubert, Ludwig and Radford, Alec and Olah, Chris},
  journal={Distill},
  volume={6},
  number={3},
  pages={e30},
  year={2021}
}

@article{olsson2022context,
  title={In-context learning and induction heads},
  author={Olsson, Catherine and Elhage, Nelson and Nanda, Neel and Joseph, Nicholas and DasSarma, Nova and Henighan, Tom and Mann, Ben and Askell, Amanda and Bai, Yuntao and Chen, Anna and others},
  journal={arXiv preprint arXiv:2209.11895},
  year={2022}
}

@inproceedings{palit2023towards,
  title={Towards vision-language mechanistic interpretability: A causal tracing tool for blip},
  author={Palit, Vedant and Pandey, Rohan and Arora, Aryaman and Liang, Paul Pu},
  booktitle={Proceedings of the IEEE/CVF International Conference on Computer Vision},
  pages={2856--2861},
  year={2023}
}

@article{neo2024towards,
  title={Towards interpreting visual information processing in vision-language models},
  author={Neo, Clement and Ong, Luke and Torr, Philip and Geva, Mor and Krueger, David and Barez, Fazl},
  journal={arXiv preprint arXiv:2410.07149},
  year={2024}
}

@article{golovanevsky2024vlms,
  title={What do vlms notice? a mechanistic interpretability pipeline for gaussian-noise-free text-image corruption and evaluation},
  author={Golovanevsky, Michal and Rudman, William and Palit, Vedant and Singh, Ritambhara and Eickhoff, Carsten},
  journal={arXiv preprint arXiv:2406.16320},
  year={2024}
}

@inproceedings{salin2022vision,
  title={Are vision-language transformers learning multimodal representations? a probing perspective},
  author={Salin, Emmanuelle and Farah, Badreddine and Ayache, St{\'e}phane and Favre, Benoit},
  booktitle={Proceedings of the AAAI Conference on Artificial Intelligence},
  volume={36},
  number={10},
  pages={11248--11257},
  year={2022}
}

@article{gao2025weight,
  title={Weight-sparse transformers have interpretable circuits},
  author={Gao, Leo and Rajaram, Achyuta and Coxon, Jacob and Govande, Soham V and Baker, Bowen and Mossing, Dan},
  journal={arXiv preprint arXiv:2511.13653},
  year={2025}
}

@article{nam2025causal,
  title={Causal Head Gating: A Framework for Interpreting Roles of Attention Heads in Transformers},
  author={Nam, Andrew and Conklin, Henry and Yang, Yukang and Griffiths, Thomas and Cohen, Jonathan and Leslie, Sarah-Jane},
  journal={arXiv preprint arXiv:2505.13737},
  year={2025}
}

@article{gandelsman2024interpreting,
  title={Interpreting the second-order effects of neurons in clip},
  author={Gandelsman, Yossi and Efros, Alexei A and Steinhardt, Jacob},
  journal={arXiv preprint arXiv:2406.04341},
  year={2024}
}

@article{huo2024mmneuron,
  title={Mmneuron: Discovering neuron-level domain-specific interpretation in multimodal large language model},
  author={Huo, Jiahao and Yan, Yibo and Hu, Boren and Yue, Yutao and Hu, Xuming},
  journal={arXiv preprint arXiv:2406.11193},
  year={2024}
}

@article{rai2024practical,
  title={A practical review of mechanistic interpretability for transformer-based language models},
  author={Rai, Daking and Zhou, Yilun and Feng, Shi and Saparov, Abulhair and Yao, Ziyu},
  journal={arXiv preprint arXiv:2407.02646},
  year={2024}
}

@article{lin2025survey,
  title={A survey on mechanistic interpretability for multi-modal foundation models},
  author={Lin, Zihao and Basu, Samyadeep and Beigi, Mohammad and Manjunatha, Varun and Rossi, Ryan A and Wang, Zichao and Zhou, Yufan and Balasubramanian, Sriram and Zarei, Arman and Rezaei, Keivan and others},
  journal={arXiv preprint arXiv:2502.17516},
  year={2025}
}

@inproceedings{liu2025mechanistic,
  title={Mechanistic Interpretability Meets Vision Language Models: Insights and Limitations},
  author={Liu, Yiming and Zhang, Yuhui and Yeung-Levy, Serena},
  booktitle={The Fourth Blogpost Track at ICLR 2025}
}

@article{meng2022locating,
  title={Locating and editing factual associations in gpt},
  author={Meng, Kevin and Bau, David and Andonian, Alex and Belinkov, Yonatan},
  journal={Advances in neural information processing systems},
  volume={35},
  pages={17359--17372},
  year={2022}
}

@inproceedings{gao2023pal,
  title={Pal: Program-aided language models},
  author={Gao, Luyu and Madaan, Aman and Zhou, Shuyan and Alon, Uri and Liu, Pengfei and Yang, Yiming and Callan, Jamie and Neubig, Graham},
  booktitle={International Conference on Machine Learning},
  pages={10764--10799},
  year={2023},
  organization={PMLR}
}

@article{ye2023satlm,
  title={Satlm: Satisfiability-aided language models using declarative prompting},
  author={Ye, Xi and Chen, Qiaochu and Dillig, Isil and Durrett, Greg},
  journal={Advances in Neural Information Processing Systems},
  volume={36},
  pages={45548--45580},
  year={2023}
}

@inproceedings{hu2024visual,
  title={Visual program distillation: Distilling tools and programmatic reasoning into vision-language models},
  author={Hu, Yushi and Stretcu, Otilia and Lu, Chun-Ta and Viswanathan, Krishnamurthy and Hata, Kenji and Luo, Enming and Krishna, Ranjay and Fuxman, Ariel},
  booktitle={Proceedings of the IEEE/CVF Conference on Computer Vision and Pattern Recognition},
  pages={9590--9601},
  year={2024}
}

@article{menon2024whiteboard,
  title={Whiteboard-of-thought: Thinking step-by-step across modalities},
  author={Menon, Sachit and Zemel, Richard and Vondrick, Carl},
  journal={arXiv preprint arXiv:2406.14562},
  year={2024}
}

@article{wang2025vgr,
  title={Vgr: Visual grounded reasoning},
  author={Wang, Jiacong and Kang, Zijian and Wang, Haochen and Jiang, Haiyong and Li, Jiawen and Wu, Bohong and Wang, Ya and Ran, Jiao and Liang, Xiao and Feng, Chao and others},
  journal={arXiv preprint arXiv:2506.11991},
  year={2025}
}

@inproceedings{olausson2023linc,
  title={LINC: A neurosymbolic approach for logical reasoning by combining language models with first-order logic provers},
  author={Olausson, Theo and Gu, Alex and Lipkin, Ben and Zhang, Cedegao and Solar-Lezama, Armando and Tenenbaum, Joshua and Levy, Roger},
  booktitle={Proceedings of the 2023 Conference on Empirical Methods in Natural Language Processing},
  pages={5153--5176},
  year={2023}
}

@article{bhuyan2024neuro,
  title={Neuro-symbolic artificial intelligence: a survey},
  author={Bhuyan, Bikram Pratim and Ramdane-Cherif, Amar and Tomar, Ravi and Singh, TP},
  journal={Neural Computing and Applications},
  volume={36},
  number={21},
  pages={12809--12844},
  year={2024},
  publisher={Springer}
}

@InProceedings{pmlr-v284-olivier25a,
  title = 	 {Towards a Neurosymbolic Reasoning System Grounded in Schematic Representations},
  author =       {Olivier, Fran\c{c}ois and Bouraoui, Zied},
  booktitle = 	 {Proceedings of The 19th International Conference on Neurosymbolic Learning and Reasoning},
  pages = 	 {420--438},
  year = 	 {2025},
  editor = 	 {H. Gilpin, Leilani and Giunchiglia, Eleonora and Hitzler, Pascal and van Krieken, Emile},
  volume = 	 {284},
  series = 	 {Proceedings of Machine Learning Research},
  month = 	 {08--10 Sep},
  publisher =    {PMLR},
  url = 	 {https://proceedings.mlr.press/v284/olivier25a.html}
}

@article{wang2025multishap,
  title={MultiSHAP: A Shapley-Based Framework for Explaining Cross-Modal Interactions in Multimodal AI Models},
  author={Wang, Zhanliang and Wang, Kai},
  journal={arXiv preprint arXiv:2508.00576},
  year={2025}
}

@inproceedings{fong2017interpretable,
  title={Interpretable explanations of black boxes by meaningful perturbation},
  author={Fong, Ruth C and Vedaldi, Andrea},
  booktitle={Proceedings of the IEEE international conference on computer vision},
  pages={3429--3437},
  year={2017}
}

@article{alvarez2018robustness,
  title={On the robustness of interpretability methods},
  author={Alvarez-Melis, David and Jaakkola, Tommi S},
  journal={arXiv preprint arXiv:1806.08049},
  year={2018}
}

@article{white2019measurable,
  title={Measurable counterfactual local explanations for any classifier},
  author={White, Adam and Garcez, Artur d'Avila},
  journal={arXiv preprint arXiv:1908.03020},
  year={2019}
}

@article{wachter2017counterfactual,
  title={Counterfactual explanations without opening the black box: Automated decisions and the GDPR},
  author={Wachter, Sandra and Mittelstadt, Brent and Russell, Chris},
  journal={Harv. JL \& Tech.},
  volume={31},
  pages={841},
  year={2017},
  publisher={HeinOnline}
}

@InProceedings{pmlr-v238-sun24b,
  title = 	 {Sparse and Faithful Explanations Without Sparse Models},
  author =       {Sun, Yiyang and Chen, Zhi and Orlandi, Vittorio and Wang, Tong and Rudin, Cynthia},
  booktitle = 	 {Proceedings of The 27th International Conference on Artificial Intelligence and Statistics},
  pages = 	 {2071--2079},
  year = 	 {2024},
  editor = 	 {Dasgupta, Sanjoy and Mandt, Stephan and Li, Yingzhen},
  volume = 	 {238},
  series = 	 {Proceedings of Machine Learning Research},
  month = 	 {02--04 May},
  publisher =    {PMLR},
  url = 	 {https://proceedings.mlr.press/v238/sun24b.html}
}

@article{samek2016evaluating,
  title={Evaluating the visualization of what a deep neural network has learned},
  author={Samek, Wojciech and Binder, Alexander and Montavon, Gr{\'e}goire and Lapuschkin, Sebastian and M{\"u}ller, Klaus-Robert},
  journal={IEEE transactions on neural networks and learning systems},
  volume={28},
  number={11},
  pages={2660--2673},
  year={2016},
  publisher={IEEE}
}

@article{petsiuk2018rise,
  title={Rise: Randomized input sampling for explanation of black-box models},
  author={Petsiuk, Vitali and Das, Abir and Saenko, Kate},
  journal={arXiv preprint arXiv:1806.07421},
  year={2018}
}

@article{yeh2019fidelity,
  title={On the (in) fidelity and sensitivity of explanations},
  author={Yeh, Chih-Kuan and Hsieh, Cheng-Yu and Suggala, Arun and Inouye, David I and Ravikumar, Pradeep K},
  journal={Advances in neural information processing systems},
  volume={32},
  year={2019}
}

@article{cambria2023survey,
  title={A survey on XAI and natural language explanations},
  author={Cambria, Erik and Malandri, Lorenzo and Mercorio, Fabio and Mezzanzanica, Mario and Nobani, Navid},
  journal={Information Processing \& Management},
  volume={60},
  number={1},
  pages={103111},
  year={2023},
  publisher={Elsevier}
}

@article{cambria2024xai,
  title={Xai meets llms: A survey of the relation between explainable ai and large language models},
  author={Cambria, Erik and Malandri, Lorenzo and Mercorio, Fabio and Nobani, Navid and Seveso, Andrea},
  journal={arXiv preprint arXiv:2407.15248},
  year={2024}
}

@article{bilal2025llms,
  title={Llms for explainable ai: A comprehensive survey},
  author={Bilal, Ahsan and Ebert, David and Lin, Beiyu},
  journal={arXiv preprint arXiv:2504.00125},
  year={2025}
}

@article{palikhe2025towards,
  title={Towards Transparent AI: A Survey on Explainable Large Language Models},
  author={Palikhe, Avash and Yu, Zhenyu and Wang, Zichong and Zhang, Wenbin},
  journal={arXiv preprint arXiv:2506.21812},
  year={2025}
}

@article{adebayo2018sanity,
  title={Sanity checks for saliency maps},
  author={Adebayo, Julius and Gilmer, Justin and Muelly, Michael and Goodfellow, Ian and Hardt, Moritz and Kim, Been},
  journal={Advances in neural information processing systems},
  volume={31},
  year={2018}
}

@article{zhang2024cross,
  title={Cross-modal consistency in multimodal large language models},
  author={Zhang, Xiang and Li, Senyu and Shi, Ning and Hauer, Bradley and Wu, Zijun and Kondrak, Grzegorz and Abdul-Mageed, Muhammad and Lakshmanan, Laks VS},
  journal={arXiv preprint arXiv:2411.09273},
  year={2024}
}

@article{hooker2019benchmark,
  title={A benchmark for interpretability methods in deep neural networks},
  author={Hooker, Sara and Erhan, Dumitru and Kindermans, Pieter-Jan and Kim, Been},
  journal={Advances in neural information processing systems},
  volume={32},
  year={2019}
}

@article{wiegreffe2019attention,
  title={Attention is not not explanation},
  author={Wiegreffe, Sarah and Pinter, Yuval},
  journal={arXiv preprint arXiv:1908.04626},
  year={2019}
}

@article{clark2019does,
  title={What does bert look at? an analysis of bert's attention},
  author={Clark, Kevin and Khandelwal, Urvashi and Levy, Omer and Manning, Christopher D},
  journal={arXiv preprint arXiv:1906.04341},
  year={2019}
}

@article{posner1980orienting,
  title={Orienting of attention},
  author={Posner, Michael I},
  journal={Quarterly journal of experimental psychology},
  volume={32},
  number={1},
  pages={3--25},
  year={1980},
  publisher={SAGE Publications Sage UK: London, England}
}

@article{he2024efficient,
  title={Efficient modality selection in multimodal learning},
  author={He, Yifei and Cheng, Runxiang and Balasubramaniam, Gargi and Tsai, Yao-Hung Hubert and Zhao, Han},
  journal={Journal of Machine Learning Research},
  volume={25},
  number={47},
  pages={1--39},
  year={2024}
}

@article{dang2024explainable,
  title={Explainable and interpretable multimodal large language models: A comprehensive survey},
  author={Dang, Yunkai and Huang, Kaichen and Huo, Jiahao and Yan, Yibo and Huang, Sirui and Liu, Dongrui and Gao, Mengxi and Zhang, Jie and Qian, Chen and Wang, Kun and others},
  journal={arXiv preprint arXiv:2412.02104},
  year={2024}
}

@article{kunz2024properties,
  title={Properties and challenges of llm-generated explanations},
  author={Kunz, Jenny and Kuhlmann, Marco},
  journal={arXiv preprint arXiv:2402.10532},
  year={2024}
}

@inproceedings{vatsa2024adventures,
  title={Adventures of trustworthy vision-language models: A survey},
  author={Vatsa, Mayank and Jain, Anubhooti and Singh, Richa},
  booktitle={Proceedings of the AAAI Conference on Artificial Intelligence},
  volume={38},
  number={20},
  pages={22650--22658},
  year={2024}
}

@article{kazmierczak2025explainability,
  title={Explainability and vision foundation models: A survey},
  author={Kazmierczak, R{\'e}mi and Berthier, Elo{\"\i}se and Frehse, Goran and Franchi, Gianni},
  journal={Information Fusion},
  volume={122},
  pages={103184},
  year={2025},
  publisher={Elsevier}
}

@article{hu2022lora,
  title={Lora: Low-rank adaptation of large language models.},
  author={Hu, Edward J and Shen, Yelong and Wallis, Phillip and Allen-Zhu, Zeyuan and Li, Yuanzhi and Wang, Shean and Wang, Lu and Chen, Weizhu and others},
  journal={ICLR},
  volume={1},
  number={2},
  pages={3},
  year={2022}
}

@article{wang2024enhancing,
  title={Enhancing the reasoning ability of multimodal large language models via mixed preference optimization},
  author={Wang, Weiyun and Chen, Zhe and Wang, Wenhai and Cao, Yue and Liu, Yangzhou and Gao, Zhangwei and Zhu, Jinguo and Zhu, Xizhou and Lu, Lewei and Qiao, Yu and others},
  journal={arXiv preprint arXiv:2411.10442},
  year={2024}
}

@article{zhu2023minigpt,
  title={Minigpt-4: Enhancing vision-language understanding with advanced large language models},
  author={Zhu, Deyao and Chen, Jun and Shen, Xiaoqian and Li, Xiang and Elhoseiny, Mohamed},
  journal={arXiv preprint arXiv:2304.10592},
  year={2023}
}

@article{yao2023tree,
  title={Tree of thoughts: Deliberate problem solving with large language models},
  author={Yao, Shunyu and Yu, Dian and Zhao, Jeffrey and Shafran, Izhak and Griffiths, Tom and Cao, Yuan and Narasimhan, Karthik},
  journal={Advances in neural information processing systems},
  volume={36},
  pages={11809--11822},
  year={2023}
}

@article{liang2022multiviz,
  title={Multiviz: Towards visualizing and understanding multimodal models},
  author={Liang, Paul Pu and Lyu, Yiwei and Chhablani, Gunjan and Jain, Nihal and Deng, Zihao and Wang, Xingbo and Morency, Louis-Philippe and Salakhutdinov, Ruslan},
  journal={arXiv preprint arXiv:2207.00056},
  year={2022}
}

@article{tang2024grounded,
  title={Grounded relational inference: Domain knowledge driven explainable autonomous driving},
  author={Tang, Chen and Srishankar, Nishan and Martin, Sujitha and Tomizuka, Masayoshi},
  journal={IEEE Transactions on Intelligent Transportation Systems},
  volume={25},
  number={9},
  pages={10617--10635},
  year={2024},
  publisher={IEEE}
}

@article{atakishiyev2024explainable,
  title={Explainable artificial intelligence for autonomous driving: A comprehensive overview and field guide for future research directions},
  author={Atakishiyev, Shahin and Salameh, Mohammad and Yao, Hengshuai and Goebel, Randy},
  journal={IEEE Access},
  volume={12},
  pages={101603--101625},
  year={2024},
  publisher={IEEE}
}

@inproceedings{kim2018textual,
  title={Textual explanations for self-driving vehicles},
  author={Kim, Jinkyu and Rohrbach, Anna and Darrell, Trevor and Canny, John and Akata, Zeynep},
  booktitle={Proceedings of the European conference on computer vision (ECCV)},
  pages={563--578},
  year={2018}
}

@article{kuznietsov2024explainable,
  title={Explainable AI for safe and trustworthy autonomous driving: A systematic review},
  author={Kuznietsov, Anton and Gyevnar, Balint and Wang, Cheng and Peters, Steven and Albrecht, Stefano V},
  journal={IEEE Transactions on Intelligent Transportation Systems},
  volume={25},
  number={12},
  pages={19342--19364},
  year={2024},
  publisher={IEEE}
}

@article{weber2024applications,
  title={Applications of Explainable Artificial Intelligence in Finance—a systematic review of Finance, Information Systems, and Computer Science literature: P. Weber et al.},
  author={Weber, Patrick and Carl, K Valerie and Hinz, Oliver},
  journal={Management Review Quarterly},
  volume={74},
  number={2},
  pages={867--907},
  year={2024},
  publisher={Springer}
}

@article{talaat2024toward,
  title={Toward interpretable credit scoring: integrating explainable artificial intelligence with deep learning for credit card default prediction},
  author={Talaat, Fatma M and Aljadani, Abdussalam and Badawy, Mahmoud and Elhosseini, Mostafa},
  journal={Neural Computing and Applications},
  volume={36},
  number={9},
  pages={4847--4865},
  year={2024},
  publisher={Springer}
}

@article{kim2024human,
  title={Human-centered evaluation of explainable AI applications: a systematic review},
  author={Kim, Jenia and Maathuis, Henry and Sent, Danielle},
  journal={Frontiers in Artificial Intelligence},
  volume={7},
  pages={1456486},
  year={2024},
  publisher={Frontiers}
}

@article{naveed2024overview,
  title={An overview of the empirical evaluation of explainable ai (xai): A comprehensive guideline for user-centered evaluation in xai},
  author={Naveed, Sidra and Stevens, Gunnar and Robin-Kern, Dean},
  journal={Applied Sciences},
  volume={14},
  number={23},
  pages={11288},
  year={2024},
  publisher={MDPI}
}

@article{lopes2022xai,
  title={XAI systems evaluation: A review of human and computer-centred methods},
  author={Lopes, Pedro and Silva, Eduardo and Braga, Cristiana and Oliveira, Tiago and Rosado, Lu{\'\i}s},
  journal={Applied Sciences},
  volume={12},
  number={19},
  pages={9423},
  year={2022},
  publisher={MDPI}
}

@article{bobek2025user,
  title={User-centric evaluation of explainability of AI with and for humans: a comprehensive empirical study},
  author={Bobek, Szymon and Koryci{\'n}ska, Paloma and Krakowska, Monika and Mozolewski, Maciej and Rak, Dorota and Zych, Magdalena and W{\'o}jcik, Magdalena and Nalepa, Grzegorz J},
  journal={International Journal of Human-Computer Studies},
  pages={103625},
  year={2025},
  publisher={Elsevier}
}

\end{document}